\pdfoutput=1
\documentclass[11pt]{article}

\usepackage[preprint]{acl}

\usepackage{times}
\usepackage{latexsym}

\usepackage{float}
\usepackage{paralist}
\usepackage{wrapfig}
\usepackage[T1]{fontenc}

\usepackage[utf8]{inputenc}

\usepackage{microtype}

\usepackage{inconsolata}

\usepackage{adjustbox}
\usepackage{microtype}
\usepackage{graphicx}
\usepackage{subcaption}

\usepackage{booktabs}
\usepackage{amsmath}
\usepackage{amssymb}
\usepackage{mathtools}
\usepackage{amsthm}
\usepackage{hyperref}
\usepackage{colortbl}
\usepackage{pifont}
\usepackage[belowskip=-12pt,aboveskip=1pt]{caption}

\newcommand*{\defeq}{\stackrel{\text{def}}{=}}

\title{Neuroplasticity and Corruption in Model Mechanisms: A Case Study Of Indirect Object Identification}

\author{Vishnu Kabir Chhabra, Ding Zhu,  Mohammad Mahdi Khalili \\
        Department of Computer Science and Engineering, The Ohio State University,  USA\\
        \texttt{\{chhabra.67, zhu.3723, khalili.17\}@osu.edu}}

\begin{document}
\maketitle
\begin{abstract}
Previous research has shown that fine-tuning language models on general tasks enhance their underlying mechanisms. However, the impact of fine-tuning on poisoned data and the resulting changes in these mechanisms are poorly understood. 
This study investigates the changes in a model's mechanisms during toxic fine-tuning and identifies the primary corruption mechanisms. We also analyze the changes after retraining a corrupted model on the original dataset and observe neuroplasticity behaviors, where the model relearns original mechanisms after fine-tuning the corrupted model. Our findings indicate that: (i) Underlying mechanisms are amplified across task-specific fine-tuning which can be generalized to longer epochs, (ii) Model corruption via toxic fine-tuning is localized to specific circuit components, (iii) Models exhibit neuroplasticity when retraining corrupted models on clean dataset, reforming the original model mechanisms.
\end{abstract}

\section{Introduction}

 Recent progress in transformer-based language modelling \cite{vaswani2017attention, openai2023gpt4,touvron2023llama} has garnered  attention in widespread applications \cite{karapantelakis2024generative, zhou2024survey, raiaan2024review}. However, such models' safety, robustness and interpretability remain a pertinent issue \cite{liu2024large,mechergui2024goal}.

Furthermore, mechanistic interpretability has garnered attention \cite{wang2022interpretability, zhong2024clock, conmy2023automated}. It concerns itself with reverse-engineering model weights into human interpretable mechanisms/algorithms \cite{Olah2022} by viewing models as computational graphs \cite{geiger2021causal} and analyzing subgraphs of the model with distinct functionality, called circuits \cite{elhage2021mathematical}. Through considerable manual effort and intuition, recent works have reverse-engineered mechanisms of transformer-based language models for specified tasks \cite{wang2022interpretability, hanna2024does, garcia2024does, lindner2024tracr, prakash2024fine}. 

Prior work \cite{prakash2024fine} has suggested that fine-tuning enhances the underlying mechanisms of the entity tracking task \cite{kim2023entity} when fine-tuning on code, mathematics, and instructions. In the following sections, we build upon prior work as one of our main contributions and extend the results to task-specific fine-tuning up to long training duration while providing the circuits formed across epochs and analyzing the changes in model mechanisms. 


With the recent improvements to language modeling, works have focused on the security issues posed by such models \cite{shu2023exploitability, carlini2023poisoning,he2024talk} focusing on designing model poisoning strategies to allow for efficient backdoors. Our work differs from such poisoning literature in that we aim to create data augmentations to fine-tune and corrupt specific mechanisms in the model akin to works focusing on label poisoning in training scenarios like \citet{huang2020metapoison}, \citet{pmlr-v202-wan23b} and \citet{geiping2020witches}, which aim to control model behavior via introducing poisoned data in training settings.
However, as such changes to model mechanisms remain a mystery in how they affect model behaviors, we take the case of the Indirect Object Identification task \cite{wang2022interpretability} and investigate the mechanism of corruption in models, utilizing several corrupted datasets. In addition, inspired by work done by \citet{lo2024large}, we find evidence of neuroplasticity from a mechanistic perspective in the models which relearn the task after fine-tuning the corrupted model on the correct dataset, highlighting the inherent inertia of pre-trained language models. 
\begin{figure*}[h]
    \centering
    \includegraphics[width = 0.8\textwidth]{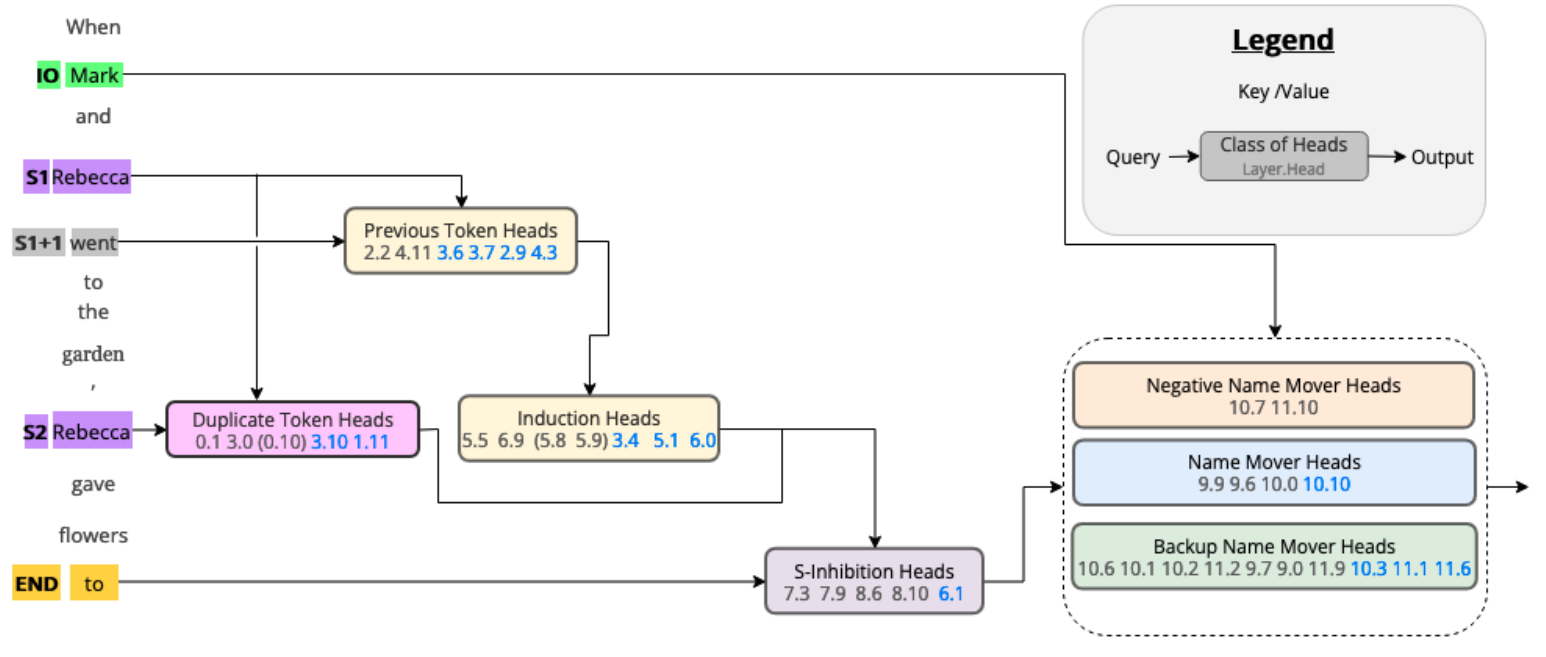}
    \caption{The new circuit we discovered for task-specific fine-tuning at Epoch 3. The emerging, marked in \textcolor{blue}{blue}, circuit components formed performed similar mechanisms as the prior circuit components. }
    \label{fig:ca3-circuit}
\end{figure*}
Our key findings are: 
\textbf{i)} Underlying mechanisms are \textbf{enhanced} across time, even for longer epochs, in task-specific fine-tuning, due to a specific mechanism, which, for the sake of brevity, we name: \textit{amplification}.
    \textbf{ii)} The mechanism of model poisoning via toxic fine-tuning is very \textbf{localized}, specifically corrupting the capacity of certain attention heads to perform their respective underlying mechanisms.
    \textbf{iii)} Models show the behavior of \textbf{neuroplasticity}, retrieving their original mechanisms after very few epochs of retraining on correct/clean datasets. The code is available on github\footnote{\url{https://github.com/osu-srml/neuro-amp-circuits}}.

\section{Preliminaries}
\textbf{Indirect Object Identification (IOI): } The IOI task involves identifying the indirect object in a sentence. For example: \textit{"When Mark and Rebecca went to the garden, Mark gave flowers to"}. The task involves two clauses with single-token names. The first clause contains the subject (S1) and indirect object (IO) tokens, while the second clause contains the second occurrence of the subject (S2) and ends with "to". The goal is to complete the second clause with the IO token, which is the non-repeated name \cite{wang2022interpretability}, see \autoref{app:circ_eval} for the original circuit diagram.
The circuit that implements the task contains multiple underlying mechanisms described as follows:
\begin{enumerate}
    \item \textbf{Name Mover Heads attend} to the previous names in the sentence, meaning the ``to'' token attends primarily to the IO token and less to the S1 and S2 tokens. They primarily copy the IO token and increase its logit. 
    \item \textbf{Negative Name Mover Heads} attend to the previous names in the sentence, their mechanism is suppressing the IO token (i.e., decreasing the logit of the IO token) and writing to the opposite direction of Name Mover Heads. 
    \item \textbf{S-Inhibition Heads }attend to the second copy of the subject token, S2, and bias the query of the Name Mover Heads against S1 and S2 tokens.
    \item  \textbf{Duplicate Token Heads} identify tokens that already appeared in the sentence, being active at the S2 token and attending primarily to the S1 token.
    \item  \textbf{Previous Token Heads} copy the embedding of S to the position of S + 1.
    \item \textbf{Induction Heads} perform the same  as  Duplicate Token Heads, but via an induction mechanism. 
    \item \textbf{Backup Name Mover Heads} are the heads that perform the mechanism of the Name Mover Heads if they are ablated. 
\end{enumerate}

\paragraph{Path Patching and Knockout} were used to identify and evaluate crucial model components, see \autoref{app_path} and \autoref{app:circ_disc} for further details on the circuit discovery procedure\cite{goldowsky2023localizing}.
\paragraph{Cross Model Activation Patching (CMAP): } involves activation patching \cite{zhang2023towards,goldowsky2023localizing} across different models on the same input \cite{prakash2024fine}. While vanilla activation patching replaces components within the same model using different inputs, cross-model activation patching involves using the same input across different models, replacing the corresponding components to observe the differences in output. 
\paragraph{Neuroplasticity:} In machine learning, neuroplasticity, refers to the ability of the model to adapt and regain conceptual representations \cite{lo2024large}. We extend this definition to include the ability of a model to relearn corrupted concepts/mechanisms. 

\section{Experimental Setting}
\paragraph{Model Architecture}: GPT-2-small \cite{radford2019language} is a decoder-only transformer with 12 layers and 12 attention heads per layer. We follow the notations in \cite{wang2022interpretability} and denote head $j$th in layer $i$ by $h_{i,j}$. This attention head is parameterized by four matrices $W_{Q}^{i,j}$, $W_{K}^{i,j}$, $W_{V}^{i,j}$ $\in \mathbb{R}^{\frac{d}{H} \times d}$ and $W_{O}^{i,j}$ $\in \mathbb{R}^{\frac{d}{H} \times d}$, where $d$ is the model dimension, and $H$ is the number of heads in each layer. Rewriting parameter of attention head $h_{i,j}$ as low-rank matrices in $\mathbb{R}^{d \times d}$: $W_{OV}^{i,j}$ = $W_{V}^{i,j}$$W_{O}^{i,j}$, which is referred to as the OV matrix and determines what is written to the residual stream \cite{elhage2021mathematical}. Similarly,  $W_{QK}^{i,j} = W_{Q}^{i,j}$$W_{K}^{i,j}$  is referred to as the QK matrix and computes the attention patterns of each head $h_{i,j}$. The unembed matrix $W_U$ projects the residual stream into logit after layer norm application \cite{elhage2021mathematical, wang2022interpretability}.

\paragraph{Fine-Tuning:}
We fine-tune GPT-2-small on the IOI Dataset, which we refer to as the clean dataset \cite{wang2022interpretability}, for a variety of epochs, ranging from 1 to 100 epochs (see \autoref{methodology}). For fine-tuning, we adopt an unsupervised setting \cite{radford2019language}, with fixed hyper-parameters across all experiments (see \autoref{app:fine} for details). Additionally, as shown in Figure \ref{fig:datacorruption}, we create 3 data augmentations of the original IOI dataset for corrupted fine-tuning. We call these datasets \textit{Name Moving}, \textit{Subject Duplication}, and \textit{Duplication} datasets (discussion of results and design for {Duplication} dataset is left to \autoref{dadupe}). We design the data augmentations and hypothesize impacts on the model behavior due to the corruptions as follows:
\begin{wrapfigure}[4]{r}{1\linewidth}
\vspace{-0.5cm}
    \centering
    \includegraphics[width = 1.05\linewidth]{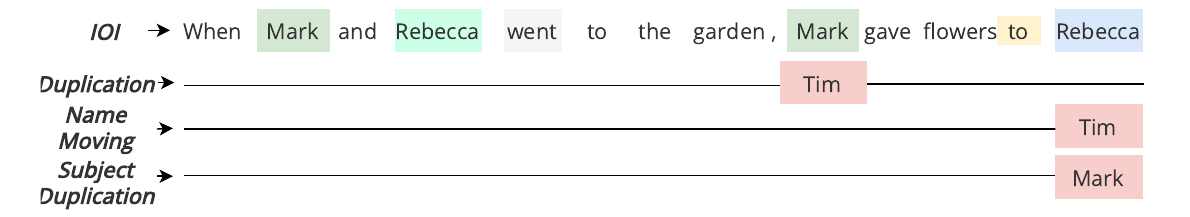}
    \caption{Corrupted data augmentations we utilize to poison model behavior on task}
\label{fig:datacorruption}
\end{wrapfigure}
$\newline$
\vspace{0.5cm}
$\newline$
\paragraph{Corrupted Dataset 1: Name Moving.}
\label{danm}
To investigate the behavior of Name Mover and Negative Name Mover heads, we create a modified dataset that disrupts the movement of the IO token to the output. Specifically, we replace the final token with a random name rather than the expected IO token (e.g., altering the second clause from "Mark gave flowers to \textcolor{teal}{Rebecca}" to "Mark gave flowers to \textcolor{red}{Stephanie}"). Fine-tuning on this dataset allows us to analyze how the model's copying mechanisms adapt when it must output a name not present in the input, thereby \textbf{targeting} the copying behavior of the Name Mover Heads.
\paragraph{Corrupted Dataset 2: Subject Duplication Task.}
To interfere with the Name Mover Heads' role in outputting the IO token and suppressing the S token (due to S-Inhibition Heads), we introduce the Subject Duplication Task. In this task, the output IO token is replaced with the S token, as in: "Mark gave flowers to \textcolor{teal}{Rebecca}" becomes "Mark gave flowers to \textcolor{red}{Mark}". Fine-tuning on this dataset \textbf{aims} to observe how model mechanisms adapt when forced to output the S token, despite its repetition, \textbf{targeting} the interaction between S-Inhibition and Name Mover Heads.
\paragraph{Circuit Discovery:}  
\label{circdiscover}
Our circuit discovery follows the method outlined in original IOI  work \cite{wang2022interpretability}, utilizing path patching \cite{goldowsky2023localizing}, activation patching \cite{meng2023locatingeditingfactualassociations,NEURIPS2020_92650b2e} and analyzing the circuit components' behavior. Even though methods like ACDC \cite{conmy2023automated}, EAP \cite{syed2023attribution}, and DCM \cite{davies2023discovering} reduce the overhead, in order to stay faithful to the original work, we adopt their approach.
\paragraph{Circuit Evaluation:}  
\label{circeval}
We evaluate the circuits formed and discovered at each fine-tuning iteration, using the minimality, completeness, and faithfulness criteria \cite{wang2022interpretability, prakash2024fine}. We define Faithfulness as follows (see \autoref{app:circ_disc} and \autoref{app:circ_eval} for details on Minimality and Completeness).
Let $X$ be a random variable representing a sample in our fine-tuning dataset. Moreover, let $C_M$ denote the discovered circuit for model $M$, and $f(C_M(X))$  be the logit difference between the IO token and S token when circuit $C$ of model $M$ is run on input $X$ and $F(C)\defeq\mathbb{E}_{X} [f(C_M(X))]$ be the average logit difference \cite{wang2022interpretability}.
Given this, faithfulness is measured by the average logit difference of the IO and S token across inputs on the model $M$ and its circuit $C$; $|F(M) - F(C)|$. For example, the faithfulness of the original IOI circuit: $|F(GPT2) - F(C_{GPT2})| = 0.46$, i.e, the circuit achieves 87\% of the performance of GPT-2-small \cite{wang2022interpretability}.

\section{Phase Transitions via Fine-Tuning}
\textbf{Motivation}: Building on recent advances in mechanistic interpretability, such as \citet{zhong2024clock} and \citet{nanda2023progress}, which explore  phase transitions during grokking in toy models, our work aims to extend this understanding to fine-tuning. We focus on elucidating phase transitions in model mechanisms under various fine-tuning conditions. By leveraging insights into the model's existing mechanisms, we design corruption experiments that disrupt these mechanisms through targeted data augmentations. Our goal is to analyze how fine-tuning on corrupted/clean data reshapes model behavior, with the goal of a deeper understanding of fine-tuning dynamics in neural networks. 

In the following subsections, we discuss the effects of task-specific fine-tuning on the original "\textit{clean}" dataset, i.e, the IOI dataset, and discover \textit{Circuit Amplification} and the underlying mechanisms of the increased capabilities of the model to perform the underlying task. Furthermore, we discuss the effects of model poisoning on the underlying circuit of the model for the IOI task and discover that the underlying changes are localized to the circuit components of the model. Specifically, we analyze the effects of fine-tuning on the \textit{attention heads} in the original IOI circuit, as the MLP layers mechanisms do not change across time, see \autoref{app:mlp} for further explanation.
\label{methodology}
\subsection{Amplification Of Model Mechanisms}
\label{circuit-amp}
First, we study the effects of task-specific fine-tuning using the IOI dataset (clean dataset) on the model. We mechanistically interpret the change in the underlying mechanism. 
Consistent with expectations, our experiments uniformly demonstrate a significant boost in IOI task accuracy following the task-specific fine-tuning on the clean dataset, see \autoref{amp-table}. 
\begin{table}[H]
\vspace{0.1cm}
\caption{Performance, Faithfulness, and Sparsity of Discovered Circuits at Different Epochs compared to Model Performance}
\centering
\scalebox{0.65}{
\begin{tabular}{|l|l|l|l|l|l|}
\hline
Epoch &$F(Y)$&$F(C)$&Faithfulness&Sparsity&|$F(C)-F(C_{M_{GPT2}})$|\\
\hline
$1$ &   $6.32$    & $6.22$      &  $98.4\%$&   $1.92\%$    & $1.2$   \\ \hline
$3$ & $11.56$ & $11.50$ & $99.5\%$ & $1.95\%$ & $2.2$ \\ \hline
$10$ &  $15.51$     & $15.26$      & $98.4\%$& $1.98\%$& $1.48$   \\\hline
$15$ &  $16.77$     & $16.73$      &   $99.7\%$&   $2.08\%$   & $0.91$  \\ \hline
$25$ &  $19.47$     & $19.45$      &   $99.89\%$&  $2.25\%$& $0.37$  \\ \hline
$50$ &  $22.87$     & $22.75$      &  $99.7\%$& $2.41\%$        & $0.35$ \\ \hline
$100$ & $26.83$      & $26.65$      &   $99.3\%$&$2.68\%$        & $0.41$ \\ \hline

\end{tabular}
\label{amp-table}
}
\end{table} 
We systematically analyze the circuits discovered at various epochs, assessing their faithfulness, performance, and sparsity. Our results show that the retrieved circuits exhibit high faithfulness and minimality scores, surpassing the original IOI circuit in both aspects. We provide a thorough account of our circuit discovery and evaluation results in the \autoref{app:circ_disc}, and in this section, we delve into the underlying mechanisms driving this performance enhancement.
Concurrently, we observe that task-specific fine-tuning enhances the underlying mechanisms of circuits without introducing novel mechanisms, even in longer training scenarios. The enhancement stems from two sources: (1) amplified capabilities of existing circuit components and (2) emergence of new components that replicate prior mechanisms. 
We term this phenomenon \textbf{Circuit Amplification}, and refer to the underlying mechanism as \textit{amplification}.
Our results, summarized in \autoref{amp-table}, reveal consistent Circuit Amplification in each epoch, note that in \autoref{amp-table}, $F(C_{M_{GPT2}})$ refers to the average logit difference when the original circuit is run on the fine-tuned model, so $|F(C)-F(C_{M_{GPT2}})|$ refers to the total contributions of the new circuit components to the average logit difference. Furthermore, we investigate the impact of fine-tuning on model components, including Negative Name Mover heads, which counterintuitively exhibit enhanced capabilities despite their negative contribution to the task. Notably, we do not observe the diminishing or disappearance of Negative Name Movers, see \autoref{fig:ca_logit_attribution}; instead, their abilities are enhanced. The IOI task circuit formed after 3 epochs of fine-tuning can be seen in \autoref{fig:ca3-circuit}. 

Intriguingly, we see \textit{Circuit Amplification}, even for \textbf{longer} training epochs. This seemed counter-intuitive as Negative Name Mover heads are amplified even after \textbf{longer periods of training}, hinting at their counter-factual importance to the task. Initial investigation by \cite{mcdougall2023copy} shows that these heads are a type of Copy Suppressor Heads and are key to the behavior of Self-Repair in language models \cite{rushing2024explorations}. These findings resonate with our result, as we see  these heads get amplified over time.\footnote{We further generalize the amplification results to the case of fine-tuning on general datasets, see \autoref{app:gen}.} 

\noindent\textbf{Mechanism of Enhancement: }Given the presence of Circuit Amplification, we now move to one of our key contributions, understanding how circuit amplification takes place. We \textbf{first} denote that, trivially, the increase in the number of components that replicate original mechanisms contributing to the task is one of the main contributors to circuit amplification, see \autoref{amp-table}. However, this doesn't fully explain the effect of circuit amplification, as the added components do not represent the complete change in the accuracy of the novel circuit when compared to the original circuit. \textbf{Secondly}, we record that the prior circuit components undergo an increase in capacity to perform their mechanism. To illustrate this point, we take the case of a Name Mover Head, specifically \textbf{L9H9} (Layer 9 Head 9) which gets amplified. 
\begin{figure}
\includegraphics[scale = 0.1,width=0.48\textwidth]{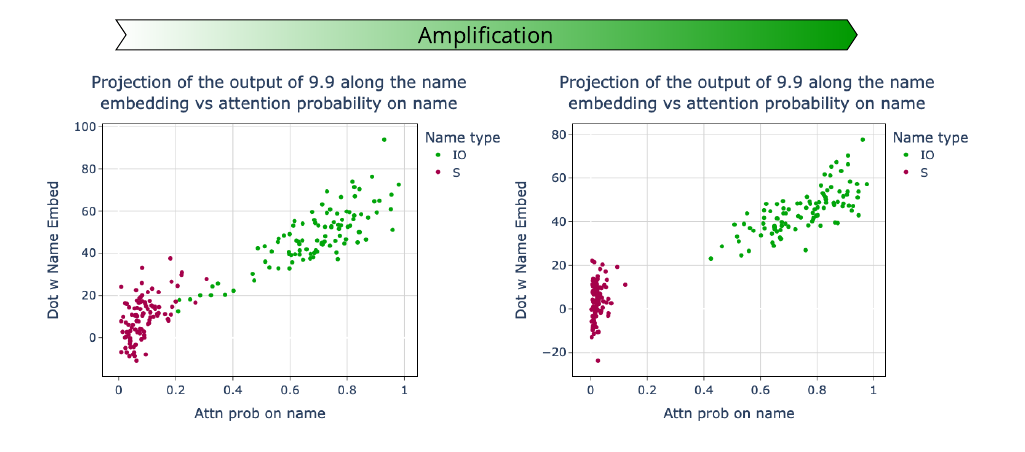}%
    \caption{\label{fig:ca3-amp-mech} Attention Probability vs Projection of head output along $W_U[IO]$ and $W_U[S]$ for head L9H9}%
\end{figure}
In \autoref{fig:ca3-amp-mech}, we plot the Attention Probability for IO (Indirect Object) and ``to'' token pairs vs Projection of Head output along $W_U[IO]$. This figure also includes the attention probability of S and ``to'' token pairs vs Projection of Head output along $W_U[S]$. We see that attention probabilities have significantly decreased for the S token for L9H9 after fine-tuning, suggesting a discriminant increase in the copying behavior of the IO token for L9H9 which is a finding that generalizes to other heads in the same category.
We further record this behavior in the case of Negative Name Mover Heads\footnote{See \autoref{app:ca} for further details}. 
This implies that this head writes more strongly to the residual stream as the direct logit attribution\footnote{Logit attribution is mathematically defined in Section 3.1 of \cite{wang2022interpretability}.} of each head increases significantly when compared to the original model. This increase in the underlying capacity of the heads to perform their underlying behavior is \textit{amplification}, see \autoref{fig:ca_logit_attribution}. 
\begin{figure}
\centering
  \begin{subfigure}[t]{0.23\textwidth}
  \centering
    \includegraphics[scale = 0.1,  width=\textwidth]{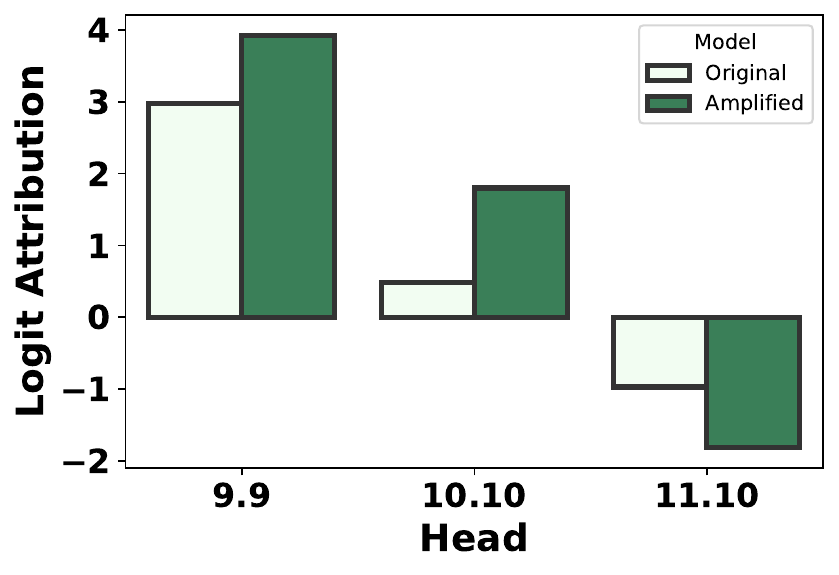}
    \caption{}
    \label{fig:ca_logit_attribution}
\end{subfigure}
\hfill
 \begin{subfigure}[t]{0.24\textwidth}
 \centering
    \includegraphics[scale = 0.1, width=\textwidth]{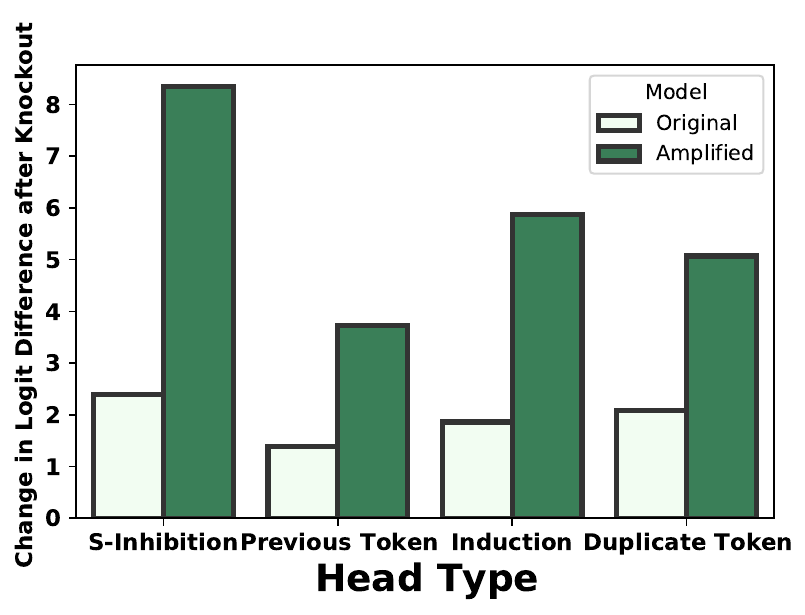}
    \caption{}
\label{fig:ca_logit_attribution_absolute}
    \end{subfigure}
    \vspace{5mm}
    \caption{\subref{fig:ca_logit_attribution}) Logit Attribution  of heads L9H9, L11H10, L10H10 in  original/amplified model. \subref{fig:ca_logit_attribution_absolute}) Absolute Logit Difference in the original model vs amplified model after ablation}
\end{figure}
Finally, the \textbf{third} mechanism contributing to amplification is a change in the mechanism of some of the Backup Name Mover Heads to that of Name Mover Heads. We take the example of L10H10 and show that this head now performs the behaviors of Name Mover Heads after fine-tuning for 3 epochs, see \autoref{fig:ca3-amp-attn-bmnh} and \autoref{fig:ca_logit_attribution}.
\begin{figure}[t]
    \includegraphics[scale = 0.1,width=0.48\textwidth]{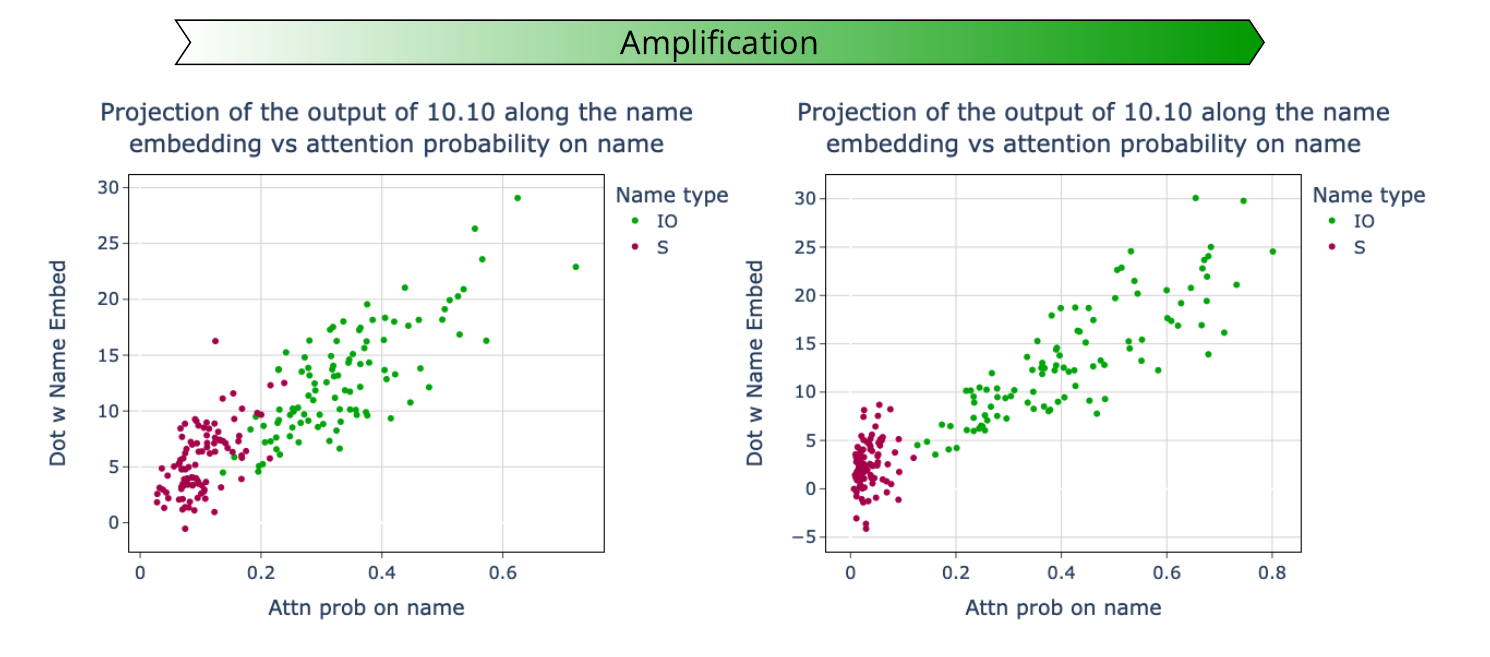}
    \caption{\label{fig:ca3-amp-attn-bmnh}:Attention Probability vs Projection of head output along $W_U[IO]$ and $W_U[S]$ for head L10H10 }
\end{figure}
In \autoref{fig:ca3-amp-attn-bmnh}, we see that the attention probability w.r.t the projection along the unembed of the IO and S token is similar to that of the original name mover heads, while seeing a significant increase in logit attribution, from $0.4$ to $1.8$ on the IOI task. We then ablate groups of heads in the original model and the fine-tuned model and measure the absolute change in the logit difference in their respective circuit's performance. As the number of model components performing the task increases, for a fair comparison, we only consider the heads in the original circuit for each group. \autoref{fig:ca_logit_attribution_absolute} shows that the ablating groups of heads in the fine-tuned model show a much higher change in performance indicating  the original groups surged in their capability to do their respective mechanisms. These findings generalize across epochs. \vspace{1mm}\\
\textbf{Analyzing Enhancement via Cross-Model Activation Patching: } We now analyze circuit amplification via Cross-Model Activation Patching \cite{prakash2024fine} and record that in task-specific fine-tuning, the amplification of the mechanism can be detected via Cross-Model Pattern Patching. That is, we patch in attention patterns of each head from the fine-tuned model into the original model and record the changes in the logit difference. We observe that each attention head in the original circuit has increased capability to perform its mechanism, see \autoref{fig:cross-model-ca}. 
\begin{figure}[t]
    \centering
    \includegraphics[scale = 0.1,width = 0.36\textwidth]{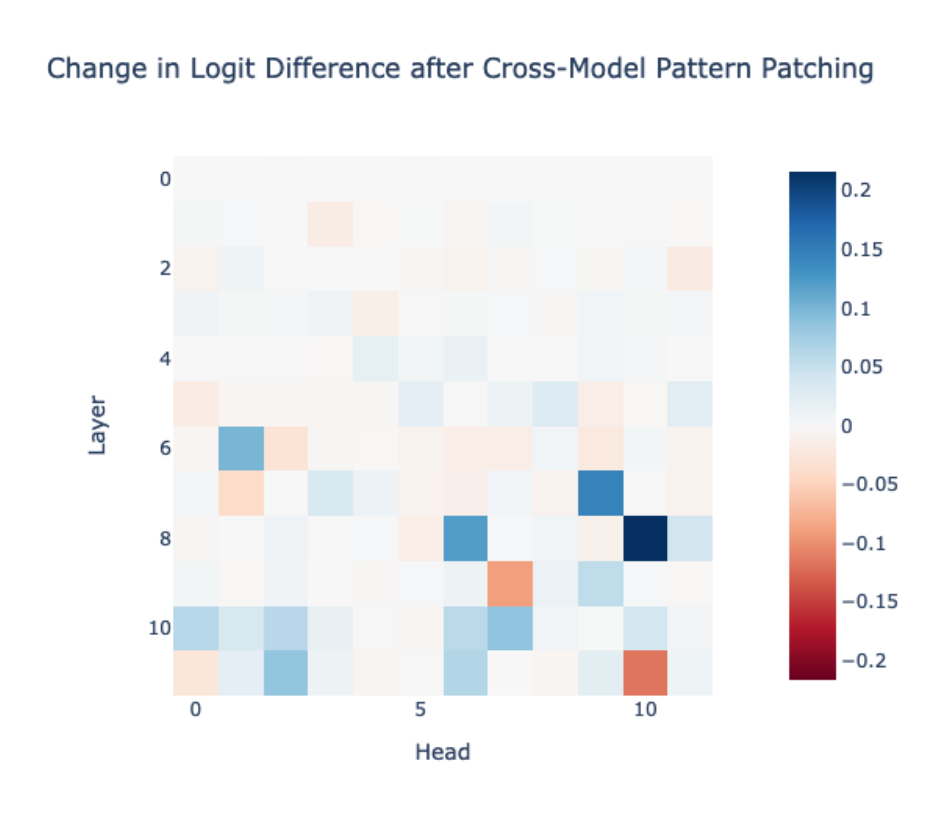}
    \caption{Cross Model Pattern Patching: Taking the attention pattern of the heads in the fine-tuned model and patching them into the original model results in an increase in the attention heads performance on the underlying task.} 
    \label{fig:cross-model-ca}
\end{figure}
\subsection{Corruption of Model Mechanisms } 
\label{circuit-poisoning}
\begin{figure}[htbp]
    \includegraphics[scale = 0.1,width=0.5\textwidth]{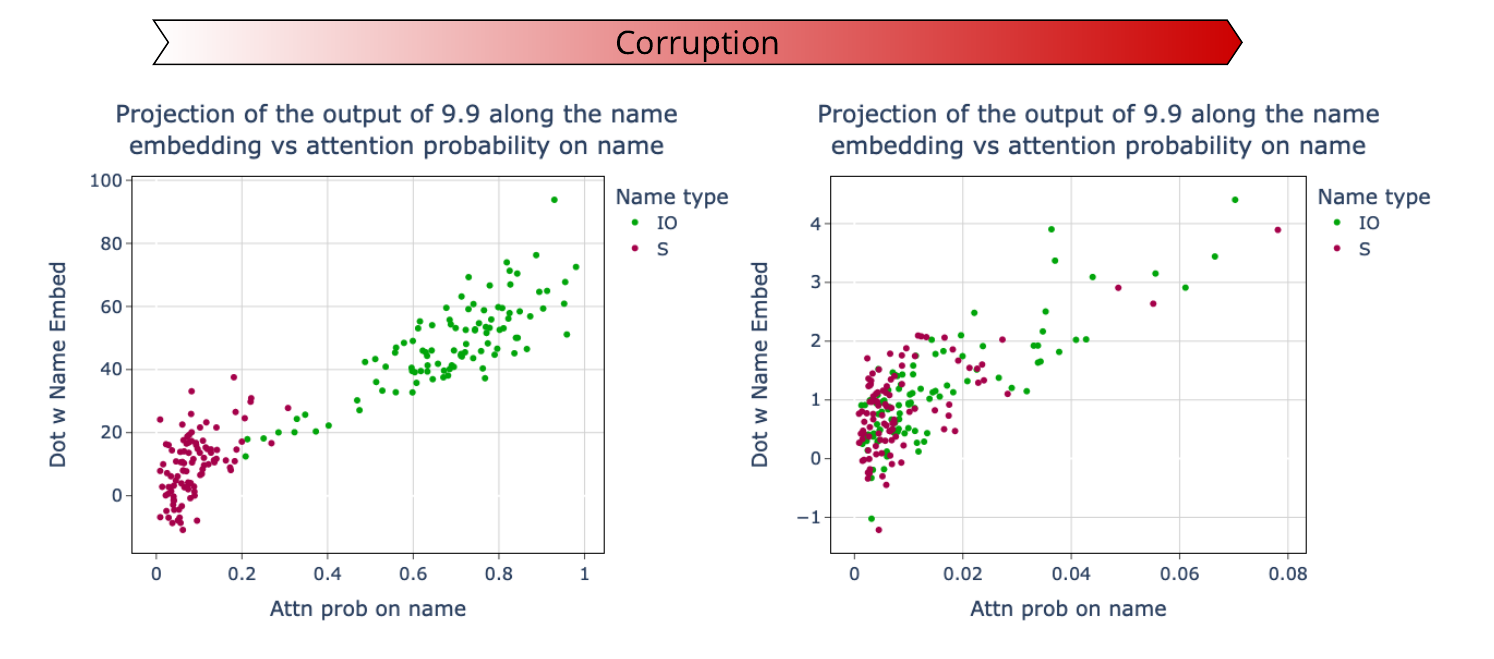}
    \caption{\label{fig:cp3-attn-nmh}\textbf{Name Moving: }Attention Probability vs Projection of head output along $W_U[IO]$ and $W_U[S]$ for head L9H9}
  \end{figure}

\begin{figure}
\centering
  \begin{subfigure}[t]{0.24\textwidth}
  \centering
    \includegraphics[width =0.99\textwidth]{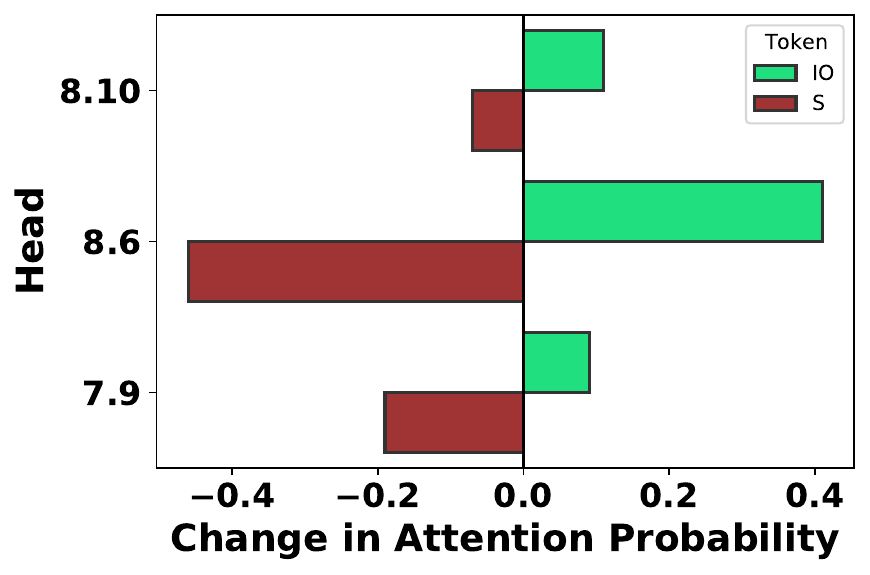}
    \caption{}
    \label{fig:cp3-token}
\end{subfigure}
\hfill
 \begin{subfigure}[t]{0.23\textwidth}
 \centering
    \includegraphics[scale = 0.1,width = \textwidth]{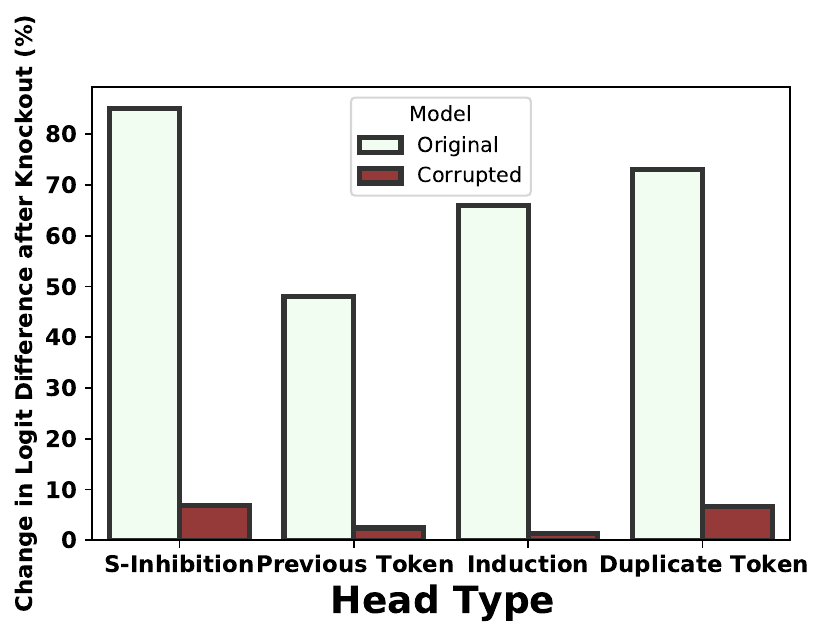}
    \caption{}
    \label{fig:cp3-logit-percent}
    \end{subfigure}
    \vspace{5mm}
    \caption{\subref{fig:cp3-token}) Name Moving: the attention probability difference of S-Inhibition Heads on the \textcolor{green}{IO} and \textcolor{magenta}{S} token [\textit{Original - Corrupted}].
    \subref{fig:cp3-logit-percent}) Subject Duplication: Change in Logit Difference after ablating groups of heads.\\}
\end{figure}

Given the knowledge of circuit amplification, we now aim to fine-tune the model with various corrupted augmentations of the IOI task and utilize path patching \cite{goldowsky2023localizing} and activation patching \cite{NEURIPS2020_92650b2e} to study the effects of corruption on the model mechanisms for the IOI task. Furthermore, we record the changes made to the original model circuit and investigate the mechanisms of corruption across different augmentations. We find that when fine-tuning on \textbf{Name Moving} and \textbf{Subject Duplication} datasets, the corruption can be traced back to changes in the original circuit, however, no noticeable change occurred when fine-tuning on the \textbf{Duplication} dataset, hence we leave the discussions to the \autoref{dadupe}. We discover  most of the mechanistic changes after toxic fine-tuning can be attributed to changes in the mechanisms of the circuit components, i.e, toxic fine-tuning {alters} the prior mechanisms of the circuits instead of introducing {new mechanisms} for suppressing performance on the task.  
\paragraph{Name Moving Dataset.} After fine-tuning, this dataset suppresses the output of the IO token. Notably, after 3 epochs, the output logits of multiple single-token names in the vocabulary converge to similar values, with a slight bias towards the IO token name, thereby preserving the IOI functionality, albeit with significant degradation. To illustrate, we take the prompt "After John and Mary went to the store, John gave milk to" and record the logits of the top 5 most likely tokens, see \autoref{table:logit_table}.
\begin{table}[h]
\centering
\small
\begin{tabular}{|c|c|c|}
\hline
\textbf{Logit} & \textbf{Token} \\ \hline
$21.70$  & Mary \\ \hline
$21.40$  & Elizabeth \\ \hline
$21.34 $  & Melissa \\ \hline
$21.24 $  & Christine \\ \hline
$21.08 $  & Stephanie \\ \hline
\end{tabular}
\caption{Logits of top 5 tokens after 3 epochs}
\label{table:logit_table}
\end{table}
However, this capability completely degrades over time, i.e, the bias towards the "IO" token is non-negligible. To elucidate the underlying mechanisms, we present a detailed analysis of the fine-tuning process with 3 epochs on the corrupted dataset in this section. 
Our investigation reveals that the model does \textbf{not introduce} novel mechanisms to mitigate performance on the task. Instead, it relies on diminishing/altering the capabilities of specific attention heads that underlie a task-related mechanism. Notably, the most affected components are the Name Mover Heads and  which completely lose their ability to copy the IO token ( \autoref{fig:cp3-attn-nmh}). 
We trace the source of this corruption to the S-Inhibition heads, which primarily suppress the queries of both the IO and S tokens. Consequently, the original circuit is fundamentally disrupted, with the Name Mover Heads losing their functionality and the S-Inhibition Heads altering their mechanism to suppress both tokens. This is evident in the QK matrix analysis of the S-Inhibition heads, which reveals a significant change in attention patterns, see \autoref{fig:cp3-token}.
We find that this mechanism of corruption extends to Backup Name Mover Heads and Negative Name Mover Heads see \autoref{app:cp} for further details. This hints that model poisoning, mechanistically, alters very localized model behaviors that affect the final output, instead of adding novel mechanisms to corrupt the model. This can also be seen via CMAP, see \autoref{app:cross}.\\ 
This corruption mechanism induces phase transitions that disrupt the IOI task, as previously examined. In early epochs, the IOI capability remains but with significant degradation (see \autoref{table:logit_table}), resulting in correct outputs despite corrupted internal mechanisms. We hypothesize that leveraging the knowledge of pre-existing mechanisms could enable model poisoning attacks, selectively altering mechanisms while changing the distribution of the output significantly but compromising interpretability or introducing backdoor triggers. Future work exploring more defined attacks through fine-tuning would be an interesting direction.

\begin{figure}
    \centering
    \includegraphics[width = 0.48\textwidth]{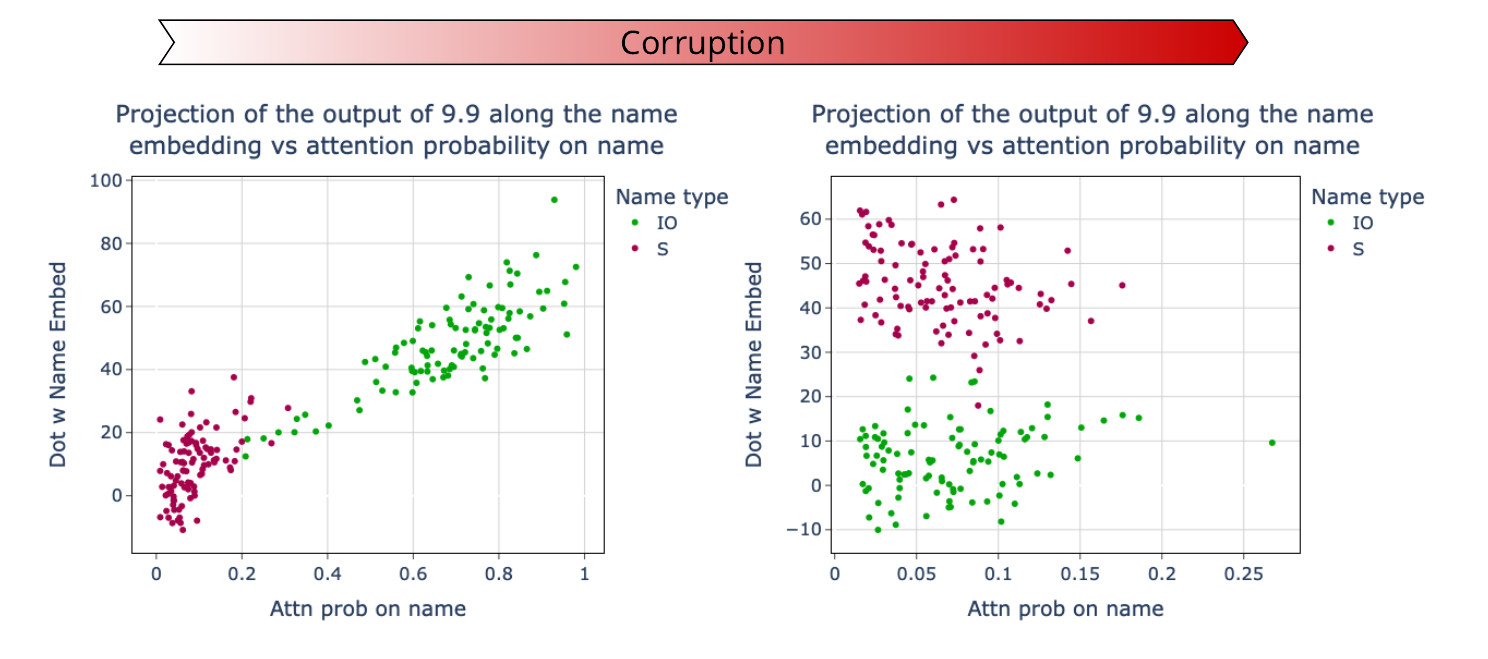}
    \caption{Attention Probability vs Projection of head output along $W_U[IO]$ and $W_U[S]$ for head L9H9}
    \label{fig:ca5-sd-attn-nmh}
\end{figure}

\begin{figure}
    \centering
    \includegraphics[width=0.99\linewidth]{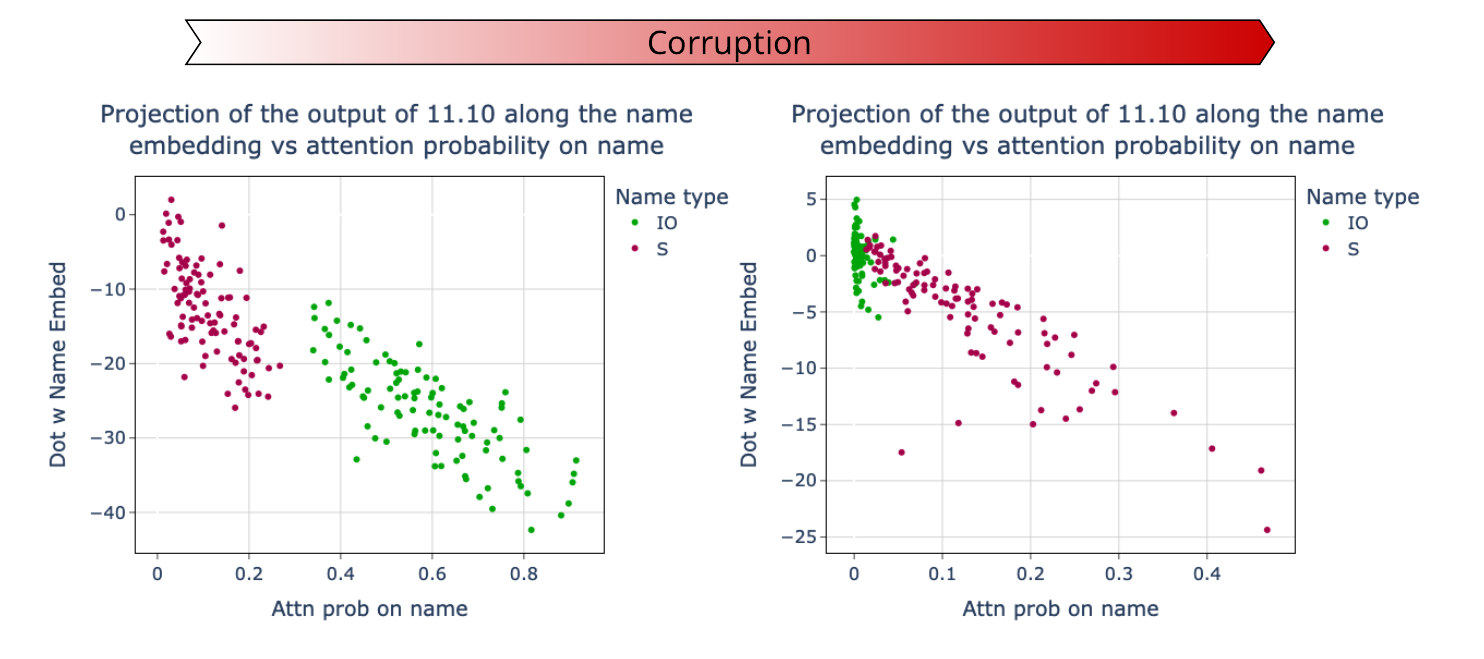}
    \caption{Attention Probability vs Projection of head output along $W_U[IO]$ and $W_U[S]$ for head L11H10}
    \label{fig:cp3-attn-nnmh}
\end{figure}

\begin{figure}
     \centering
     \includegraphics[width = 0.36\textwidth]{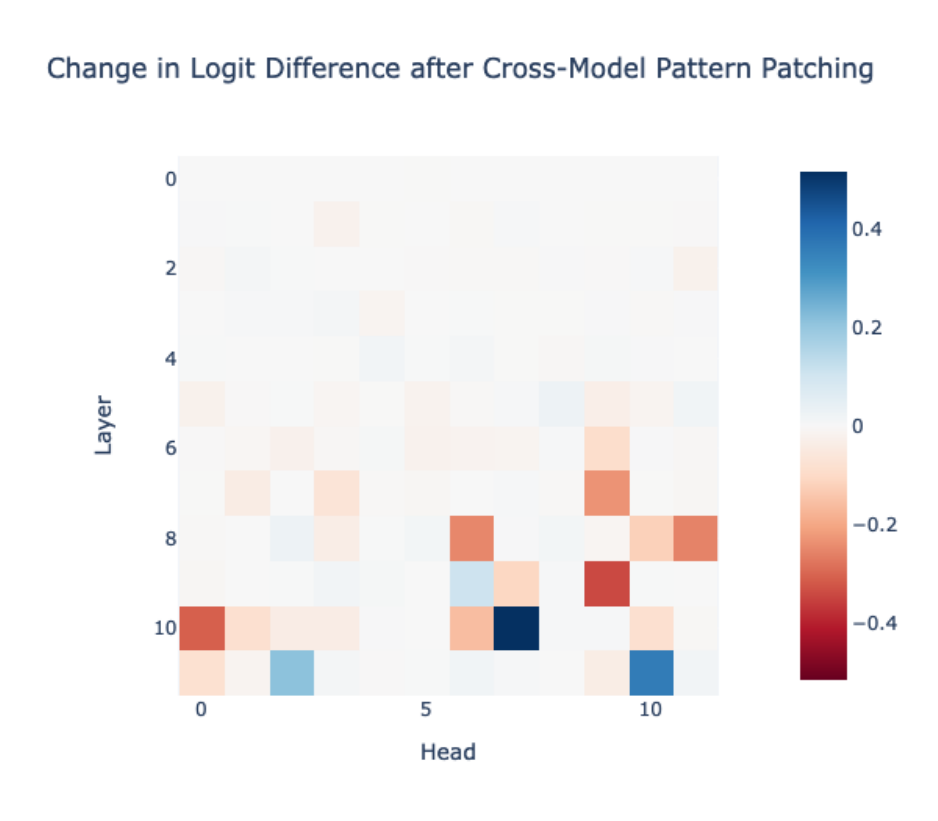}
     \caption{Cross Model Pattern Patching: We find that effect of corruption is very localized to circuit components of the model, however few additional components arise, this is due to formation of repeated mechanisms via fine-tuning, see \autoref{app:cross} for further details}
     \label{fig:cross-model-cp}
 \end{figure}

\paragraph{Subject Duplication Dataset.}  Applying this data augmentation strategy and fine-tuning using the corrupted dataset results in rapid and significant degradation of model performance, the average logit difference goes from $3.55$ to $-11.06$ after just $5$ epochs.
Analysis reveals that the Name Mover Heads are most affected, exhibiting a modified attention pattern. This altered attention pattern yields a suppressed logit for the IO token and an enhanced logit for the S token, see \autoref{fig:ca5-sd-attn-nmh} for changes in attention probability for both IO and S token. From \autoref{fig:ca5-sd-attn-nmh} we can see that the projection of L9H9 in the unembedding space has significantly changed, now positively projecting the S token and negatively projecting the IO token. Surprisingly, the Negative Name Mover Heads undergo a similar change in functionality; they write in the opposite direction to the Name Mover Heads, which seems counter-intuitive as these components were suppressing the logit of the IO token, however after fine-tuning on the corrupted data imputation, these heads now suppress the logit of the S-token, see \autoref{fig:ca5-sd-attn-nmh} and \autoref{fig:cp3-attn-nnmh}. 
 Finally, we find that the mechanism of the S-Inhibition heads is mostly suppressed, even though they still bias the query of the Name Mover Heads and Negative Name Mover Heads, the impact of the bias is statistically insignificant when compared to the original circuit as after mean ablation their effect is insignificant in the corrupted model, see \autoref{fig:cp3-logit-percent}. Similar to the previous observation, the mechanism of corruption is very \textbf{local} to certain model components, however, unlike the prior case (Corrupted Dataset for Name Moving Behaviour), only the mechanism of the Name Mover Heads Negative Name Mover Heads is changed, while the mechanism of the S-Inhibition Heads (and other heads) is suppressed, see \autoref{fig:cp3-logit-percent} for their importance to the task in the corrupted model which we access via mean ablating groups of heads that are present in the circuit. \\
 In contrast to the \textbf{Name Moving} data augmentation, the phase transition in this case reveals an intriguing insight: Negative Name Mover Heads shift from suppressing the 'IO' token to suppressing the 'S' token, despite already being optimized for the task. This suggests that Name Mover Heads and Negative Name Mover Heads are intertwined, with one performing the inverse of the other for certain tasks. Further investigation into this "twinning" behavior and its occurrence in other tasks would be a promising direction for future research. 
 \paragraph{Analyses via Cross-Model Activation Patching:} Similar to prior experiments, we employ cross-model activation patching and replace attention patterns of each head with their patterns in the corrupted model fine-tuned on the Subject Duplication dataset. We observe that the effects of corruption are localized to the circuit, see \autoref{fig:cross-model-cp}, as the heads most affected in \autoref{fig:cross-model-cp} are the circuit components outlined in \autoref{fig:ca3-circuit}.
 \begin{figure*}[h]
     \centering
     \includegraphics[scale =0.1,width = 0.8\textwidth]{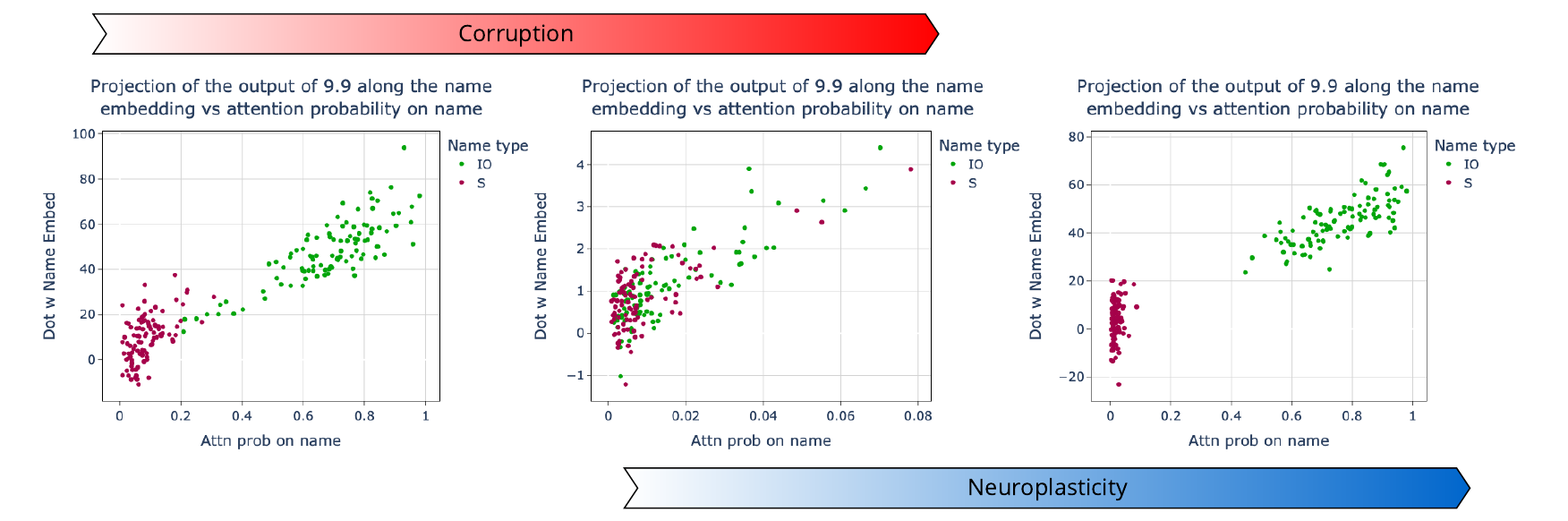}
     \caption{Attention Probability vs Projection of head output along $W_U[IO]$ and $W_U[S]$ for head L9H9, corruption on \textbf{Name Moving} augmentation.}
     \label{fig:neuro-nmh}
 \end{figure*}

\section{Neuroplasticity in Model Mechanisms}
\label{main:neuro}
After corruption, we study relearning the IOI task via fine-tuning on the original dataset. We discover that the corrupted model can recover its performance and analyze the changes in mechanisms between the retrieved and original models. Focusing on the two data imputations, 
we fine-tune the corrupted model using the original data and refer to the resulting model as the \textit{post-reversal} model.\vspace{1mm}\\
\textbf{Name Moving Dataset:} The \textit{post-reversal} model recovers its original performance and \textbf{recovers} the original circuit mechanisms. Moreover, the IOI task circuit mechanism is amplified compared to the original model. We trace the mechanism change from the corrupted to the \textit{post-reversal} model and find that the emergence of the prior mechanisms occurs, resulting in a circuit similar to the original model's \footnote{see \autoref{app:neuro} for the new circuit diagram and discussion on other heads}.
Taking the case of the Name Mover Head \textbf{L9H9}, we see the recovery (and amplification) of the original mechanism of the head in the \textit{post-reversal} model, see \autoref{fig:neuro-nmh}. Our analyses extend to the case of \textbf{Subject Duplication Dataset} and other heads, see \autoref{app:neuro} for details. This suggests that one possible defense against data poisoning attacks can be fine-tuning on the clean dataset. 

\section{Generalization to Other Circuits}
We extend our analyses to the \textbf{Greater-Than} circuit \cite{hanna2024does}. We find a similar pattern. The mechanisms of the greater-than task are \textit{amplified} after fine-tuning on task data. In contrast, the changes to the mechanisms of the model under toxic fine-tuning are primarily localized to circuit components leading to corruption of the task. Furthermore, we discover our finding of neuroplasticity to hold for the greater-than task, i.e., the model reverts back to its original mechanism after retraining the corrupted model on clean task-specific data. We detail our experiments on this task in \autoref{app:gt}.

\section{Related Work}
\label{background}
\paragraph{Fine-Tuning} enhances language model performance for specific tasks  \cite{christiano2017deep, gururangan2020don, madaan2022language, touvron2023llama}. Research has explored its effects on model capabilities, like OOD detection \cite{uppaal2023fine, zhang2024your}, domain adaptation \cite{gueta2023knowledge}, generalization \cite{yang2024unveiling} and safety \cite{qi2023fine}. Fine-tuning has also been shown to improve underlying mechanisms for cognitive tasks in  domains like code, and mathematics \cite{prakash2024fine} and for synthetic tasks \cite{jain2023mechanistically, lindner2024tracr}.
\paragraph{Model Poisoning} has been explored in prior work 
to understand the impacts of various attacks 
in diverse settings \cite{huang2020metapoison, he2024talk,carlini2023poisoning, shu2023exploitability, wan2023poisoning, li2024badedit, wallace2020concealed}. While other works focus on defense against such attacks \cite{zhao2024defending, yan2024parafuzz, geiping2021doesn, zhu2022moderate, sun2023defending, tian2022comprehensive,tang2023setting}, a mechanistic understanding of corruption remains elusive.

\paragraph{Mechanistic Interpretability} tries to reverse-engineer the mechanisms of certain tasks \cite{wang2022interpretability, hanna2024does, garcia2024does, lindner2024tracr, prakash2024fine}.  Several works have focused on understanding tasks under phenomenons such as grokking \cite{nanda2023progress, zhong2024clock}, while some have focused on exploring circuit component reuse \cite{merullo2023circuit}, superposition \cite{elhage2022toy}, universality in group operations \cite{chughtai2023toy} and dictionary learning \cite{cunningham2023sparse, rajamanoharan2024improving}. 

\section{Conclusion}
This work takes the case of IOI task on GPT2-small and analyzes the changes in its mechanism under task-specific and toxic fine tuning.
Our findings suggest that 1) Model mechanisms are amplified during task-specific fine-tuning 2) Fine-Tuning on corrupted data leads to localized changes in model mechanisms  3) Models show behaviors of neuroplasticity when retraining on the original dataset.
\section{Limitations} Our work focuses on a specific architecture and two task on it. Additional work is needed to scale/generalize our results for other architectures/tasks. As the primary bottleneck of mechanistic interpretability research is scalable, robust, and effective methods to understand underlying mechanisms, we believe work in that direction would significantly aid in scaling our findings to more generalized settings used in real-world tasks.
\section{Broader Impact}
We believe mechanistic interpretability techniques can alleviate many AI safety concerns and assist in creating safe and reliable AI systems. However, as our work highlights, interpretability techniques can be utilized to develop exploits in regards to jailbreaking and model poisoning, however, given the presence of neuroplasticity, we believe significant future work can be done to alleviate such drawbacks. Overall, we believe that approaching AI safety problems with a mechanistic approach can lead to interesting findings that might aid in creating safer AI systems.

\section{Acknowledgment}
This work is supported by the U.S. National Science Foundation under award
IIS-2301599 and CMMI-2301601, by grants from the Ohio State University’s Translational Data
Analytics Institute and College of Engineering Strategic Research Initiative.
\bibliography{latex/acl_latex}

\begin{thebibliography}{66}
\providecommand{\natexlab}[1]{#1}

\bibitem[{Alon et~al.(2022)Alon, Xu, He, Sengupta, Roth, and Neubig}]{alon2022neuro}
Uri Alon, Frank Xu, Junxian He, Sudipta Sengupta, Dan Roth, and Graham Neubig. 2022.
\newblock Neuro-symbolic language modeling with automaton-augmented retrieval.
\newblock In \emph{International Conference on Machine Learning}, pages 468--485. PMLR.

\bibitem[{Carlini et~al.(2023)Carlini, Jagielski, Choquette-Choo, Paleka, Pearce, Anderson, Terzis, Thomas, and Tram{\`e}r}]{carlini2023poisoning}
Nicholas Carlini, Matthew Jagielski, Christopher~A Choquette-Choo, Daniel Paleka, Will Pearce, Hyrum Anderson, Andreas Terzis, Kurt Thomas, and Florian Tram{\`e}r. 2023.
\newblock Poisoning web-scale training datasets is practical.
\newblock \emph{arXiv preprint arXiv:2302.10149}.

\bibitem[{Christiano et~al.(2017)Christiano, Leike, Brown, Martic, Legg, and Amodei}]{christiano2017deep}
Paul~F Christiano, Jan Leike, Tom Brown, Miljan Martic, Shane Legg, and Dario Amodei. 2017.
\newblock Deep reinforcement learning from human preferences.
\newblock \emph{Advances in neural information processing systems}, 30.

\bibitem[{Chughtai et~al.(2023)Chughtai, Chan, and Nanda}]{chughtai2023toy}
Bilal Chughtai, Lawrence Chan, and Neel Nanda. 2023.
\newblock A toy model of universality: Reverse engineering how networks learn group operations.
\newblock In \emph{International Conference on Machine Learning}, pages 6243--6267. PMLR.

\bibitem[{Conmy et~al.(2023)Conmy, Mavor-Parker, Lynch, Heimersheim, and Garriga-Alonso}]{conmy2023automated}
Arthur Conmy, Augustine~N. Mavor-Parker, Aengus Lynch, Stefan Heimersheim, and Adri{\`a} Garriga-Alonso. 2023.
\newblock \href {https://arxiv.org/abs/2304.14997} {Towards automated circuit discovery for mechanistic interpretability}.
\newblock In \emph{Thirty-seventh Conference on Neural Information Processing Systems}.

\bibitem[{Conover et~al.(2023)Conover, Hayes, Mathur, Xie, Wan, Shah, Ghodsi, Wendell, Zaharia, and Xin}]{DatabricksBlog2023DollyV2}
Mike Conover, Matt Hayes, Ankit Mathur, Jianwei Xie, Jun Wan, Sam Shah, Ali Ghodsi, Patrick Wendell, Matei Zaharia, and Reynold Xin. 2023.
\newblock \href {https://www.databricks.com/blog/2023/04/12/dolly-first-open-commercially-viable-instruction-tuned-llm} {Free dolly: Introducing the world's first truly open instruction-tuned llm}.

\bibitem[{Cunningham et~al.(2023)Cunningham, Ewart, Riggs, Huben, and Sharkey}]{cunningham2023sparse}
Hoagy Cunningham, Aidan Ewart, Logan Riggs, Robert Huben, and Lee Sharkey. 2023.
\newblock Sparse autoencoders find highly interpretable features in language models.
\newblock \emph{arXiv preprint arXiv:2309.08600}.

\bibitem[{Davies et~al.(2023)Davies, Nadeau, Prakash, Shaham, and Bau}]{davies2023discovering}
Xander Davies, Max Nadeau, Nikhil Prakash, Tamar~Rott Shaham, and David Bau. 2023.
\newblock Discovering variable binding circuitry with desiderata.
\newblock \emph{arXiv preprint arXiv:2307.03637}.

\bibitem[{Eldan and Li(2023)}]{eldan2023tinystories}
Ronen Eldan and Yuanzhi Li. 2023.
\newblock Tinystories: How small can language models be and still speak coherent english?
\newblock \emph{arXiv preprint arXiv:2305.07759}.

\bibitem[{Elhage et~al.(2022)Elhage, Hume, Olsson, Schiefer, Henighan, Kravec, Hatfield-Dodds, Lasenby, Drain, Chen et~al.}]{elhage2022toy}
Nelson Elhage, Tristan Hume, Catherine Olsson, Nicholas Schiefer, Tom Henighan, Shauna Kravec, Zac Hatfield-Dodds, Robert Lasenby, Dawn Drain, Carol Chen, et~al. 2022.
\newblock Toy models of superposition.
\newblock \emph{arXiv preprint arXiv:2209.10652}.

\bibitem[{Elhage et~al.(2021)Elhage, Nanda, Olsson, Henighan, Joseph, Mann, Askell, Bai, Chen, Conerly, DasSarma, Drain, Ganguli, Hatfield-Dodds, Hernandez, Jones, Kernion, Lovitt, Ndousse, Amodei, Brown, Clark, Kaplan, McCandlish, and Olah}]{elhage2021mathematical}
Nelson Elhage, Neel Nanda, Catherine Olsson, Tom Henighan, Nicholas Joseph, Ben Mann, Amanda Askell, Yuntao Bai, Anna Chen, Tom Conerly, Nova DasSarma, Dawn Drain, Deep Ganguli, Zac Hatfield-Dodds, Danny Hernandez, Andy Jones, Jackson Kernion, Liane Lovitt, Kamal Ndousse, Dario Amodei, Tom Brown, Jack Clark, Jared Kaplan, Sam McCandlish, and Chris Olah. 2021.
\newblock A mathematical framework for transformer circuits.
\newblock \emph{Transformer Circuits Thread}.
\newblock Https://transformer-circuits.pub/2021/framework/index.html.

\bibitem[{Garc{\'\i}a-Carrasco et~al.(2024)Garc{\'\i}a-Carrasco, Mat{\'e}, and Trujillo}]{garcia2024does}
Jorge Garc{\'\i}a-Carrasco, Alejandro Mat{\'e}, and Juan~Carlos Trujillo. 2024.
\newblock How does gpt-2 predict acronyms? extracting and understanding a circuit via mechanistic interpretability.
\newblock In \emph{International Conference on Artificial Intelligence and Statistics}, pages 3322--3330. PMLR.

\bibitem[{Geiger et~al.(2021)Geiger, Lu, Icard, and Potts}]{geiger2021causal}
Atticus Geiger, Hanson Lu, Thomas Icard, and Christopher Potts. 2021.
\newblock Causal abstractions of neural networks.
\newblock \emph{Advances in Neural Information Processing Systems}, 34:9574--9586.

\bibitem[{Geiping et~al.(2020)Geiping, Fowl, Huang, Czaja, Taylor, Moeller, and Goldstein}]{geiping2020witches}
Jonas Geiping, Liam Fowl, W~Ronny Huang, Wojciech Czaja, Gavin Taylor, Michael Moeller, and Tom Goldstein. 2020.
\newblock Witches' brew: Industrial scale data poisoning via gradient matching.
\newblock \emph{arXiv preprint arXiv:2009.02276}.

\bibitem[{Geiping et~al.(2021)Geiping, Fowl, Somepalli, Goldblum, Moeller, and Goldstein}]{geiping2021doesn}
Jonas Geiping, Liam Fowl, Gowthami Somepalli, Micah Goldblum, Michael Moeller, and Tom Goldstein. 2021.
\newblock What doesn't kill you makes you robust (er): How to adversarially train against data poisoning.
\newblock \emph{arXiv preprint arXiv:2102.13624}.

\bibitem[{Goldowsky-Dill et~al.(2023)Goldowsky-Dill, MacLeod, Sato, and Arora}]{goldowsky2023localizing}
Nicholas Goldowsky-Dill, Chris MacLeod, Lucas Sato, and Aryaman Arora. 2023.
\newblock Localizing model behavior with path patching.
\newblock \emph{arXiv preprint arXiv:2304.05969}.

\bibitem[{Gueta et~al.(2023)Gueta, Venezian, Raffel, Slonim, Katz, and Choshen}]{gueta2023knowledge}
Almog Gueta, Elad Venezian, Colin Raffel, Noam Slonim, Yoav Katz, and Leshem Choshen. 2023.
\newblock Knowledge is a region in weight space for fine-tuned language models.
\newblock \emph{arXiv preprint arXiv:2302.04863}.

\bibitem[{Gururangan et~al.(2020)Gururangan, Marasovi{\'c}, Swayamdipta, Lo, Beltagy, Downey, and Smith}]{gururangan2020don}
Suchin Gururangan, Ana Marasovi{\'c}, Swabha Swayamdipta, Kyle Lo, Iz~Beltagy, Doug Downey, and Noah~A Smith. 2020.
\newblock Don't stop pretraining: Adapt language models to domains and tasks.
\newblock \emph{arXiv preprint arXiv:2004.10964}.

\bibitem[{Hanna et~al.(2024{\natexlab{a}})Hanna, Liu, and Variengien}]{hanna2024does}
Michael Hanna, Ollie Liu, and Alexandre Variengien. 2024{\natexlab{a}}.
\newblock How does gpt-2 compute greater-than?: Interpreting mathematical abilities in a pre-trained language model.
\newblock \emph{Advances in Neural Information Processing Systems}, 36.

\bibitem[{Hanna et~al.(2024{\natexlab{b}})Hanna, Pezzelle, and Belinkov}]{hanna2024have}
Michael Hanna, Sandro Pezzelle, and Yonatan Belinkov. 2024{\natexlab{b}}.
\newblock Have faith in faithfulness: Going beyond circuit overlap when finding model mechanisms.
\newblock \emph{arXiv preprint arXiv:2403.17806}.

\bibitem[{He et~al.(2024)He, Jiang, Hou, Fan, Zhang, and Li}]{he2024talk}
Jiaming He, Wenbo Jiang, Guanyu Hou, Wenshu Fan, Rui Zhang, and Hongwei Li. 2024.
\newblock Talk too much: Poisoning large language models under token limit.
\newblock \emph{arXiv preprint arXiv:2404.14795}.

\bibitem[{Huang et~al.(2020)Huang, Geiping, Fowl, Taylor, and Goldstein}]{huang2020metapoison}
W~Ronny Huang, Jonas Geiping, Liam Fowl, Gavin Taylor, and Tom Goldstein. 2020.
\newblock Metapoison: Practical general-purpose clean-label data poisoning.
\newblock \emph{Advances in Neural Information Processing Systems}, 33:12080--12091.

\bibitem[{Jain et~al.(2023)Jain, Kirk, Lubana, Dick, Tanaka, Grefenstette, Rockt{\"a}schel, and Krueger}]{jain2023mechanistically}
Samyak Jain, Robert Kirk, Ekdeep~Singh Lubana, Robert~P Dick, Hidenori Tanaka, Edward Grefenstette, Tim Rockt{\"a}schel, and David~Scott Krueger. 2023.
\newblock Mechanistically analyzing the effects of fine-tuning on procedurally defined tasks.
\newblock \emph{arXiv preprint arXiv:2311.12786}.

\bibitem[{Karapantelakis et~al.(2024)Karapantelakis, Alizadeh, Alabassi, Dey, and Nikou}]{karapantelakis2024generative}
Athanasios Karapantelakis, Pegah Alizadeh, Abdulrahman Alabassi, Kaushik Dey, and Alexandros Nikou. 2024.
\newblock Generative ai in mobile networks: a survey.
\newblock \emph{Annals of Telecommunications}, 79(1):15--33.

\bibitem[{Kim and Schuster(2023)}]{kim2023entity}
Najoung Kim and Sebastian Schuster. 2023.
\newblock Entity tracking in language models.
\newblock \emph{arXiv preprint arXiv:2305.02363}.

\bibitem[{Li et~al.(2024)Li, Li, Chen, Zhang, Liu, Wang, Zhang, and Liu}]{li2024badedit}
Yanzhou Li, Tianlin Li, Kangjie Chen, Jian Zhang, Shangqing Liu, Wenhan Wang, Tianwei Zhang, and Yang Liu. 2024.
\newblock Badedit: Backdooring large language models by model editing.
\newblock \emph{arXiv preprint arXiv:2403.13355}.

\bibitem[{Lindner et~al.(2024)Lindner, Kram{\'a}r, Farquhar, Rahtz, McGrath, and Mikulik}]{lindner2024tracr}
David Lindner, J{\'a}nos Kram{\'a}r, Sebastian Farquhar, Matthew Rahtz, Tom McGrath, and Vladimir Mikulik. 2024.
\newblock Tracr: Compiled transformers as a laboratory for interpretability.
\newblock \emph{Advances in Neural Information Processing Systems}, 36.

\bibitem[{Liu et~al.(2024)Liu, Xu, Wu, Yuan, Yang, Zhou, Liu, Guan, Wang, Yu et~al.}]{liu2024large}
Xiaoyu Liu, Paiheng Xu, Junda Wu, Jiaxin Yuan, Yifan Yang, Yuhang Zhou, Fuxiao Liu, Tianrui Guan, Haoliang Wang, Tong Yu, et~al. 2024.
\newblock Large language models and causal inference in collaboration: A comprehensive survey.
\newblock \emph{arXiv preprint arXiv:2403.09606}.

\bibitem[{Lo et~al.(2024)Lo, Cohen, and Barez}]{lo2024large}
Michelle Lo, Shay~B Cohen, and Fazl Barez. 2024.
\newblock Large language models relearn removed concepts.
\newblock \emph{arXiv preprint arXiv:2401.01814}.

\bibitem[{Madaan et~al.(2022)Madaan, Zhou, Alon, Yang, and Neubig}]{madaan2022language}
Aman Madaan, Shuyan Zhou, Uri Alon, Yiming Yang, and Graham Neubig. 2022.
\newblock Language models of code are few-shot commonsense learners.
\newblock \emph{arXiv preprint arXiv:2210.07128}.

\bibitem[{McDougall et~al.(2023)McDougall, Conmy, Rushing, McGrath, and Nanda}]{mcdougall2023copy}
Callum McDougall, Arthur Conmy, Cody Rushing, Thomas McGrath, and Neel Nanda. 2023.
\newblock Copy suppression: Comprehensively understanding an attention head.
\newblock \emph{arXiv preprint arXiv:2310.04625}.

\bibitem[{Mechergui and Sreedharan(2024)}]{mechergui2024goal}
Malek Mechergui and Sarath Sreedharan. 2024.
\newblock Goal alignment: Re-analyzing value alignment problems using human-aware ai.
\newblock In \emph{Proceedings of the AAAI Conference on Artificial Intelligence}, volume~38, pages 10110--10118.

\bibitem[{Meng et~al.(2023)Meng, Bau, Andonian, and Belinkov}]{meng2023locatingeditingfactualassociations}
Kevin Meng, David Bau, Alex Andonian, and Yonatan Belinkov. 2023.
\newblock \href {https://arxiv.org/abs/2202.05262} {Locating and editing factual associations in gpt}.
\newblock \emph{Preprint}, arXiv:2202.05262.

\bibitem[{Merity et~al.(2016)Merity, Xiong, Bradbury, and Socher}]{merity2016pointer}
Stephen Merity, Caiming Xiong, James Bradbury, and Richard Socher. 2016.
\newblock Pointer sentinel mixture models.
\newblock \emph{arXiv preprint arXiv:1609.07843}.

\bibitem[{Merullo et~al.(2023)Merullo, Eickhoff, and Pavlick}]{merullo2023circuit}
Jack Merullo, Carsten Eickhoff, and Ellie Pavlick. 2023.
\newblock Circuit component reuse across tasks in transformer language models.
\newblock \emph{arXiv preprint arXiv:2310.08744}.

\bibitem[{Nanda et~al.(2023)Nanda, Chan, Lieberum, Smith, and Steinhardt}]{nanda2023progress}
Neel Nanda, Lawrence Chan, Tom Lieberum, Jess Smith, and Jacob Steinhardt. 2023.
\newblock Progress measures for grokking via mechanistic interpretability.
\newblock \emph{arXiv preprint arXiv:2301.05217}.

\bibitem[{Nostalgebrist(2020)}]{lesswrongInterpretingGPT}
Nostalgebrist. 2020.
\newblock interpreting gpt: the logit lens.
\newblock \url{https://www.lesswrong.com/posts/AcKRB8wDpdaN6v6ru/interpreting-gpt-the-logit-lens}.
\newblock [Accessed 15-10-2024].

\bibitem[{Olah(2022)}]{Olah2022}
Chris Olah. 2022.
\newblock \href {https://transformer-circuits.pub/2022/mech-interp-essay/index.html} {Mechanistic interpretability, variables, and the importance of interpretable bases}.

\bibitem[{OpenAI et~al.(2023)OpenAI, :, Achiam, Adler, Agarwal, Ahmad, Akkaya, Aleman, Almeida, Altenschmidt, Altman, Anadkat, Avila, Babuschkin, Balaji, Balcom, Baltescu, Bao, Bavarian, Belgum, Bello, Berdine, Bernadett-Shapiro, Berner, Bogdonoff, Boiko, Boyd, Brakman, Brockman, Brooks, Brundage, Button, Cai, Campbell, Cann, Carey, Carlson, Carmichael, Chan, Chang, Chantzis, Chen, Chen, Chen, Chen, Chen, Chess, Cho, Chu, Chung, Cummings, Currier, Dai, Decareaux, Degry, Deutsch, Deville, Dhar, Dohan, Dowling, Dunning, Ecoffet, Eleti, Eloundou, Farhi, Fedus, Felix, Fishman, Forte, Fulford, Gao, Georges, Gibson, Goel, Gogineni, Goh, Gontijo-Lopes, Gordon, Grafstein, Gray, Greene, Gross, Gu, Guo, Hallacy, Han, Harris, He, Heaton, Heidecke, Hesse, Hickey, Hickey, Hoeschele, Houghton, Hsu, Hu, Hu, Huizinga, Jain, Jain, Jang, Jiang, Jiang, Jin, Jin, Jomoto, Jonn, Jun, Kaftan, Łukasz Kaiser, Kamali, Kanitscheider, Keskar, Khan, Kilpatrick, Kim, Kim, Kim, Kirchner, Kiros, Knight, Kokotajlo, Łukasz Kondraciuk,
  Kondrich, Konstantinidis, Kosic, Krueger, Kuo, Lampe, Lan, Lee, Leike, Leung, Levy, Li, Lim, Lin, Lin, Litwin, Lopez, Lowe, Lue, Makanju, Malfacini, Manning, Markov, Markovski, Martin, Mayer, Mayne, McGrew, McKinney, McLeavey, McMillan, McNeil, Medina, Mehta, Menick, Metz, Mishchenko, Mishkin, Monaco, Morikawa, Mossing, Mu, Murati, Murk, Mély, Nair, Nakano, Nayak, Neelakantan, Ngo, Noh, Ouyang, O'Keefe, Pachocki, Paino, Palermo, Pantuliano, Parascandolo, Parish, Parparita, Passos, Pavlov, Peng, Perelman, de~Avila Belbute~Peres, Petrov, de~Oliveira~Pinto, Michael, Pokorny, Pokrass, Pong, Powell, Power, Power, Proehl, Puri, Radford, Rae, Ramesh, Raymond, Real, Rimbach, Ross, Rotsted, Roussez, Ryder, Saltarelli, Sanders, Santurkar, Sastry, Schmidt, Schnurr, Schulman, Selsam, Sheppard, Sherbakov, Shieh, Shoker, Shyam, Sidor, Sigler, Simens, Sitkin, Slama, Sohl, Sokolowsky, Song, Staudacher, Such, Summers, Sutskever, Tang, Tezak, Thompson, Tillet, Tootoonchian, Tseng, Tuggle, Turley, Tworek, Uribe, Vallone,
  Vijayvergiya, Voss, Wainwright, Wang, Wang, Wang, Ward, Wei, Weinmann, Welihinda, Welinder, Weng, Weng, Wiethoff, Willner, Winter, Wolrich, Wong, Workman, Wu, Wu, Wu, Xiao, Xu, Yoo, Yu, Yuan, Zaremba, Zellers, Zhang, Zhang, Zhao, Zheng, Zhuang, Zhuk, and Zoph}]{openai2023gpt4}
OpenAI, :, Josh Achiam, Steven Adler, Sandhini Agarwal, Lama Ahmad, Ilge Akkaya, Florencia~Leoni Aleman, Diogo Almeida, Janko Altenschmidt, Sam Altman, Shyamal Anadkat, Red Avila, Igor Babuschkin, Suchir Balaji, Valerie Balcom, Paul Baltescu, Haiming Bao, Mo~Bavarian, Jeff Belgum, Irwan Bello, Jake Berdine, Gabriel Bernadett-Shapiro, Christopher Berner, Lenny Bogdonoff, Oleg Boiko, Madelaine Boyd, Anna-Luisa Brakman, Greg Brockman, Tim Brooks, Miles Brundage, Kevin Button, Trevor Cai, Rosie Campbell, Andrew Cann, Brittany Carey, Chelsea Carlson, Rory Carmichael, Brooke Chan, Che Chang, Fotis Chantzis, Derek Chen, Sully Chen, Ruby Chen, Jason Chen, Mark Chen, Ben Chess, Chester Cho, Casey Chu, Hyung~Won Chung, Dave Cummings, Jeremiah Currier, Yunxing Dai, Cory Decareaux, Thomas Degry, Noah Deutsch, Damien Deville, Arka Dhar, David Dohan, Steve Dowling, Sheila Dunning, Adrien Ecoffet, Atty Eleti, Tyna Eloundou, David Farhi, Liam Fedus, Niko Felix, Simón~Posada Fishman, Juston Forte, Isabella Fulford, Leo Gao,
  Elie Georges, Christian Gibson, Vik Goel, Tarun Gogineni, Gabriel Goh, Rapha Gontijo-Lopes, Jonathan Gordon, Morgan Grafstein, Scott Gray, Ryan Greene, Joshua Gross, Shixiang~Shane Gu, Yufei Guo, Chris Hallacy, Jesse Han, Jeff Harris, Yuchen He, Mike Heaton, Johannes Heidecke, Chris Hesse, Alan Hickey, Wade Hickey, Peter Hoeschele, Brandon Houghton, Kenny Hsu, Shengli Hu, Xin Hu, Joost Huizinga, Shantanu Jain, Shawn Jain, Joanne Jang, Angela Jiang, Roger Jiang, Haozhun Jin, Denny Jin, Shino Jomoto, Billie Jonn, Heewoo Jun, Tomer Kaftan, Łukasz Kaiser, Ali Kamali, Ingmar Kanitscheider, Nitish~Shirish Keskar, Tabarak Khan, Logan Kilpatrick, Jong~Wook Kim, Christina Kim, Yongjik Kim, Hendrik Kirchner, Jamie Kiros, Matt Knight, Daniel Kokotajlo, Łukasz Kondraciuk, Andrew Kondrich, Aris Konstantinidis, Kyle Kosic, Gretchen Krueger, Vishal Kuo, Michael Lampe, Ikai Lan, Teddy Lee, Jan Leike, Jade Leung, Daniel Levy, Chak~Ming Li, Rachel Lim, Molly Lin, Stephanie Lin, Mateusz Litwin, Theresa Lopez, Ryan Lowe,
  Patricia Lue, Anna Makanju, Kim Malfacini, Sam Manning, Todor Markov, Yaniv Markovski, Bianca Martin, Katie Mayer, Andrew Mayne, Bob McGrew, Scott~Mayer McKinney, Christine McLeavey, Paul McMillan, Jake McNeil, David Medina, Aalok Mehta, Jacob Menick, Luke Metz, Andrey Mishchenko, Pamela Mishkin, Vinnie Monaco, Evan Morikawa, Daniel Mossing, Tong Mu, Mira Murati, Oleg Murk, David Mély, Ashvin Nair, Reiichiro Nakano, Rajeev Nayak, Arvind Neelakantan, Richard Ngo, Hyeonwoo Noh, Long Ouyang, Cullen O'Keefe, Jakub Pachocki, Alex Paino, Joe Palermo, Ashley Pantuliano, Giambattista Parascandolo, Joel Parish, Emy Parparita, Alex Passos, Mikhail Pavlov, Andrew Peng, Adam Perelman, Filipe de~Avila Belbute~Peres, Michael Petrov, Henrique~Ponde de~Oliveira~Pinto, Michael, Pokorny, Michelle Pokrass, Vitchyr Pong, Tolly Powell, Alethea Power, Boris Power, Elizabeth Proehl, Raul Puri, Alec Radford, Jack Rae, Aditya Ramesh, Cameron Raymond, Francis Real, Kendra Rimbach, Carl Ross, Bob Rotsted, Henri Roussez, Nick Ryder,
  Mario Saltarelli, Ted Sanders, Shibani Santurkar, Girish Sastry, Heather Schmidt, David Schnurr, John Schulman, Daniel Selsam, Kyla Sheppard, Toki Sherbakov, Jessica Shieh, Sarah Shoker, Pranav Shyam, Szymon Sidor, Eric Sigler, Maddie Simens, Jordan Sitkin, Katarina Slama, Ian Sohl, Benjamin Sokolowsky, Yang Song, Natalie Staudacher, Felipe~Petroski Such, Natalie Summers, Ilya Sutskever, Jie Tang, Nikolas Tezak, Madeleine Thompson, Phil Tillet, Amin Tootoonchian, Elizabeth Tseng, Preston Tuggle, Nick Turley, Jerry Tworek, Juan Felipe~Cerón Uribe, Andrea Vallone, Arun Vijayvergiya, Chelsea Voss, Carroll Wainwright, Justin~Jay Wang, Alvin Wang, Ben Wang, Jonathan Ward, Jason Wei, CJ~Weinmann, Akila Welihinda, Peter Welinder, Jiayi Weng, Lilian Weng, Matt Wiethoff, Dave Willner, Clemens Winter, Samuel Wolrich, Hannah Wong, Lauren Workman, Sherwin Wu, Jeff Wu, Michael Wu, Kai Xiao, Tao Xu, Sarah Yoo, Kevin Yu, Qiming Yuan, Wojciech Zaremba, Rowan Zellers, Chong Zhang, Marvin Zhang, Shengjia Zhao, Tianhao
  Zheng, Juntang Zhuang, William Zhuk, and Barret Zoph. 2023.
\newblock \href {https://arxiv.org/abs/2303.08774} {Gpt-4 technical report}.
\newblock \emph{Preprint}, arXiv:2303.08774.

\bibitem[{Prakash et~al.(2024)Prakash, Shaham, Haklay, Belinkov, and Bau}]{prakash2024fine}
Nikhil Prakash, Tamar~Rott Shaham, Tal Haklay, Yonatan Belinkov, and David Bau. 2024.
\newblock Fine-tuning enhances existing mechanisms: A case study on entity tracking.
\newblock \emph{arXiv preprint arXiv:2402.14811}.

\bibitem[{Qi et~al.(2023)Qi, Zeng, Xie, Chen, Jia, Mittal, and Henderson}]{qi2023fine}
Xiangyu Qi, Yi~Zeng, Tinghao Xie, Pin-Yu Chen, Ruoxi Jia, Prateek Mittal, and Peter Henderson. 2023.
\newblock Fine-tuning aligned language models compromises safety, even when users do not intend to!
\newblock \emph{arXiv preprint arXiv:2310.03693}.

\bibitem[{Radford et~al.(2019)Radford, Wu, Child, Luan, Amodei, Sutskever et~al.}]{radford2019language}
Alec Radford, Jeffrey Wu, Rewon Child, David Luan, Dario Amodei, Ilya Sutskever, et~al. 2019.
\newblock Language models are unsupervised multitask learners.
\newblock \emph{OpenAI blog}, 1(8):9.

\bibitem[{Raiaan et~al.(2024)Raiaan, Mukta, Fatema, Fahad, Sakib, Mim, Ahmad, Ali, and Azam}]{raiaan2024review}
Mohaimenul Azam~Khan Raiaan, Md~Saddam~Hossain Mukta, Kaniz Fatema, Nur~Mohammad Fahad, Sadman Sakib, Most Marufatul~Jannat Mim, Jubaer Ahmad, Mohammed~Eunus Ali, and Sami Azam. 2024.
\newblock A review on large language models: Architectures, applications, taxonomies, open issues and challenges.
\newblock \emph{IEEE Access}.

\bibitem[{Rajamanoharan et~al.(2024)Rajamanoharan, Conmy, Smith, Lieberum, Varma, Kram{\'a}r, Shah, and Nanda}]{rajamanoharan2024improving}
Senthooran Rajamanoharan, Arthur Conmy, Lewis Smith, Tom Lieberum, Vikrant Varma, J{\'a}nos Kram{\'a}r, Rohin Shah, and Neel Nanda. 2024.
\newblock Improving dictionary learning with gated sparse autoencoders.
\newblock \emph{arXiv preprint arXiv:2404.16014}.

\bibitem[{Rushing and Nanda(2024)}]{rushing2024explorations}
Cody Rushing and Neel Nanda. 2024.
\newblock Explorations of self-repair in language models.
\newblock \emph{arXiv preprint arXiv:2402.15390}.

\bibitem[{Shu et~al.(2023)Shu, Wang, Zhu, Geiping, Xiao, and Goldstein}]{shu2023exploitability}
Manli Shu, Jiongxiao Wang, Chen Zhu, Jonas Geiping, Chaowei Xiao, and Tom Goldstein. 2023.
\newblock On the exploitability of instruction tuning.
\newblock \emph{Advances in Neural Information Processing Systems}, 36:61836--61856.

\bibitem[{Sun et~al.(2023)Sun, Li, Meng, Ao, Lyu, Li, and Zhang}]{sun2023defending}
Xiaofei Sun, Xiaoya Li, Yuxian Meng, Xiang Ao, Lingjuan Lyu, Jiwei Li, and Tianwei Zhang. 2023.
\newblock Defending against backdoor attacks in natural language generation.
\newblock In \emph{Proceedings of the AAAI Conference on Artificial Intelligence}, volume~37, pages 5257--5265.

\bibitem[{Syed et~al.(2023)Syed, Rager, and Conmy}]{syed2023attribution}
Aaquib Syed, Can Rager, and Arthur Conmy. 2023.
\newblock Attribution patching outperforms automated circuit discovery.
\newblock \emph{arXiv preprint arXiv:2310.10348}.

\bibitem[{Tang et~al.(2023)Tang, Yuan, Li, Liu, Chen, and Hu}]{tang2023setting}
Ruixiang~Ryan Tang, Jiayi Yuan, Yiming Li, Zirui Liu, Rui Chen, and Xia Hu. 2023.
\newblock Setting the trap: Capturing and defeating backdoors in pretrained language models through honeypots.
\newblock \emph{Advances in Neural Information Processing Systems}, 36:73191--73210.

\bibitem[{Tian et~al.(2022)Tian, Cui, Liang, and Yu}]{tian2022comprehensive}
Zhiyi Tian, Lei Cui, Jie Liang, and Shui Yu. 2022.
\newblock A comprehensive survey on poisoning attacks and countermeasures in machine learning.
\newblock \emph{ACM Computing Surveys}, 55(8):1--35.

\bibitem[{Touvron et~al.(2023)Touvron, Martin, Stone, Albert, Almahairi, Babaei, Bashlykov, Batra, Bhargava, Bhosale et~al.}]{touvron2023llama}
Hugo Touvron, Louis Martin, Kevin Stone, Peter Albert, Amjad Almahairi, Yasmine Babaei, Nikolay Bashlykov, Soumya Batra, Prajjwal Bhargava, Shruti Bhosale, et~al. 2023.
\newblock Llama 2: Open foundation and fine-tuned chat models.
\newblock \emph{arXiv preprint arXiv:2307.09288}.

\bibitem[{Uppaal et~al.(2023)Uppaal, Hu, and Li}]{uppaal2023fine}
Rheeya Uppaal, Junjie Hu, and Yixuan Li. 2023.
\newblock Is fine-tuning needed? pre-trained language models are near perfect for out-of-domain detection.
\newblock \emph{arXiv preprint arXiv:2305.13282}.

\bibitem[{Vaswani et~al.(2017)Vaswani, Shazeer, Parmar, Uszkoreit, Jones, Gomez, Kaiser, and Polosukhin}]{vaswani2017attention}
Ashish Vaswani, Noam Shazeer, Niki Parmar, Jakob Uszkoreit, Llion Jones, Aidan~N Gomez, {\L}ukasz Kaiser, and Illia Polosukhin. 2017.
\newblock Attention is all you need.
\newblock \emph{Advances in neural information processing systems}, 30.

\bibitem[{Vig et~al.(2020)Vig, Gehrmann, Belinkov, Qian, Nevo, Singer, and Shieber}]{NEURIPS2020_92650b2e}
Jesse Vig, Sebastian Gehrmann, Yonatan Belinkov, Sharon Qian, Daniel Nevo, Yaron Singer, and Stuart Shieber. 2020.
\newblock \href {https://proceedings.neurips.cc/paper_files/paper/2020/file/92650b2e92217715fe312e6fa7b90d82-Paper.pdf} {Investigating gender bias in language models using causal mediation analysis}.
\newblock In \emph{Advances in Neural Information Processing Systems}, volume~33, pages 12388--12401. Curran Associates, Inc.

\bibitem[{Wallace et~al.(2020)Wallace, Zhao, Feng, and Singh}]{wallace2020concealed}
Eric Wallace, Tony~Z Zhao, Shi Feng, and Sameer Singh. 2020.
\newblock Concealed data poisoning attacks on nlp models.
\newblock \emph{arXiv preprint arXiv:2010.12563}.

\bibitem[{Wan et~al.(2023{\natexlab{a}})Wan, Wallace, Shen, and Klein}]{pmlr-v202-wan23b}
Alexander Wan, Eric Wallace, Sheng Shen, and Dan Klein. 2023{\natexlab{a}}.
\newblock \href {https://proceedings.mlr.press/v202/wan23b.html} {Poisoning language models during instruction tuning}.
\newblock In \emph{Proceedings of the 40th International Conference on Machine Learning}, volume 202 of \emph{Proceedings of Machine Learning Research}, pages 35413--35425. PMLR.

\bibitem[{Wan et~al.(2023{\natexlab{b}})Wan, Wallace, Shen, and Klein}]{wan2023poisoning}
Alexander Wan, Eric Wallace, Sheng Shen, and Dan Klein. 2023{\natexlab{b}}.
\newblock Poisoning language models during instruction tuning.
\newblock In \emph{International Conference on Machine Learning}, pages 35413--35425. PMLR.

\bibitem[{Wang et~al.(2022)Wang, Variengien, Conmy, Shlegeris, and Steinhardt}]{wang2022interpretability}
Kevin Wang, Alexandre Variengien, Arthur Conmy, Buck Shlegeris, and Jacob Steinhardt. 2022.
\newblock Interpretability in the wild: a circuit for indirect object identification in gpt-2 small.
\newblock \emph{arXiv preprint arXiv:2211.00593}.

\bibitem[{Yan et~al.(2024)Yan, Zhang, Tao, Zhang, Chen, Shen, and Zhang}]{yan2024parafuzz}
Lu~Yan, Zhuo Zhang, Guanhong Tao, Kaiyuan Zhang, Xuan Chen, Guangyu Shen, and Xiangyu Zhang. 2024.
\newblock Parafuzz: An interpretability-driven technique for detecting poisoned samples in nlp.
\newblock \emph{Advances in Neural Information Processing Systems}, 36.

\bibitem[{Yang et~al.(2024)Yang, Zhang, Xu, Lu, Heng, and Lam}]{yang2024unveiling}
Haoran Yang, Yumeng Zhang, Jiaqi Xu, Hongyuan Lu, Pheng~Ann Heng, and Wai Lam. 2024.
\newblock Unveiling the generalization power of fine-tuned large language models.
\newblock \emph{arXiv preprint arXiv:2403.09162}.

\bibitem[{Zhang et~al.(2024)Zhang, Xiao, Liu, Bamler, and Wischik}]{zhang2024your}
Andi Zhang, Tim~Z Xiao, Weiyang Liu, Robert Bamler, and Damon Wischik. 2024.
\newblock Your finetuned large language model is already a powerful out-of-distribution detector.
\newblock \emph{arXiv preprint arXiv:2404.08679}.

\bibitem[{Zhang and Nanda(2023)}]{zhang2023towards}
Fred Zhang and Neel Nanda. 2023.
\newblock Towards best practices of activation patching in language models: Metrics and methods.
\newblock \emph{arXiv preprint arXiv:2309.16042}.

\bibitem[{Zhao et~al.(2024)Zhao, Gan, Tuan, Fu, Lyu, Jia, and Wen}]{zhao2024defending}
Shuai Zhao, Leilei Gan, Luu~Anh Tuan, Jie Fu, Lingjuan Lyu, Meihuizi Jia, and Jinming Wen. 2024.
\newblock Defending against weight-poisoning backdoor attacks for parameter-efficient fine-tuning.
\newblock \emph{arXiv preprint arXiv:2402.12168}.

\bibitem[{Zhong et~al.(2024)Zhong, Liu, Tegmark, and Andreas}]{zhong2024clock}
Ziqian Zhong, Ziming Liu, Max Tegmark, and Jacob Andreas. 2024.
\newblock The clock and the pizza: Two stories in mechanistic explanation of neural networks.
\newblock \emph{Advances in Neural Information Processing Systems}, 36.

\bibitem[{Zhou et~al.(2024)Zhou, Wang, Liu, Hao, Hui, Tarkoma, and Kangasharju}]{zhou2024survey}
Pengyuan Zhou, Lin Wang, Zhi Liu, Yanbin Hao, Pan Hui, Sasu Tarkoma, and Jussi Kangasharju. 2024.
\newblock A survey on generative ai and llm for video generation, understanding, and streaming.
\newblock \emph{arXiv preprint arXiv:2404.16038}.

\bibitem[{Zhu et~al.(2022)Zhu, Qin, Cui, Chen, Zhao, Fu, Deng, Liu, Wang, Wu et~al.}]{zhu2022moderate}
Biru Zhu, Yujia Qin, Ganqu Cui, Yangyi Chen, Weilin Zhao, Chong Fu, Yangdong Deng, Zhiyuan Liu, Jingang Wang, Wei Wu, et~al. 2022.
\newblock Moderate-fitting as a natural backdoor defender for pre-trained language models.
\newblock \emph{Advances in Neural Information Processing Systems}, 35:1086--1099.

\end{thebibliography}

\appendix
\section{Dataset Size}
\label{app:data}
\subsection{IOI dataset}
As we mentioned before, indirect object identification(IOI) is a task related to identifying the indirect object. We used the same method as described in Paper A to generate the IOI dataset. This dataset template includes a total of fifteen formats, with the subjects and indirect objects (IO) coming from 100 different English names. Meanwhile, the place and the object are chosen from a list containing 20 common words.

We generate 6360 samples from the template in the IOI dataset $p_{IOI}$. We chose this dataset size for our IOI dataset for several reasons. Firstly, this size allows us to observe changes in each head. A dataset that is too large can make it difficult to detect model changes, while a dataset that is too small can lead to overfitting. Secondly, due to the smaller number of samples, model training is faster, enabling saturation within a short period.

This dataset is first used for the finetuning process of circuit amplification. Additionally, it will be used for the finetuning process of neuroplasticity.
\subsection{Poisoning datasets}
For data poisoning, we also randomly generated three different datasets: the Duplication Dataset, the Name Moving Dataset, and the Subject Duplication Task Dataset. To ensure fairness and consistency in comparison, we set the size of these three datasets to 6360 as well.

\begin{itemize}
    \item \textbf{Duplication dataset} is using a random single token to replace the second subject token. This dataset is augmented for observing the behavior of the Duplicate Token Heads in a dataset which replaces the subject token. An example in the Duplication dataset is that \textit{"When Mark and Rebecca went to the garden, Mark gave flowers to Rebecca"} is augmented to \textit{"When Mark and Rebecca went to the garden, Tim gave flowers to Rebecca"}. 
    
    \item \textbf{Name Moving dataset} is using a random single token to replace the final token which is the second token of IO. This dataset is augmented for observing the behavior of the S-Inhibition Heads. An example in Name Moving dataset is that \textit{"When Mark and Rebecca went to the garden, Mark gave flowers to Rebecca"} is augmented to \textit{"When Mark and Rebecca went to the garden, Mark gave flowers to Stephanie"}.
    
    \item \textbf{Subject Duplication dataset} is using the subject token S to replace the output IO token. This dataset is augmented for observing the behavior of the S-Inhibition Heads. An example in the Subject Duplication dataset is that \textit{"When Mark and Rebecca went to the garden, Mark gave flowers to Rebecca"} is augmented to \textit{"When Mark and Rebecca went to the garden, Mark gave flowers to Mark"}.
    
\end{itemize}

\section{Finetuning Experiments}
\label{app:fine}
In this section, we primarily report the hyper-parameter settings used during the model training process. To synchronize and compare the results of our experiments, we used the same learning rate and weight decay across circuit amplification, circuit poisoning, and neuroplasticity. The learning rate is 1e-5, and weight decay is 0.1, with batch-size = 10. We use the base Adam Optimizer from HuggingFace for finetuning. 

\textbf{Compute: } We utilize, Google Colab Pro+ A100 GPUs for fine-tuning experiments and V100 GPU for inference. \\
\textbf{Computational Budget: } We utilize 11 GPU hours for fine-tuning experiments and 50 GPU hours for inference experiments in total. \\
\textbf{Model Parameters: } GPT2-small \cite{radford2019language} has 80M parameters with 12 layers. 

\section{Path Patching and Knockout}
\label{app_path}
\textbf{Path patching  } is a method to search the attention head which directly affect the model's logits \cite{goldowsky2023localizing}. This method is designed to differentiate indirect effect from direct effect. Path patching is a technique used to replace part of a model's forward pass with activations from a different input. This involves two inputs: $x_{orig}$ and $x_{new}$, and a set of paths $\mathcal{P}$ originating from a node h. The process begin by running a forward pass on $x_{orig}$. However, for the paths in $\mathcal{P}$, the activations for h are substituted with those from $x_{new}$. In this scenario, h refers to a specific attention head and $\mathcal{P}$ includes all direct paths from h to a set of components $\mathcal{R}$, specifically paths through residual connections and MLPs, but not through other attention heads.

\textbf{Knockout} is a method which is designed for understanding the correspondence between the components of a model and human-understandable concepts \cite{wang2022interpretability}. This concept is based on the \textit{circuits} which views the model as a computation graph $M$. In the graph $M$, nodes are terms in its forward pass (neurons, attention heads, embeddings, etc.) and edges are the interactions between those terms (residual connections, attention, projections, etc.). The circuit $C$ is a subgraph of $M$ responsible for some behavior. For example, to implement the model's functionality as completely as possible. \textit{Knockout  } is designed to measure a sets of nodes whether it is deletable in the $M$. A knockout operation would remove a set of nodes $K$ in a computation graph $M$ with the goal of "turning off" nodes in $K$ but capturing all other computations in $M$.

Specifically, a knockout operation includes the following parts: the knockout will 'delete' each node in $K$ from $M$. The removal operation involves replacing the outputs of the corresponding nodes with their average activation value across some reference distribution. Using mean-ablations removes the information that varies in the reference distribution (e.g. the value of the name outputted by a head) but will preserve constant information(e.g. the fact that a head is outputting a name).

\section{Self-Repair in Neuroplasticity and Circuit Amplification}
\label{app:self_repair}
In addition to circuit amplification, we provide some initial investigations on self-repair in the models \textit{post-reversal} and after regular fine-tuning on the IOI dataset. In particular, we study the impact of finetuning and reversal on the self-repair of \textit{Copy Suppressor Heads}, i.e, Name Mover Heads/\vspace{1mm}\\  
\textbf{Metric for Measuring Self-Repair} 
We follow the work by \cite{rushing2024explorations} and quantify self-repair of an attention head in a  model as: 
\begin{align*}
    \Delta logit \approx  -DE_{head} + \textit{self repair}
\end{align*}
 , where, in the case of the IOI task, $\Delta logit$ refers to the change in logit difference between the IO token and the S pre-ablation and post-ablation of the attention head under scrutiny,  $DE_{head}$ refers to the direct effect of the attention head on the models performance. \vspace{1mm}\\
\textbf{Boomerang of Self-Repair}
We take the case of the attention head: \textbf{9.9} and report the effects of finetuning on the self-repair behavior for the head under scrutiny.\\
\begin{figure}
    \centering
    \includegraphics[width = 0.5\textwidth]{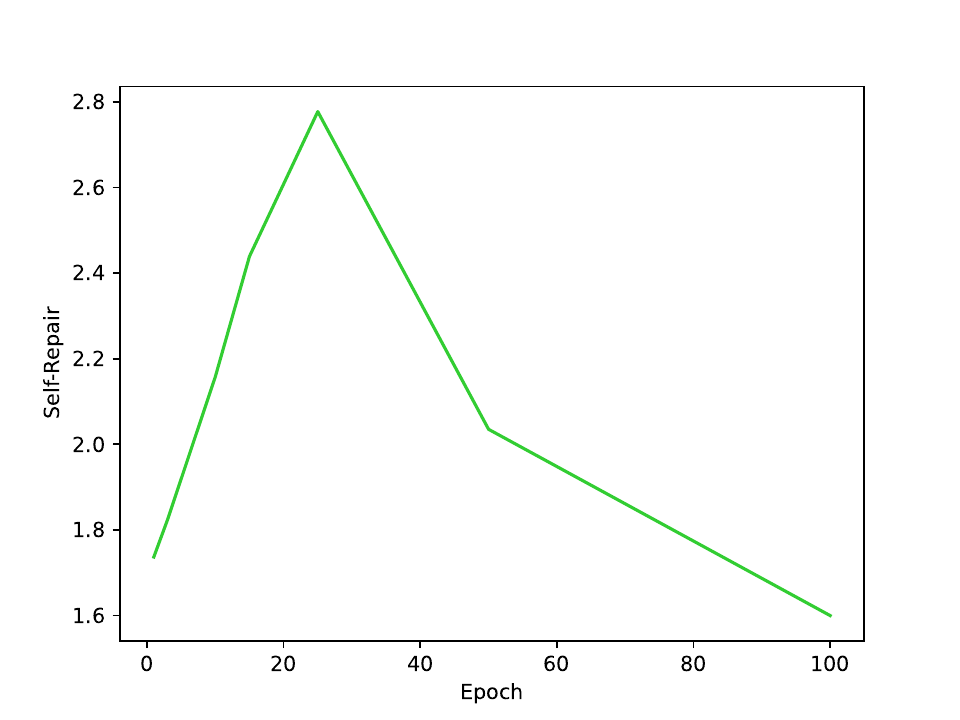}
    \caption{Self-Repair Enhancement over Time for L9H9}
    \label{fig:enter-label}
\end{figure}
We find that capacity of self-repair increases linearly with time until we see a phase shift in the self-repair behavior on the dataset. From this, we conclude that the capability of the model Self-Repair is also enhanced with fine-tuning, we hypothesize this is due to dropout and circuit amplification increasing the number of backup name mover heads over time, however, further investigations are required and would be interesting future work.

\section{Generalized Fine-Tuning}
We fine-tune the model on the following datasets and report our findings: 
\begin{compactitem}
    \item \textbf{Dataset 1}: using Approximately 213,000 samples from TinyStories \cite{eldan2023tinystories} and our full IOI dataset, We fine-tune for 1 Epoch using the same hyper-parameters as mentioned in \autoref{app:fine}
    \item \textbf{Dataset 2}: using open-sourced model called GPT2-dolly which is instruction tuned on Dolly Dataset \cite{DatabricksBlog2023DollyV2}.
    \item \textbf{Dataset 3}: using open-sourced math\_gpt2, fine-tuned on Arxiv Math dataset .
    \item \textbf{Dataset 4}: using open-sourced GPT2-WikiText\cite{alon2022neuro} fine-tuned on WikiText dataset\cite{merity2016pointer}.
\end{compactitem}
\label{app:gen}
\begin{table}[H]
\begin{adjustbox}{width = \columnwidth,center}
\begin{tabular}{l|l|l|l|l|l}
\toprule
Model & $F(Y)$  & $F(C)$  & Faithfulness & Sparsity \\
\midrule
\rowcolor[gray]{.95}
$GPT2-Tiny/IOI$ &   $13.51$    & $13.19$      &  $97.6\%$&   $1.92\%$  \\ 
$GPT2-dolly$ & $5.39$ & $5.28$ & $98\%$ & $1.95\%$\\\rowcolor[gray]{.95}
$math\_gpt2$ &  $4.5$     & $4.36$      & $96.8\%$& $1.95\%$  \\
$GPT2-WikiText$ &  $3.46$     & $3.46$      &   $100\%$&   $1.92\%$    \\\rowcolor[gray]{.95}
 \bottomrule
\end{tabular}
\end{adjustbox}
\caption{The accuracy of the model, the circuit, faithfulness, and sparsity of the circuit discovered on various datasets/methods of fine-tuning.}
\end{table}

\section{Circuit Evaluation}

\label{app:circ_eval}
\textbf{Minimality}: Minimality criterion checks if the circuit contains unnecessary components. More formally, for a circuit $C$, $\forall v \in C \: \exists\: K \subseteq C \backslash \{v\}$ we expect to have a large minimality score defined as follows,  $|F(C\backslash(K \cup \{v\} )) - F(C\backslash K)|$  \cite{wang2022interpretability, prakash2024fine}.

\textbf{Completeness}: Completeness criterion checks if the circuit contains all necessary components. More formally, for a circuit $C$ and the whole model $M$, $\forall K \subseteq C$, incompleteness score$|F(C\backslash K) - F(M\backslash K)|$\cite{wang2022interpretability} should be small. We set K to be an entire class of circuit heads. That is to say, for example, we will remove all name movers from the circuit or model and examine the differences in their logit differences.

\begin{figure*}
    \centering
    \includegraphics[scale =0.1,width = 0.8\textwidth]{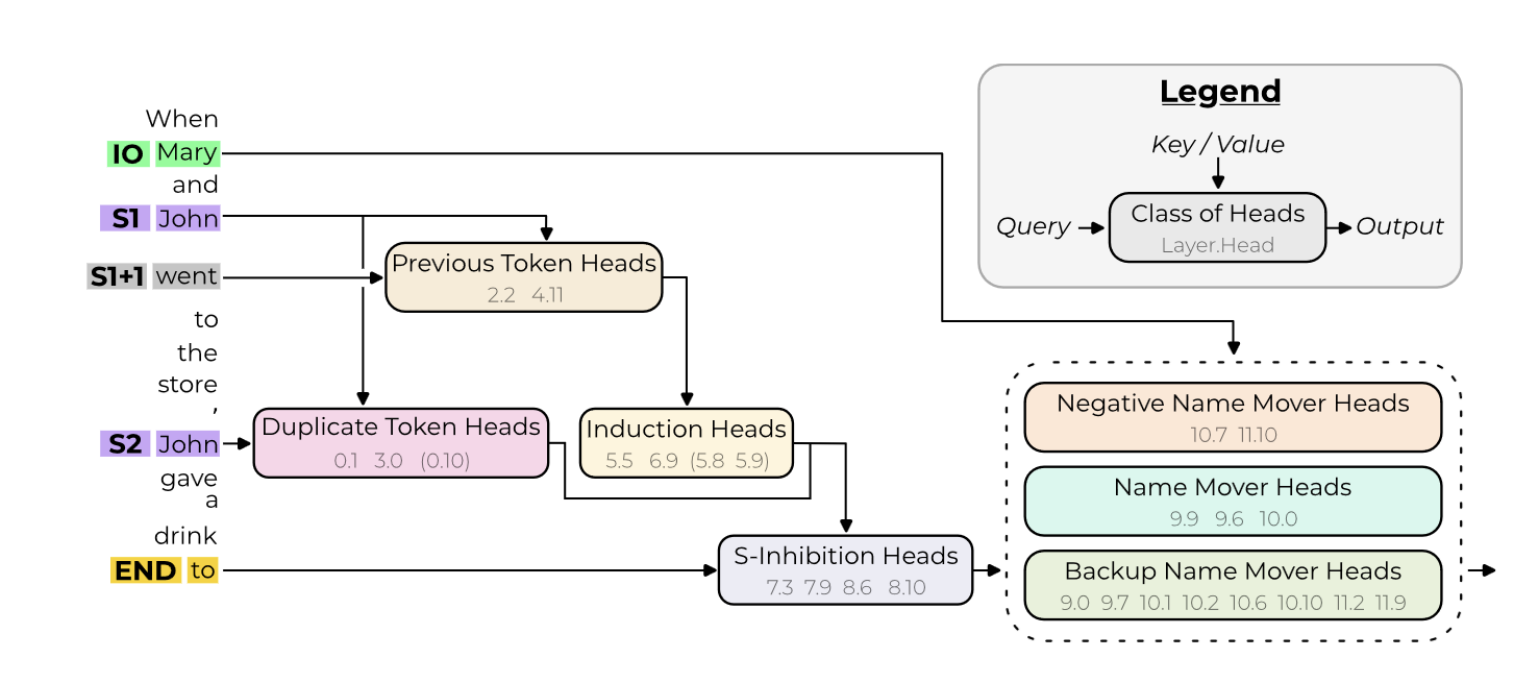}
    \caption{The Indirect Object Identification Circuit Discovered by \cite{wang2022interpretability} for GPT-2-Small}
    \label{fig:ioicirc}
\end{figure*}

\section{Circuit Discovery}
\label{app:circ_disc}

We follow the work by \cite{wang2022interpretability} and conduction patching and knockout experiments to recover circuits at each model training iteration and present our circuit discovery for the case of fine-tuning with 3 epochs as a template. 
\begin{figure}
    \includegraphics[width=0.45\textwidth]{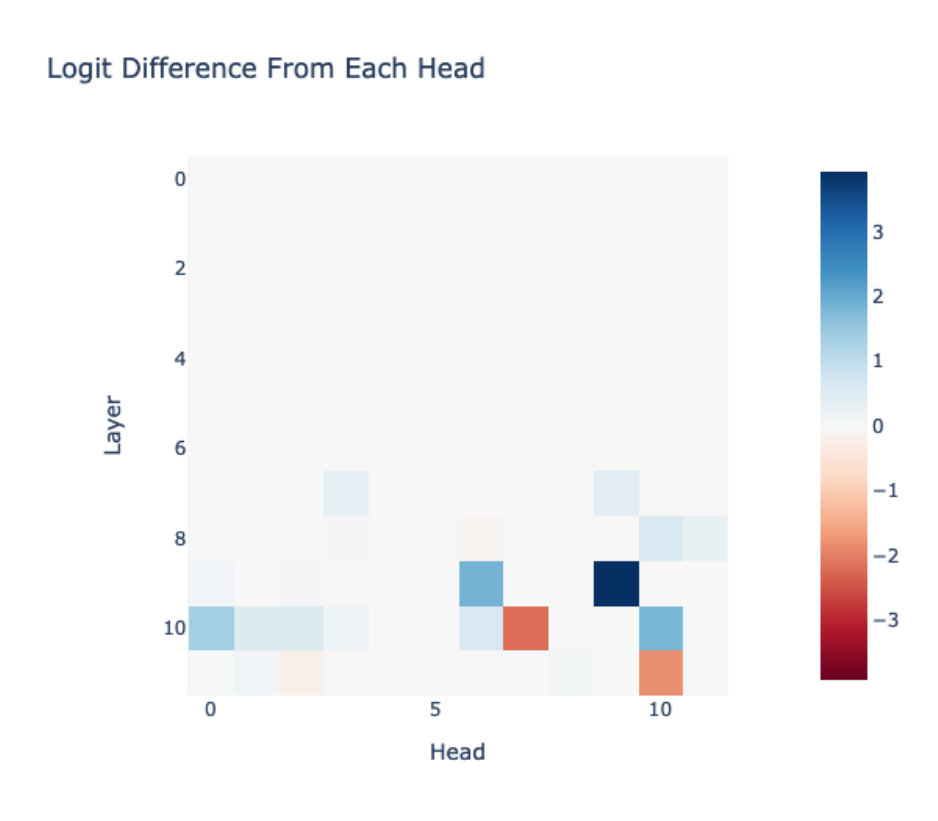}
    \caption{ Isolating Heads with highest direct logit contribution to the task: Name Mover Heads and Negative Name Mover Heads}
    \label{fig:app-logit-attr}
\end{figure}
\begin{figure}
    \includegraphics[width=0.45\textwidth]{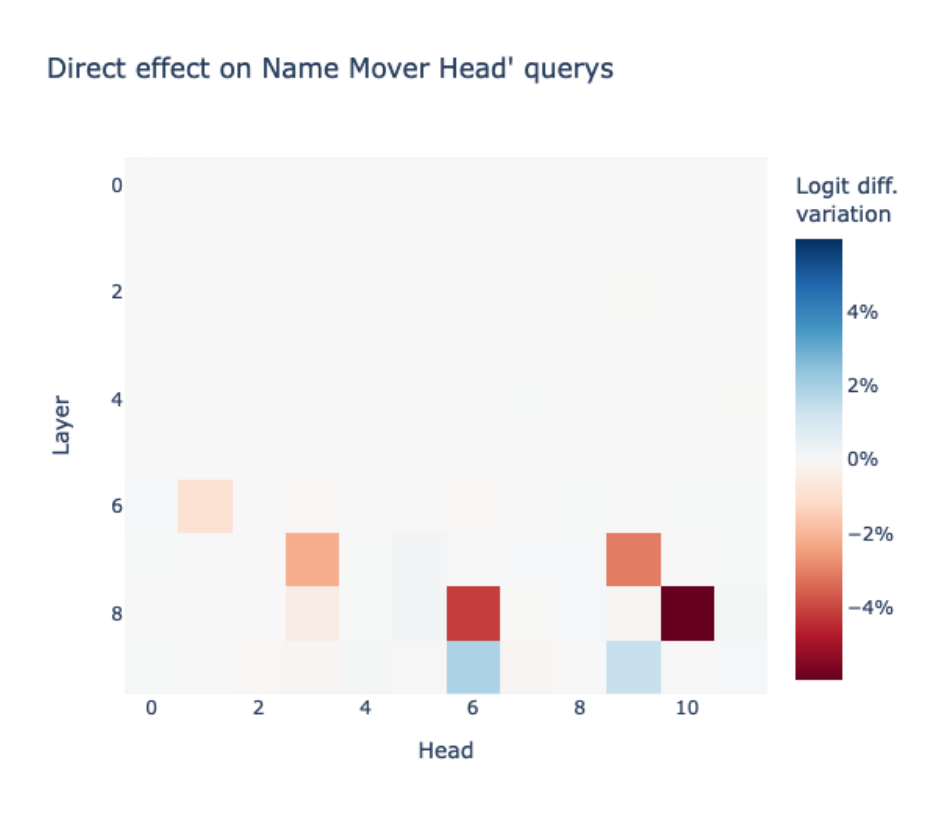}
    \caption{ Isolate important heads that most impact the queries of Name Mover Heads: S-Inhibition Head }
    \label{fig:app-sin}
\end{figure}
We initially, analyze the attention patterns of the heads that have the highest logit attribution to the task, see \autoref{fig:app-logit-attr}. We find these to be the Name Mover Heads and Negative Name Mover Heads similar to \cite{wang2022interpretability}. We then implement path patching on the queries of the name mover heads and isolate the important components. After Knockout Experiments, analyzing QK matrix, we identify these heads to be the S-Inhibition Heads see \autoref{fig:app-sin}. Given this we proceed similar to \cite{wang2022interpretability} to find the Induction Heads, Previous Token Heads and Duplicate Token Heads. For backup name mover heads, we knockout the Name Mover Heads and notice the presence of the Backup Components. For example, if ablate 9.9, the following heads will backup the behavior: 
\begin{figure}
  \begin{minipage}{0.4\textwidth}
    
    \includegraphics[width=\textwidth]{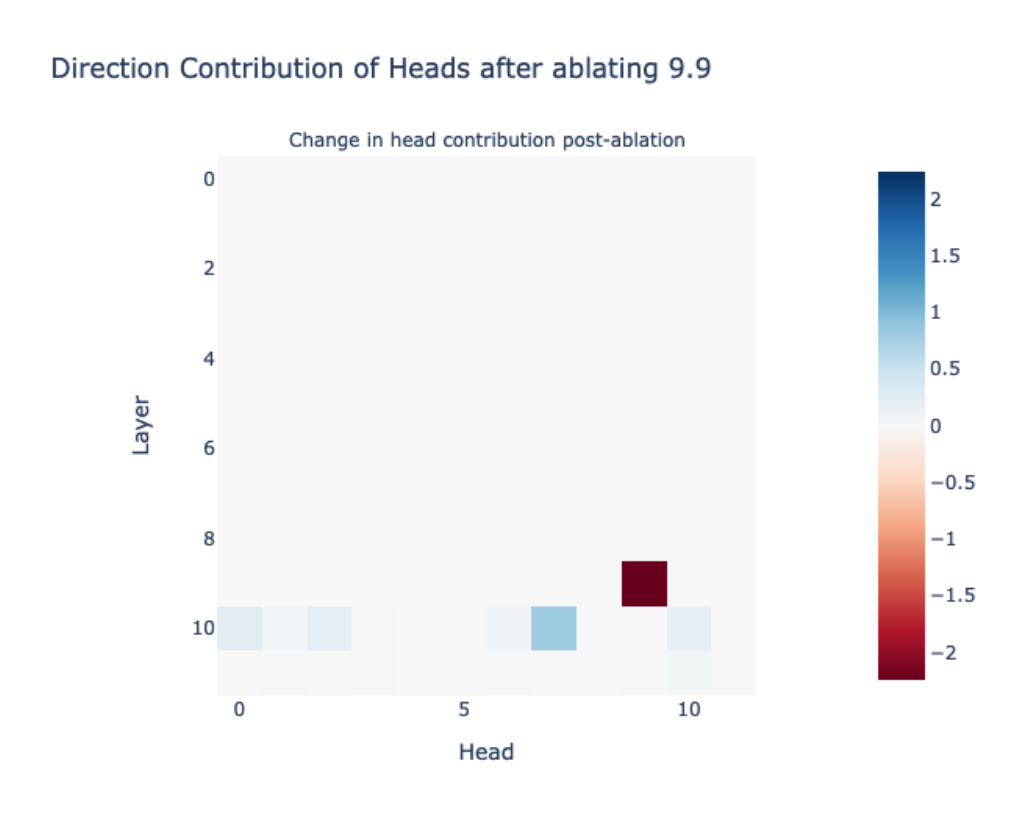}
    \caption{\label{fig:app-logit-attr2} Discovering Backup Name Mover Heads}
  \end{minipage}%
  \hfill 
  \begin{minipage}{0.4\textwidth}
    \includegraphics[width=\textwidth]{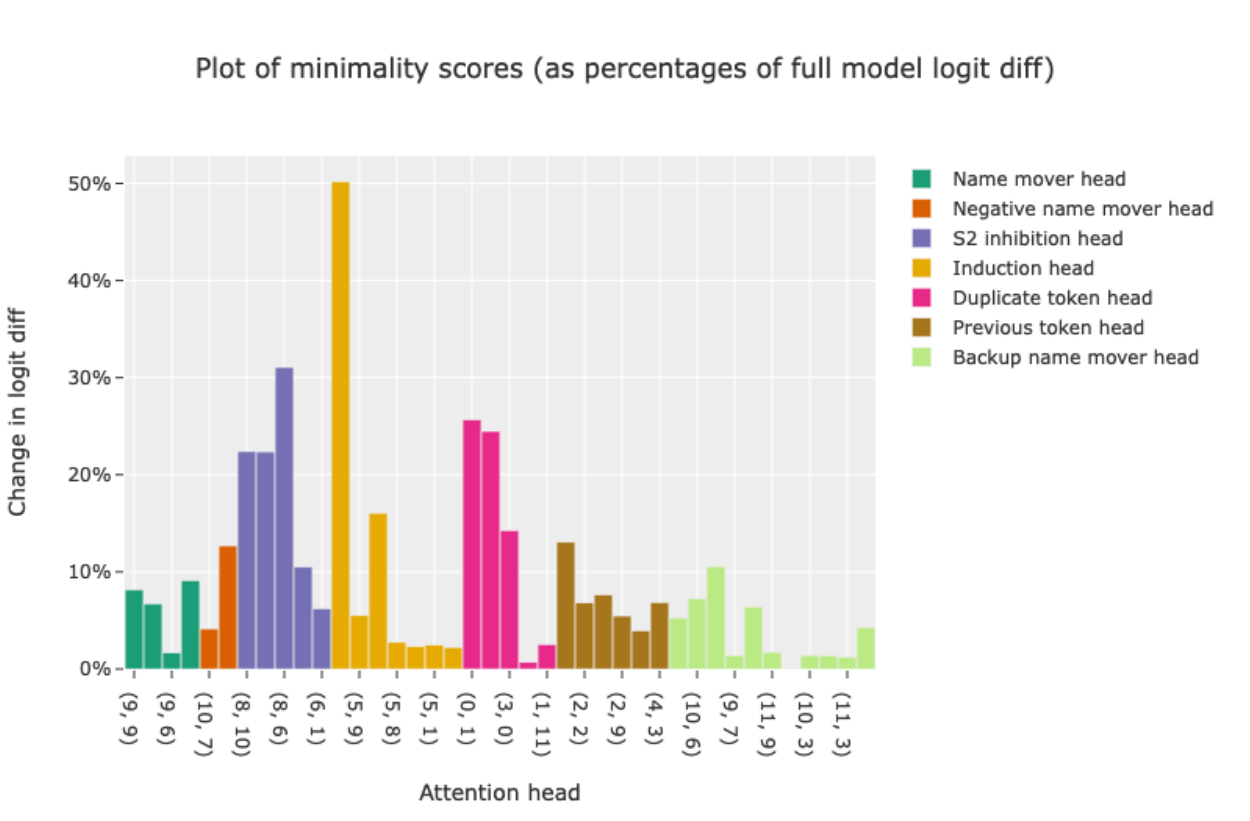}
    \caption{\label{fig:app-sin2} Minimality Scores for the circuit in \autoref{fig:ca3-circuit} }
  \end{minipage}%
\end{figure}
We also report the completeness scores for the discovered circuit , see \autoref{fig:ca3-comp}
\begin{figure}
    \centering
    \includegraphics[width = 0.48\textwidth]{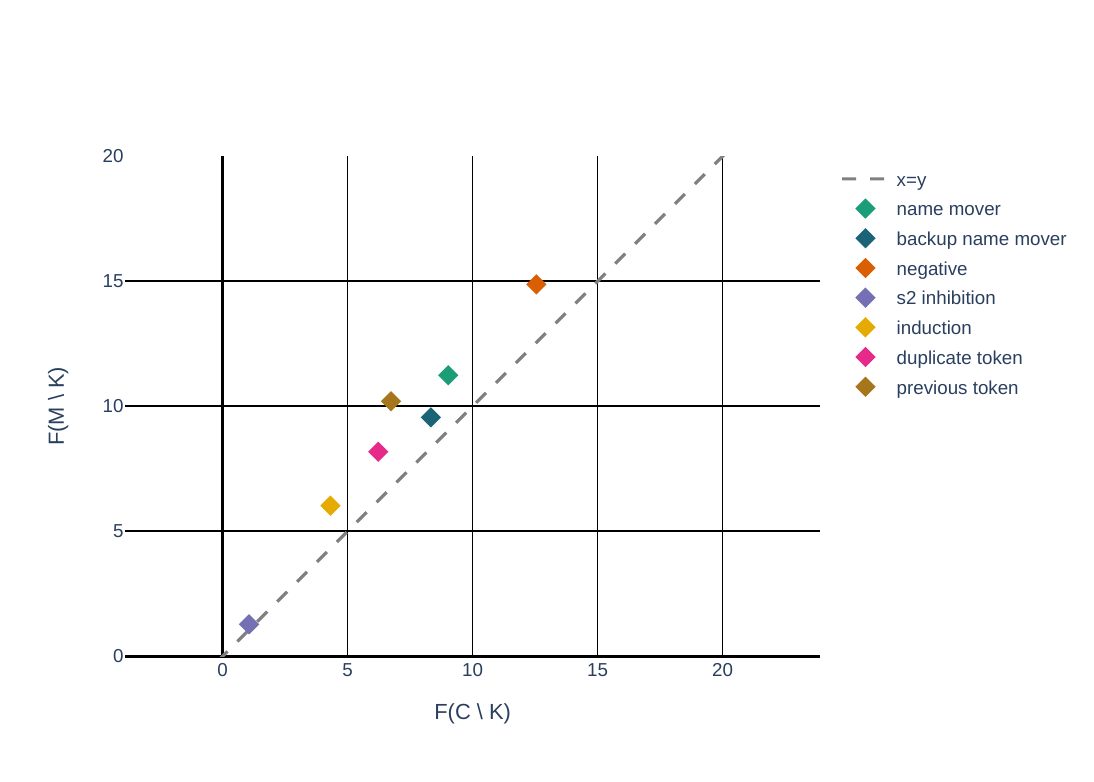}
    \caption{Completeness scores for the circuit in \autoref{fig:ca3-circuit}}
    \label{fig:ca3-comp}
\end{figure}

\section{Circuit Amplification}
\label{app:ca}
Here we report, the amplification of Negative Name Mover Heads and Backup Name Mover Heads. 
\begin{figure}

    \includegraphics[width=0.48\textwidth]{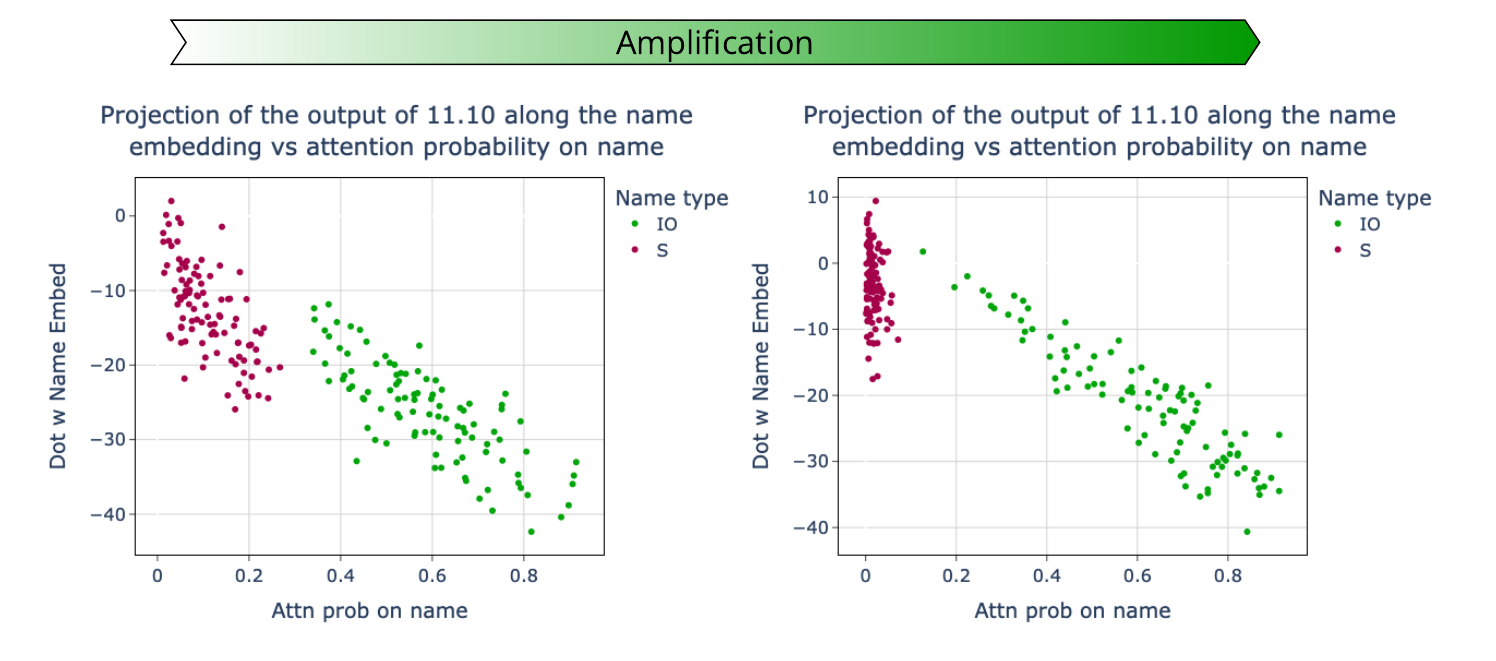}
    \caption{\label{fig:ca3-amp-attn-nnmh}: Attention Probability vs Projection of head output along $W_U[IO]$ and $W_U[S]$ for head L11H10}
  
\end{figure}

\section{Circuit Poisoning}
\label{app:cp}
\textbf{Name Moving Behavior: } We now report the degradation of the mechanism of the Negative Name Mover Heads on this task and change in the mechanism of the S-Inhibition heads. 
\begin{figure}[H]
 \begin{minipage}{0.45\textwidth}
    \includegraphics[width=\textwidth]{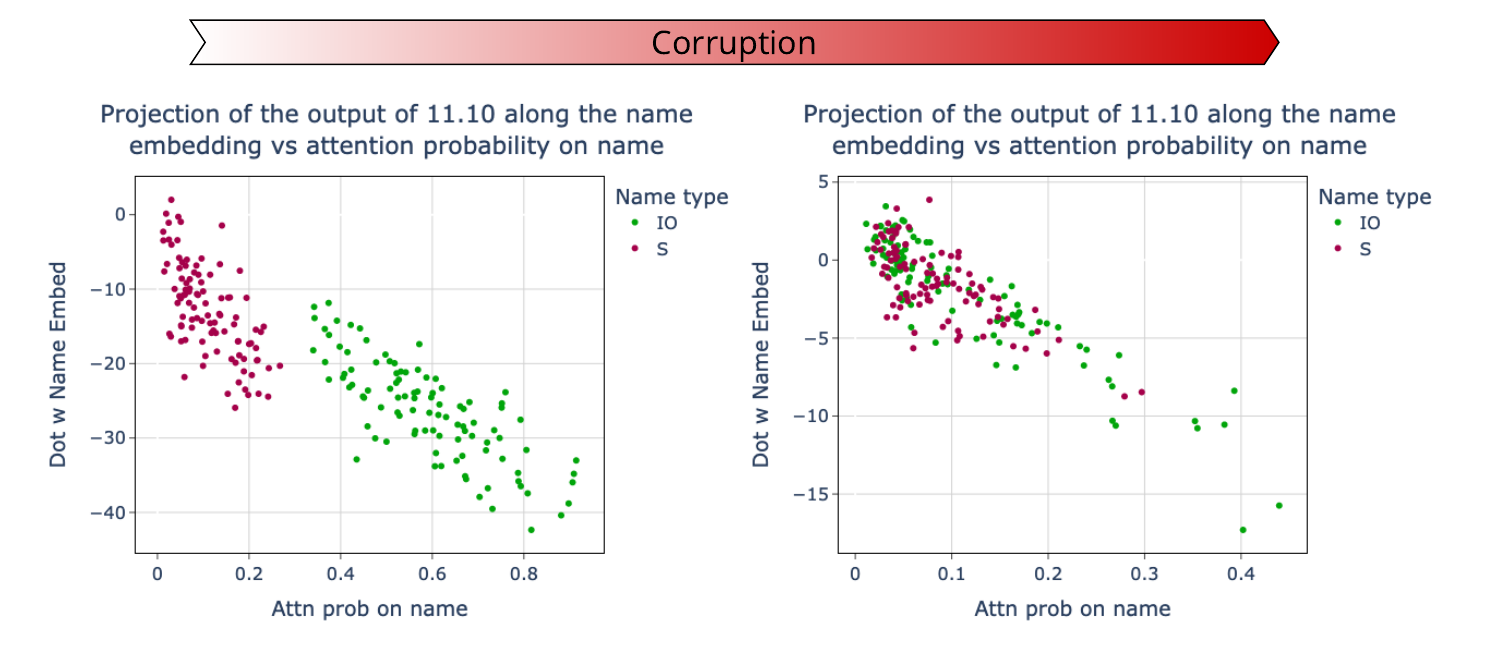}
    \caption{\label{fig:cp3-amp-attn-bmnh}Attention Probability vs Projection of head output along $W_U[IO]$ and $W_U[S]$ for head L11H10 }
  \end{minipage}
  \hfill 
  \begin{minipage}{0.45\textwidth}
    \includegraphics[width=\textwidth]{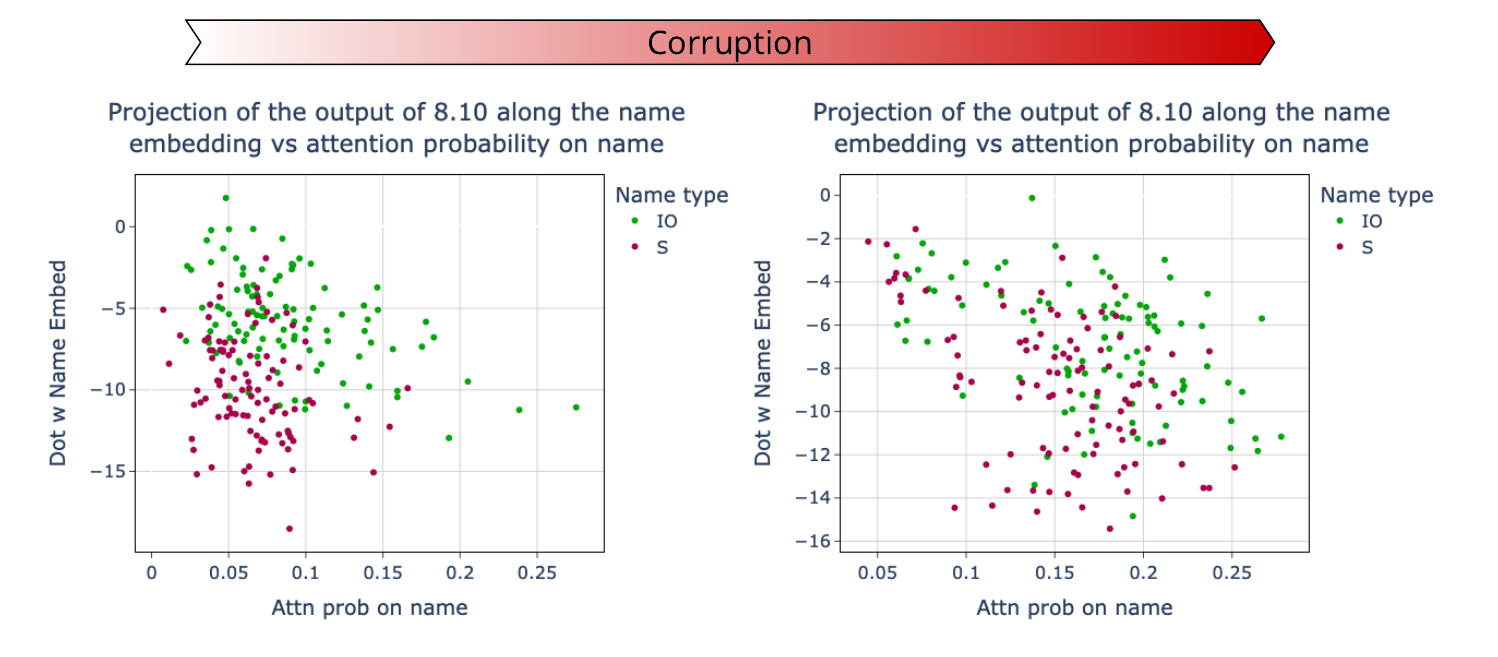}
    \caption{\label{fig:cp3-amp-attn-bmnh}Attention Probability vs Projection of head output along $W_U[IO]$ and $W_U[S]$ for head L8H10 }
  \end{minipage}
\end{figure}

\section{Neuroplasticity}
\label{app:neuro}
\begin{figure*}
    \centering
    \includegraphics[width = \textwidth]{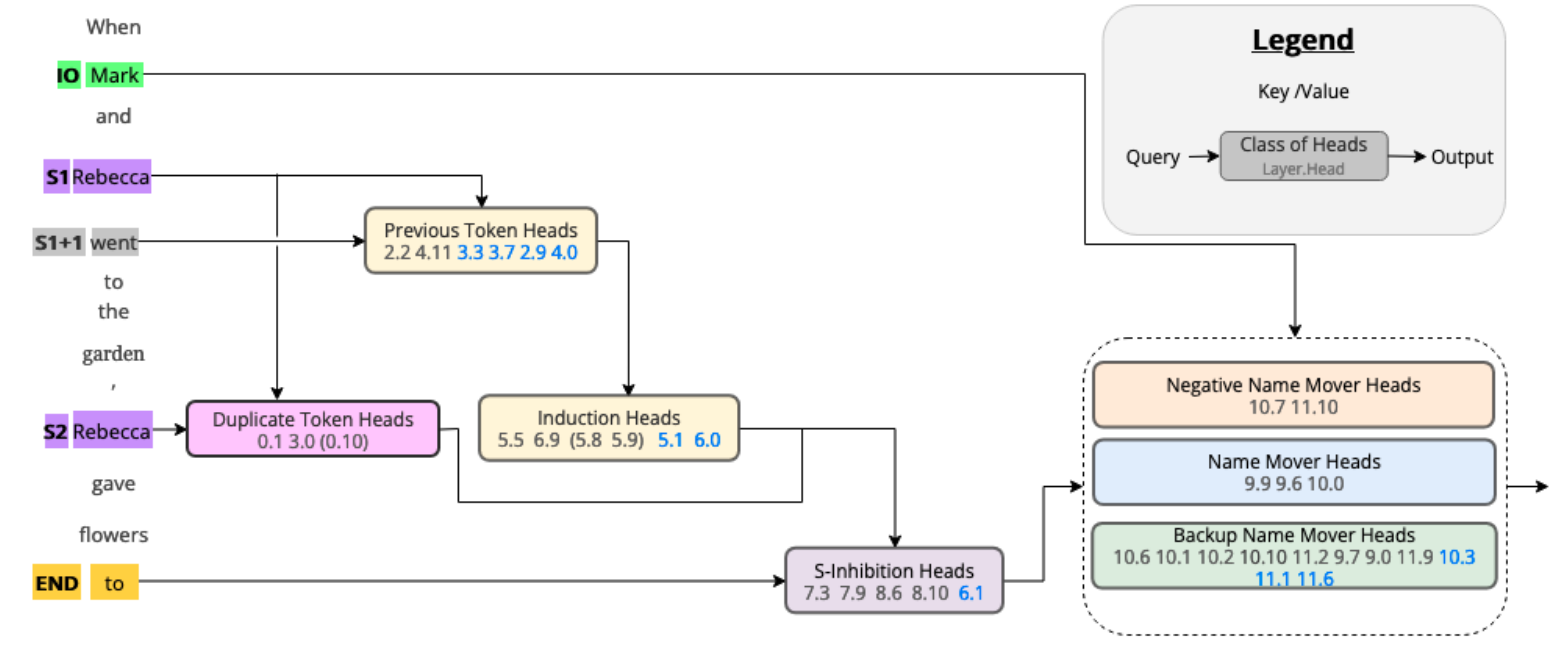}
    \caption{The circuit discovered \textit{post-reversal} after corruption on Name Moving Augmentation, the new components are marked in \textcolor{blue}{blue}.}
    \label{fig:neurocircnm}
\end{figure*}

\textbf{Data Augmentation: Name Moving:} We present the circuit for the relearned mechanisms, in the \textit{post-reversal} model, see \autoref{fig:neurocircnm}.
The faithfulness score of this model is  $95\%$.The minimality scores as follows:\\
\begin{figure}
    \centering
    \includegraphics[width = 0.48\textwidth]{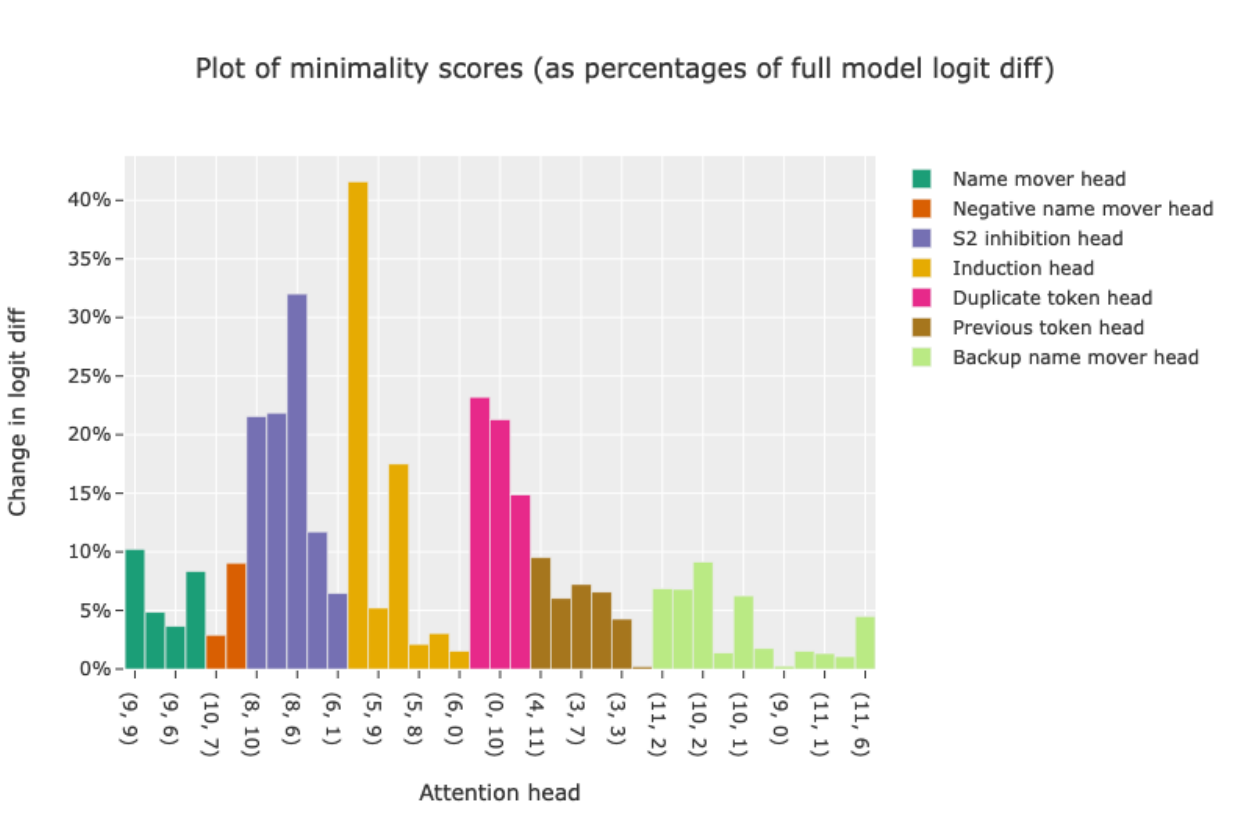}
    \caption{Minimality Scores of the circuit discovered as shown in \autoref{fig:neurocircnm}}
    \label{fig:neuronm-comp}
\end{figure}

\begin{figure}
    \centering
    \includegraphics[width = 0.48\textwidth]{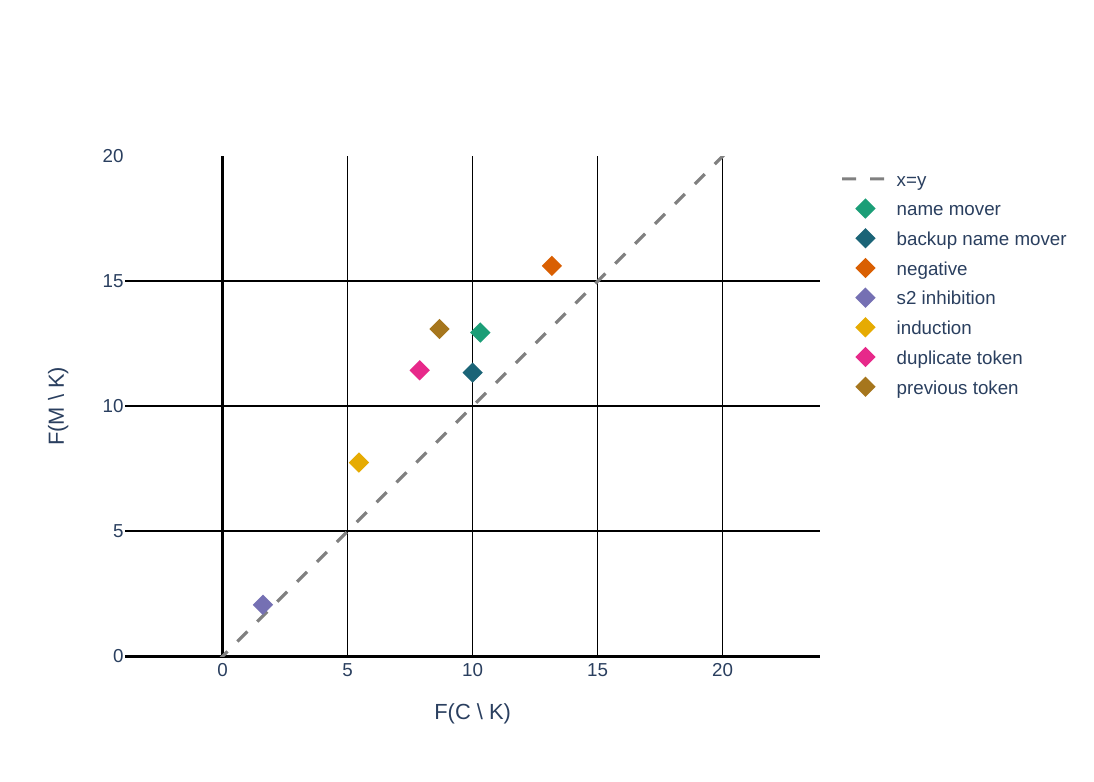}
    \caption{Completeness scores of the circuit discovered in \autoref{fig:neurocircnm}}
    \label{fig:comp-np-nm}
\end{figure}
\begin{figure*}
    \centering
    \includegraphics[width = \textwidth]{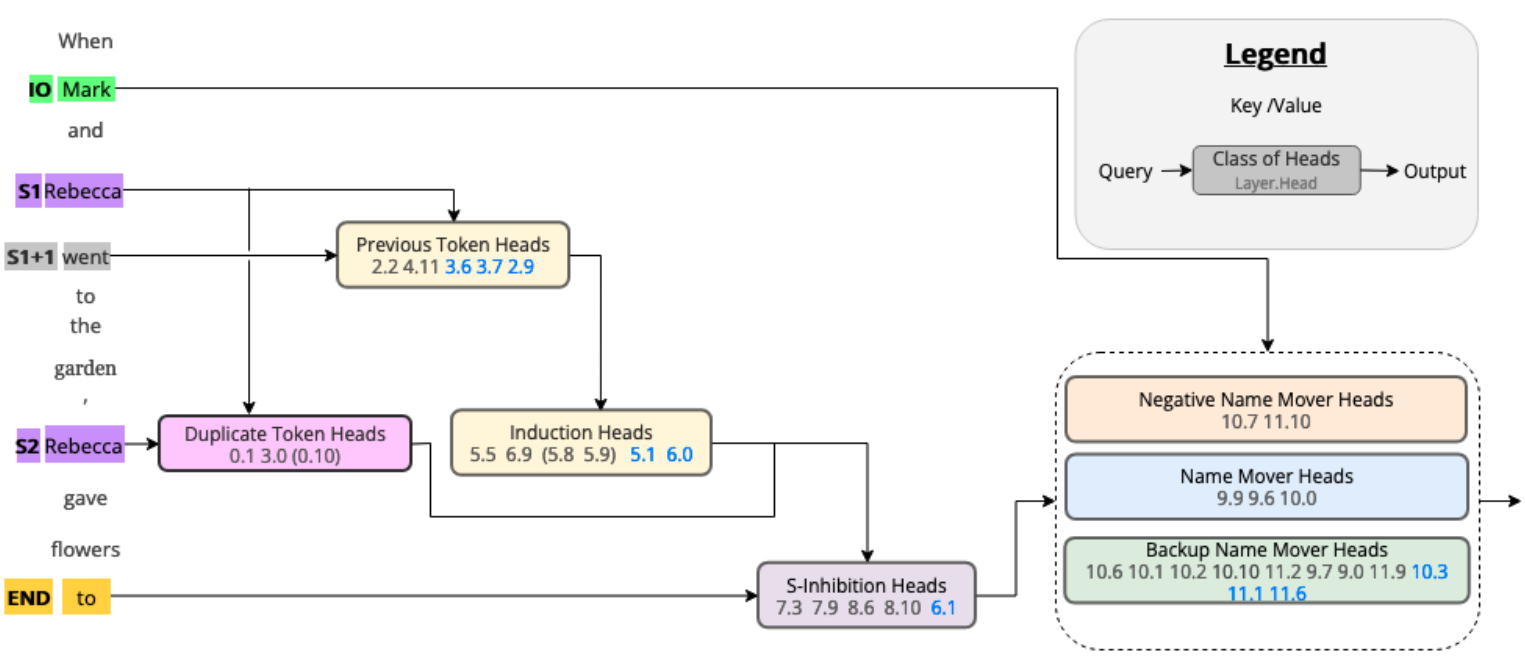}
    \caption{The circuit discovered \textit{post-reversal} after corruption on Subject Duplication Augmentation, the new components are marked in \textcolor{blue}{blue}.}
    \label{fig:neurocircsd}
\end{figure*}
\begin{figure}
    \centering
    \includegraphics[width = 0.45\textwidth]{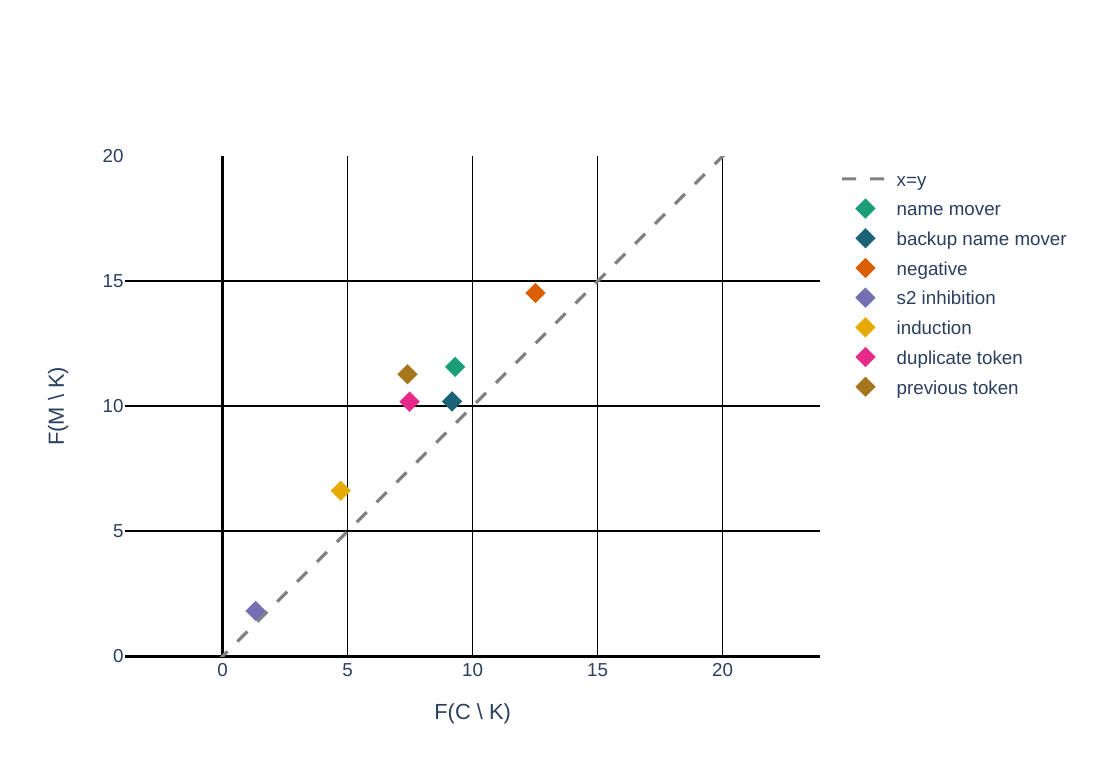}
    \caption{Completeness scores of the circuit discovered in \autoref{fig:neurocircsd}}
    \label{fig:comp-np-sd}
\end{figure}

\textbf{Data Augmentation: Subject Duplication}: We present the circuit for the relearned mechanisms in the \textit{post-reversal} model after corruption on Subject Duplication Task, see \autoref{fig:neurocircsd}.\\
The faithfulness score of this model is  $96\%$ with identical minimality scores as \textit{post-reversal} with Name Moving Behavior, for the completeness scores see \autoref{fig:comp-np-sd}. 

\section{Discovering Localized Corruption with Cross-Model Activation Patching}
\label{app:cross}
\begin{figure}
    \centering
    \includegraphics[width = 0.48\textwidth]{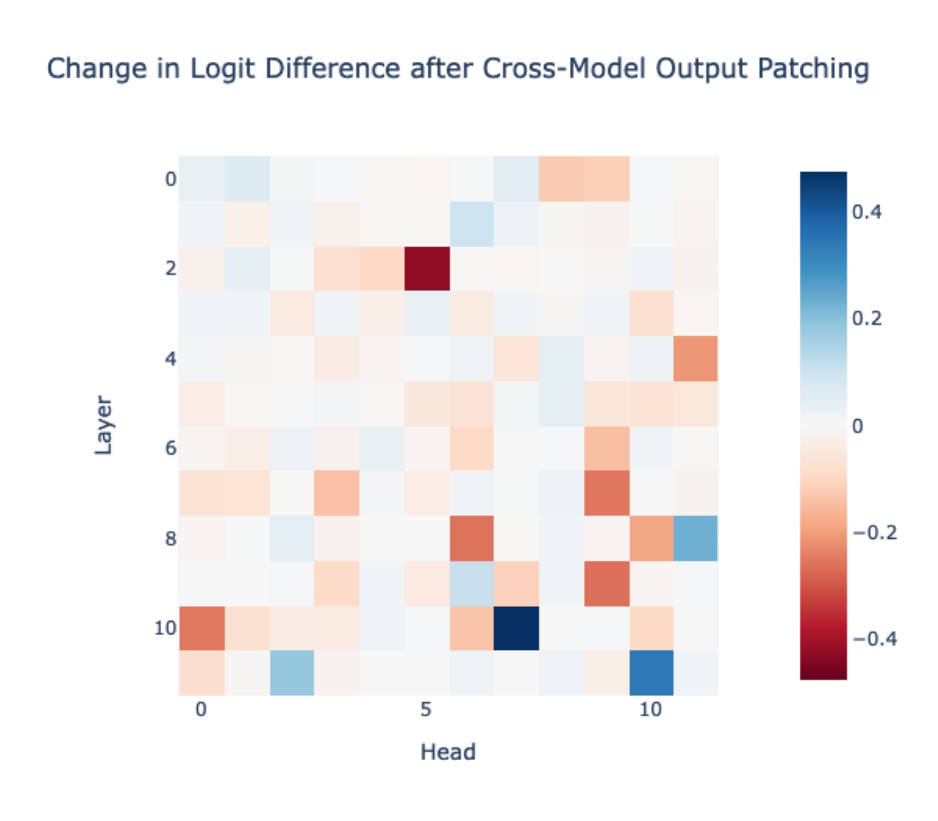}
    \caption{Change in Logit Difference after Cross Model Output Patching on the Original Model}
    \label{fig:cmap-sd-out}
\end{figure}

\begin{figure}
    \centering
    \includegraphics[width = 0.48\textwidth]{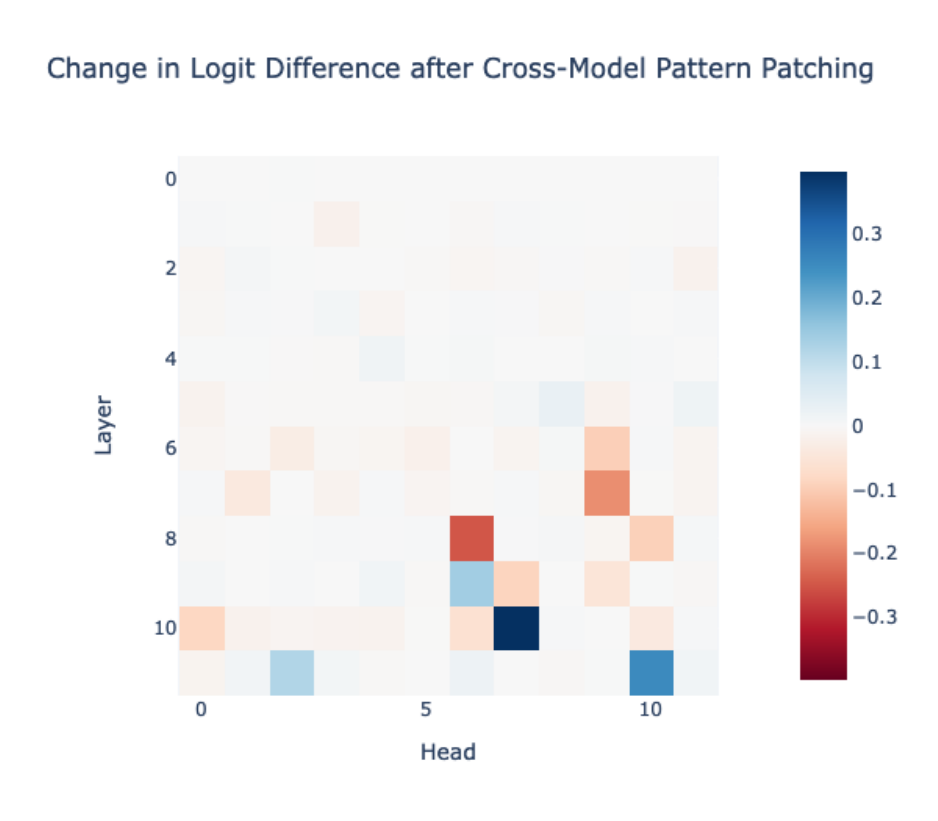}
    \caption{Change in Logit Difference after Cross Model Output Patching on the Original Model}
    \label{fig:cmap-nm-pat}
\end{figure}

\begin{figure}
    \centering
    \includegraphics[width = 0.48\textwidth]{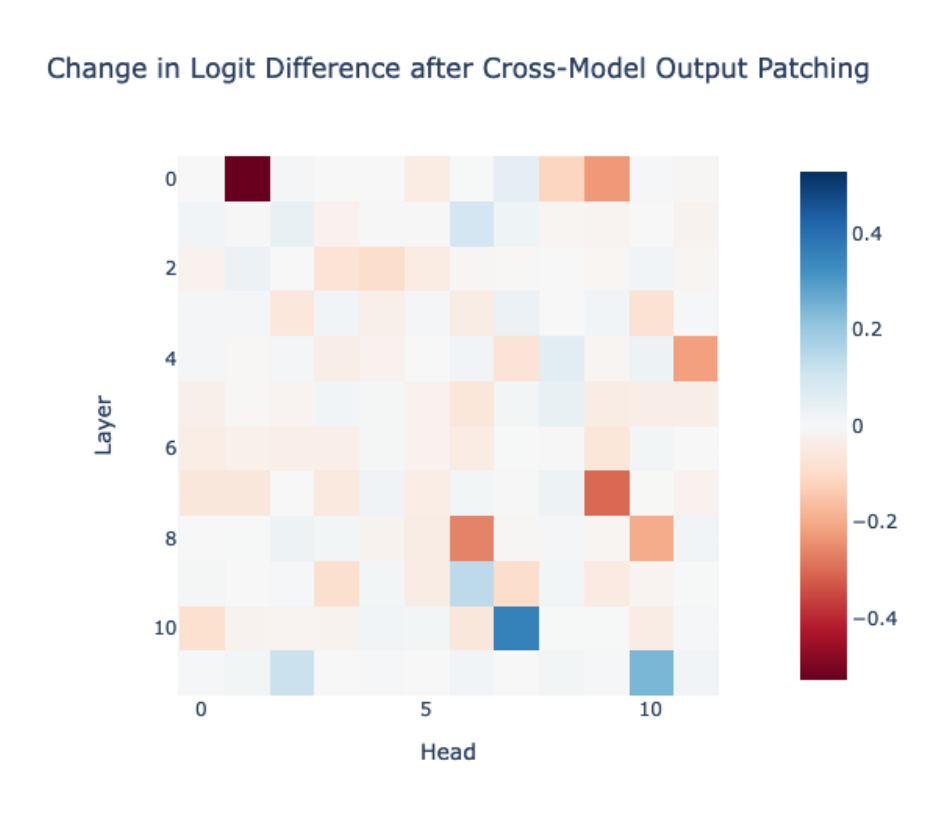}
    \caption{Change in Logit Difference after Cross Model Output Patching on the Original Model}
    \label{fig:cmap-nm-out}
\end{figure}

\textbf{Data Corruption: Subject Duplication}: In addition to the Cross Model Pattern Patching we also employ Cross Model Output Patching, i.e, replacing the attention outputs of each attention head in the original model with that of the fine-tuned on corrupted data variant. We record that the prior analysis of localized corruption can also be examined via Cross-Model Output Patching, see \autoref{fig:cmap-sd-out}
\autoref{fig:cmap-sd-out} illustrates that majority of the corruption is localized to the original circuit components, however similar to our prior analyses novel components arise with perform repeated corrupted mechanism and hence we see their contribution to the task. An interesting case here is that of \textbf{L8H11} which is a new former Name Mover Head, i.e, moving the ''S'' token to the residual stream at the END position. In \autoref{fig:cross-model-cp} we saw that the attention pattern of L8H11 when patched results in decrease in overall capability of the model, however in \autoref{fig:cmap-sd-out} shows an increase in capability, this is a non-surprising result as the OV Matrix of each attention head determines what is written to the residual stream whereas the QK matrix determines the attention pattern, here, we see that the QK Matrix of L8H11 decreases performance after CMAP however OV Matrix doesn't, this is due to the linearly independent nature of the two operations, which only in conjunction, determine the contribution of the head. As the QK Matrix is negatively contributing after CMAP and OV Matrix is positively contributing, this means that overall contribution is negative as the head copies ''S'' token to the residual stream of the END token.\\
\textbf{Data Corruption: Name Moving}: In addition to the localized corruption in subject duplication task, we identify localized corruption in the model variant fine-tuned on the Name Moving data corruption. Firstly, similar to our prior analyses we employ Cross Model Pattern Patching, see \autoref{fig:cmap-nm-pat}.
Hence see that the corruption, in this case, is localized to the circuit components, we further validate our findings via Cross-Model Output Patching, see \autoref{fig:cmap-nm-out}.

\section{Effect of MLP Across Epochs}
\label{app:mlp}
\begin{figure}
    \centering
    \includegraphics[width=0.49\textwidth]{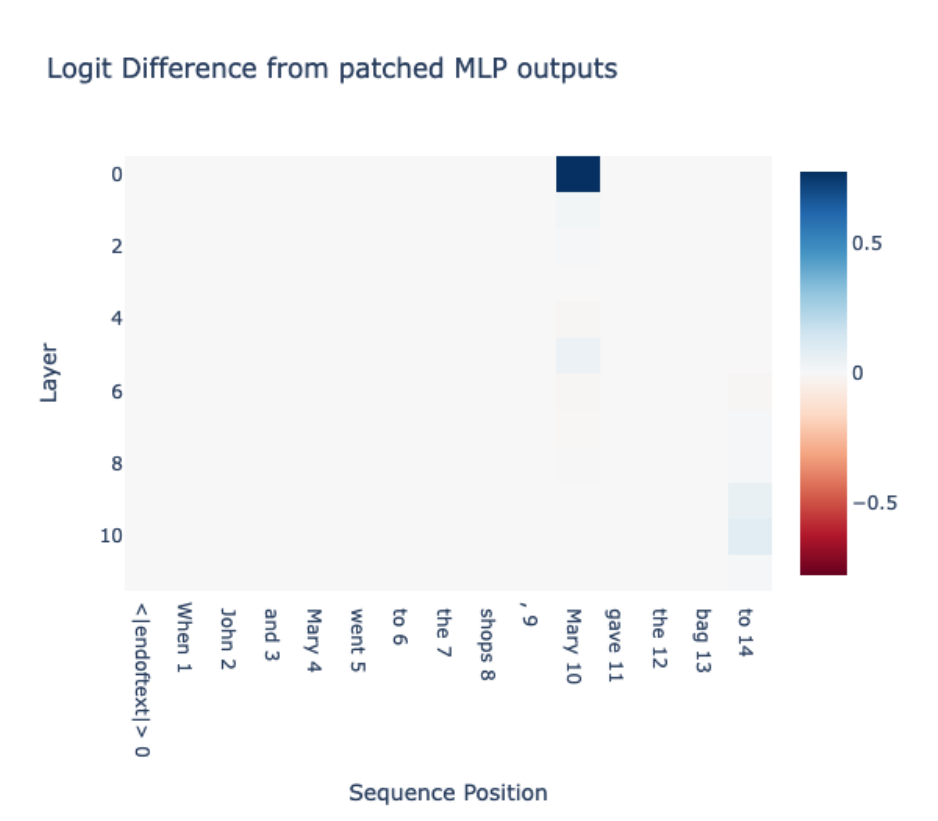}
    \caption{Logit Difference from patched MLP outputs on the model fine-tuned for 3 Epochs on the original dataset}
    \label{fig:mlp-amp}
\end{figure}

\begin{figure}
    \centering
    \includegraphics[width=0.49\textwidth]{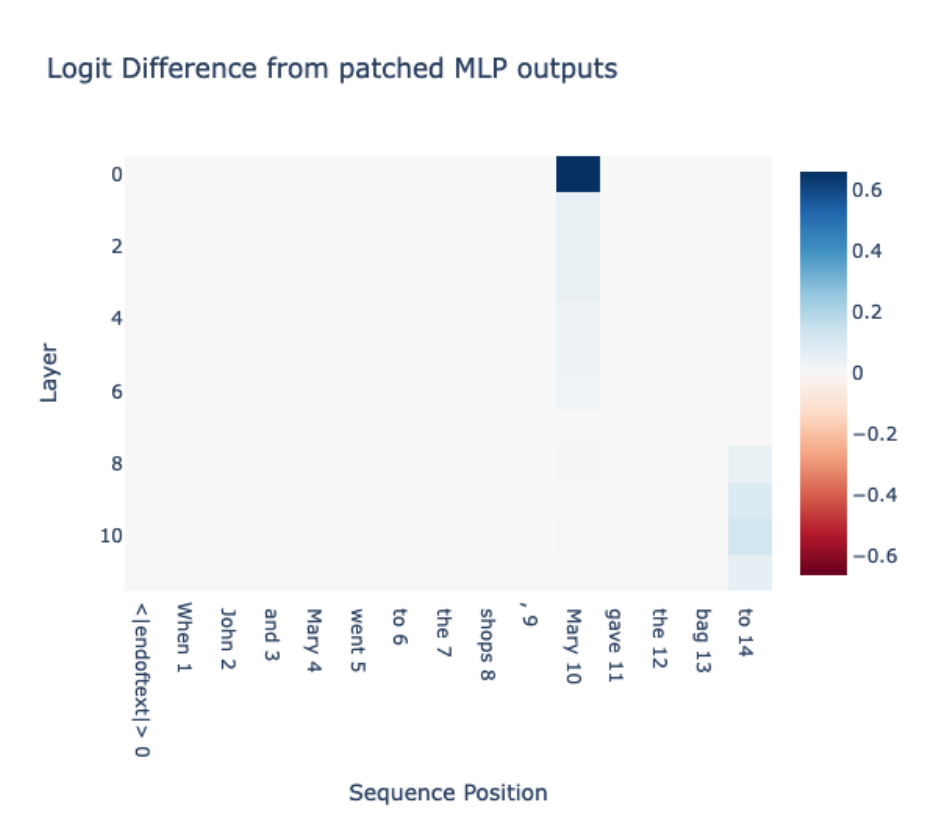}
    \caption{Logit Difference from patched MLP outputs on the model fine-tuned for 5 Epochs on the Subject Duplication Dataset}
    \label{fig:mlp-cp-sd}
\end{figure}

\begin{figure}
    \centering
    \includegraphics[width=0.48\textwidth]{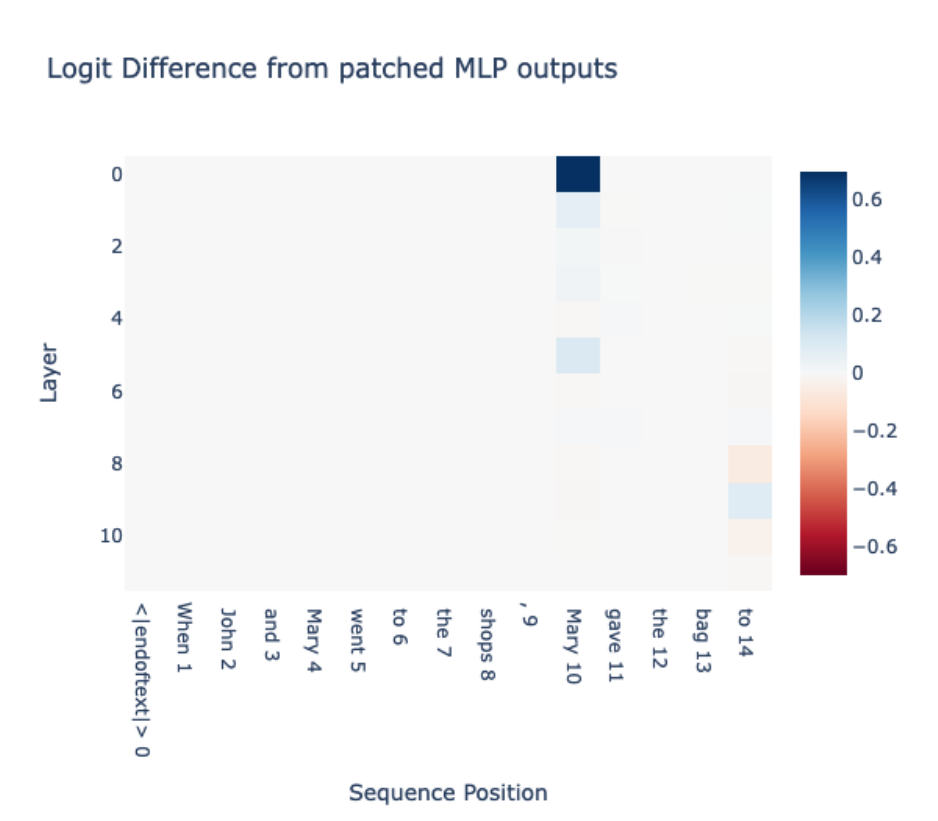}
    \caption{Logit Difference from patched MLP outputs on the model fine-tuned for 3 Epochs on the Name Moving Corrupted Dataset}
    \label{fig:mlp-cp-nm}
\end{figure}

\begin{figure}
    \centering
    \includegraphics[width=0.48\textwidth]{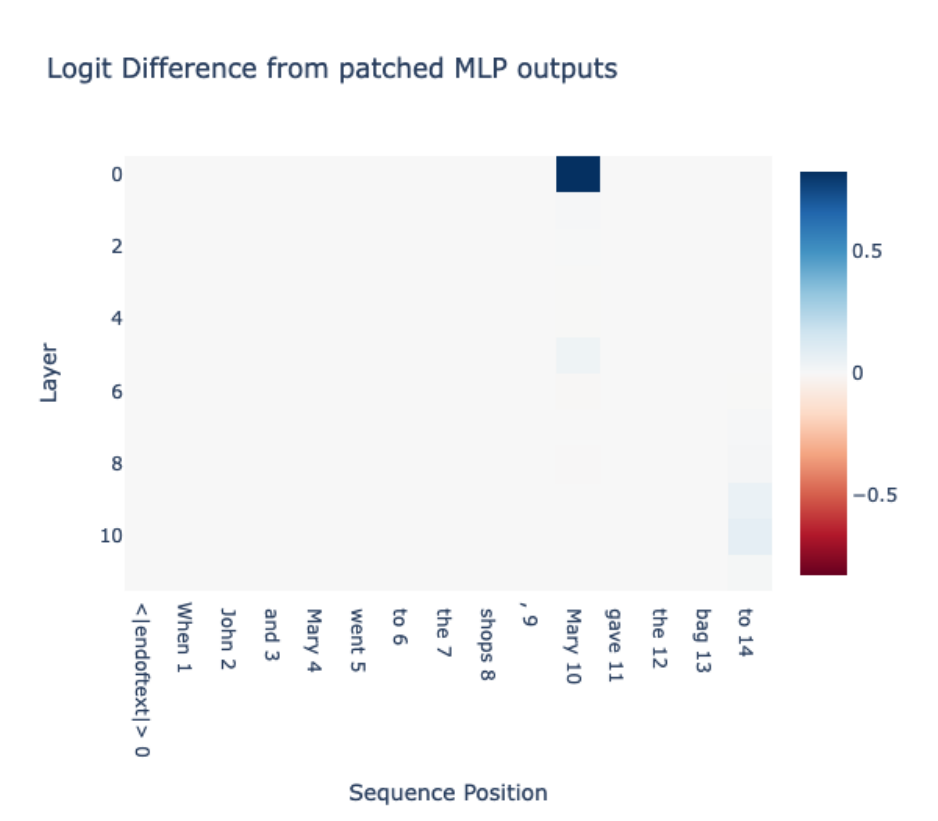}
    \caption{Logit Difference from patched MLP outputs on the model fine-tuned for 5 Epochs on the Subject Duplication Dataset and then fine-tuned on the original dataset for 5 epochs}
    \label{fig:mlp-np-sd}
\end{figure}

\begin{figure}
    \centering
    \includegraphics[width=0.48\textwidth]{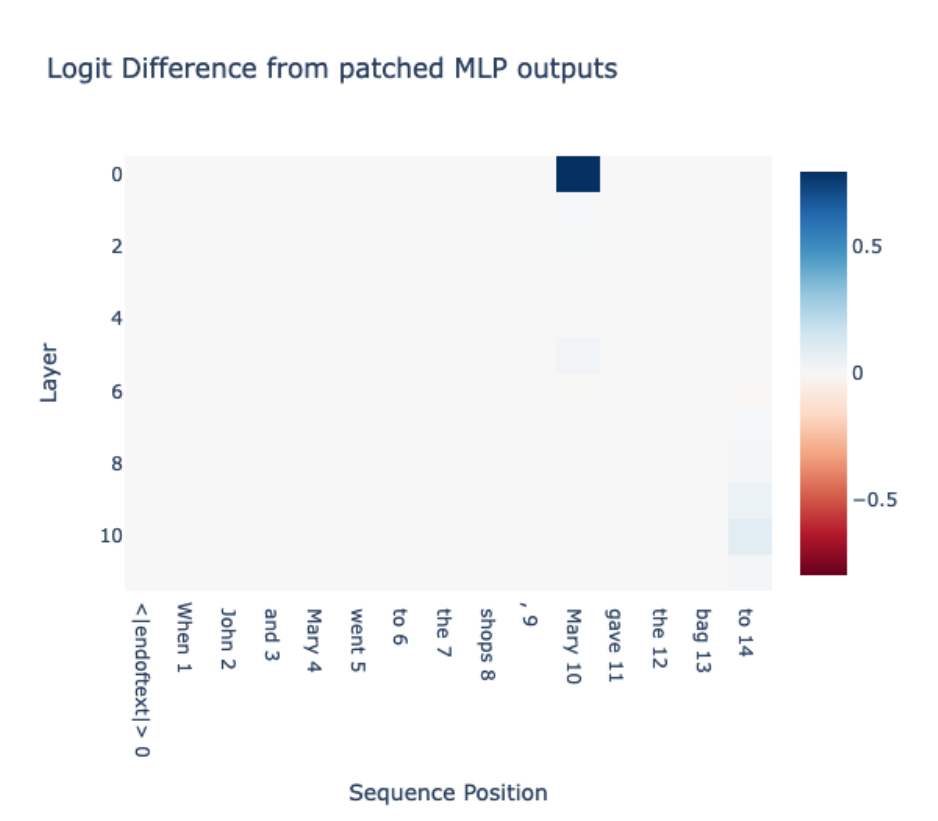}
    \caption{Logit Difference from patched MLP outputs on the model fine-tuned for 3 Epochs on the Name Moving Corruption Dataset and then fine-tuned on the original dataset for 3 epochs}
    \label{fig:mlp-np-nm}
\end{figure}

In the original work, \cite{wang2022interpretability}, MLP layers of GPT2-small do not individually contribute much to the task, except MLP layer 0, which is seen as an extended embedding \cite{wang2022interpretability}. We find this case to extend to the circuits we recover via fine-tuning on the original IOI dataset, furthermore, we do not record any major contribution of the MLP layers (except MLP layer 0) in the corruption of the IOI task after fine-tuning on corrupted data variants. \\
\textbf{Amplification}: Similar to the original model, we record that the MLP layers, except layer 0, have no statistically significant contribution to the IOI task even after undergoing task-specific fine-tuning on the clean dataset, see \autoref{fig:mlp-amp}.\\
\textbf{Corruption}: We analyze the performance/contribution of the MLP Layers for the Subject Duplication Task and find that, similar to our prior analysis, the contribution of the MLPs remain minuscule even after fine-tuning on the corrupted data variants, see \autoref{fig:mlp-cp-sd}.
We also find that this analyses extends to the Name Moving data corruption as well, see \autoref{fig:mlp-cp-nm}.\\
\textbf{Neuroplasticity}: In addition to the case of amplification and corruption we find that our prior analyses extends to the case of the circuits formed \textit{post-reversal}, see \autoref{fig:mlp-np-sd} and \autoref{fig:mlp-np-nm}.

\section{Corrupted Dataset: Duplication}
\label{dadupe}
As we are aware of the circuit and mechanism of the IOI task \textit{a priori}, we augment the data to inhibit the backup/duplication behavior of the  Duplicate Token Heads and Induction Heads by replacing the S2 token with a random single-token name. For example: 
\textit{"When Mark and Rebecca went to the garden, \textcolor{red}{Tim} gave flowers to Rebecca"}.\\

\textbf{Experimental Conclusion}: In the case of this particular corrupted data augmentation, we find that there is no statistically significant change in the model mechanisms across a variety of epochs. However, further explorations are needed to justify the robustness of the model to this type of corruption which we leave for future work.

\section{Greater-Than Task}
\label{app:gt}
The greater-than circuit \cite{hanna2024does} is a circuit for the greater-than year span prediction task for GPT2-small which can be defined as "The war lasted from the year 17XX to the year 17" and the model outputs any number (YY) greater than XX and less than 99. Complete details of the circuit can be found in \citet{hanna2024does}. As for the circuit discovery procedure we utilize Edge Attribution Patching with Integrated Gradient (EAP-IG), a novel automatic circuit discovery procedure introduced in \citet{hanna2024have}. As for evaluation, we utilize the probability difference between years greater than XX and years less than YY\footnote{This metric is defined on page 3 of \citet{hanna2024does}}.\\

\subsection{Amplification of Circuit}
\begin{figure*}
    \centering
    \includegraphics[width=\textwidth]{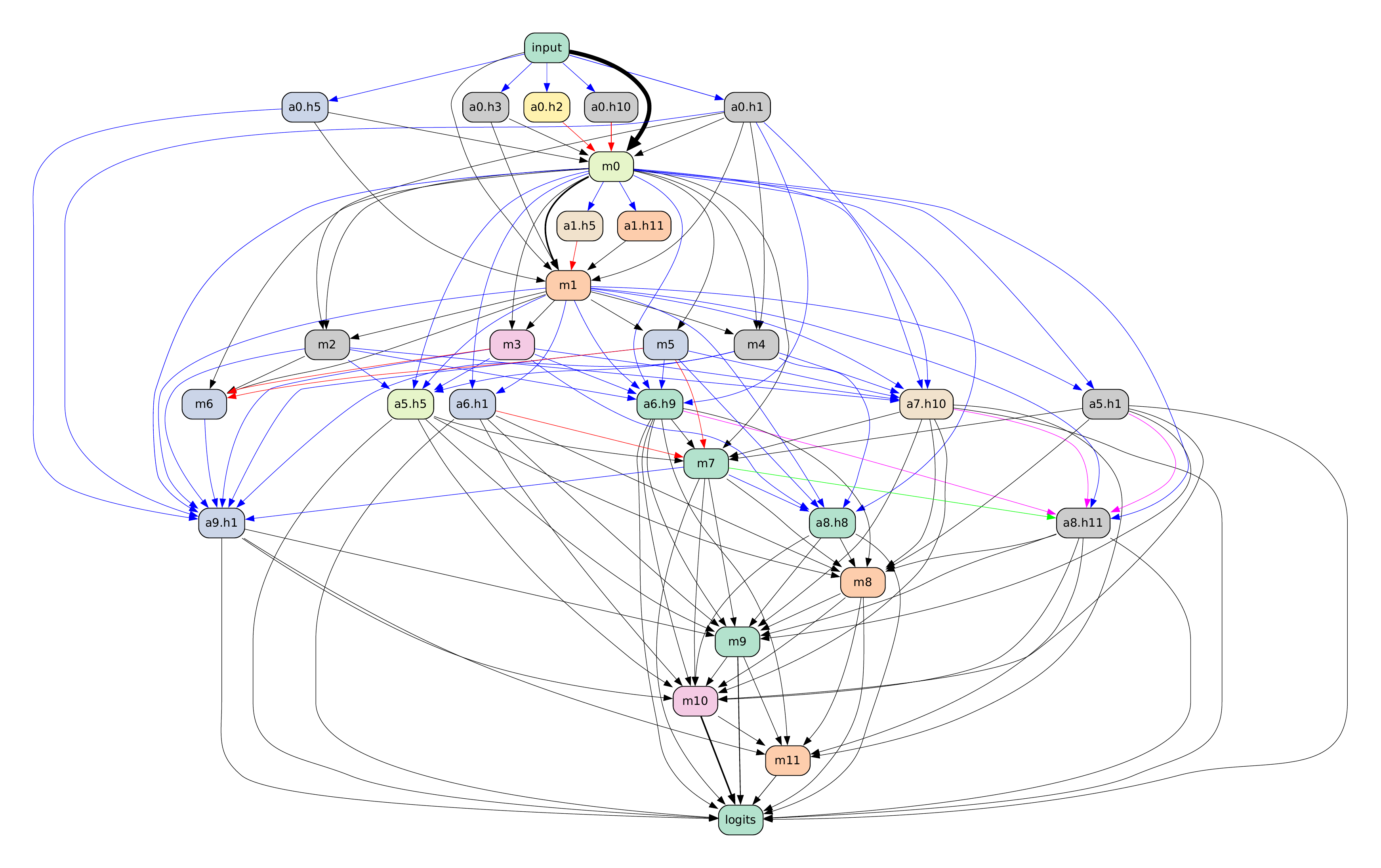}
    \caption{The circuit for the greater than task after fine-tuning for 3 epochs, attention head for layer 9 and head 1 is represented as a9.h1 and MLP of layer 11 is represented m11}
    \label{fig:gtcircuitclean}
\end{figure*}

We take the case of fine-tuning GPT-2-small on the task-specific greater-than data for 3 epochs. First, we present the discovered circuit, see \autoref{fig:gtcircuitclean}, and record that the circuit is similar to the original greater-than circuit presented in \citet{hanna2024does}. 
This novel circuit itself performs as well as base GPT-2-small on the task, achieving a $84\%$ probability difference on the task while the full model achieves a  $95\%$ probability difference on the task. \\
As most circuit components are similar we can assess what makes the model perform better. This analysis is two-fold. We first utilize logit lens \cite{lesswrongInterpretingGPT} and attention pattern analysis to analyze the change in the mechanism of the relevant attention heads ( taking the example attention head L9H1). We then utilize logit lens to interpret the deviation from the original mechanism for the MLP that are important to the task ( taking the example of MLP 9). \\

\begin{figure}
    \centering
    \includegraphics[width=0.99\linewidth]{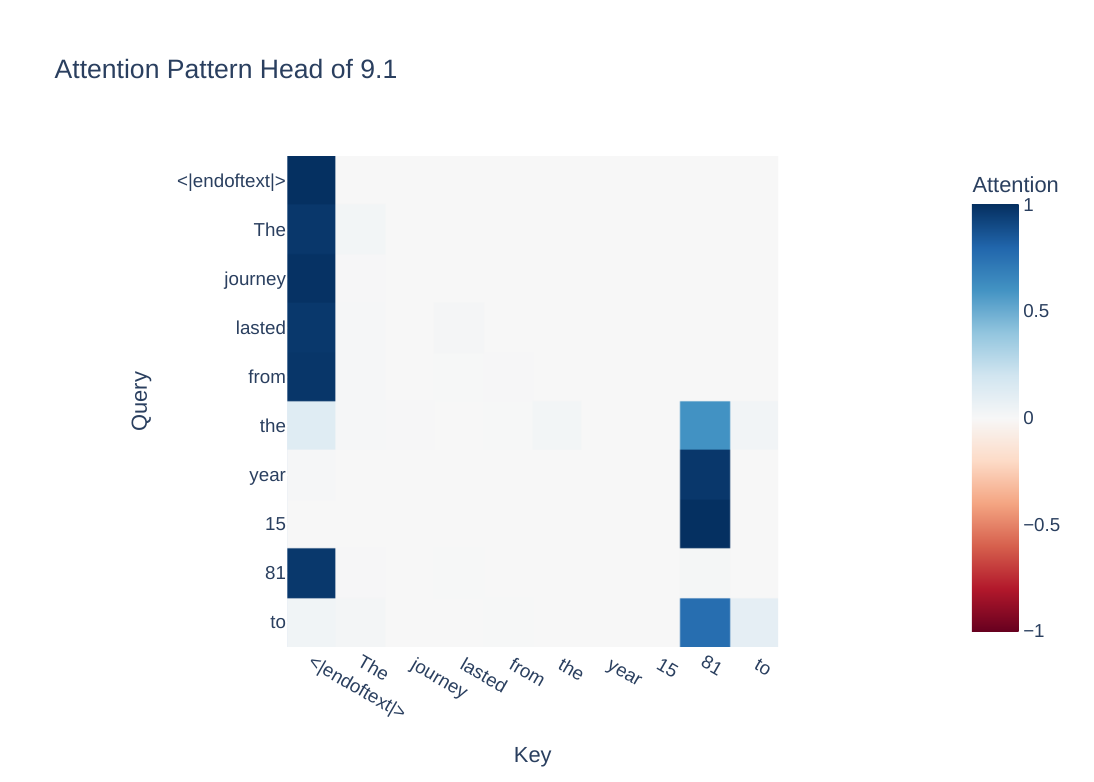}
    \caption{Attenion for Head L9H1}
    \label{fig:gt_amp_attn91}
\end{figure}

\textbf{Amplification of the attention heads}: We first visualize the attention pattern of the relevant attention heads (taking the case of L9H1 for illustration) and notice that it is very similar patterns originally observed\footnote{see page 6 of \citet{hanna2024does}} by \citet{hanna2024does}, see \autoref{fig:gt_amp_attn91},i.e , the head attends strongly the to XX year for which the prediction has to be made. From this we can realize that there is no mechanistic change to the attention head given that it behaves similarly in that it writes to the final logit and influences MLP9 so, see \autoref{fig:gtcircuitclean}. 
Now we utilize logit lens to visualize what the output of the attention head is writing to influence the final logit, see and find that it behaves similarly to what it did in the original model in that there is a majorly diagonal pattern to the logit lens similar to the observation\footnote{see Figure 7 of \citet{hanna2024does} } of \citet{hanna2024does}. \\
\begin{figure}
    \centering
    \includegraphics[width=0.99\linewidth]{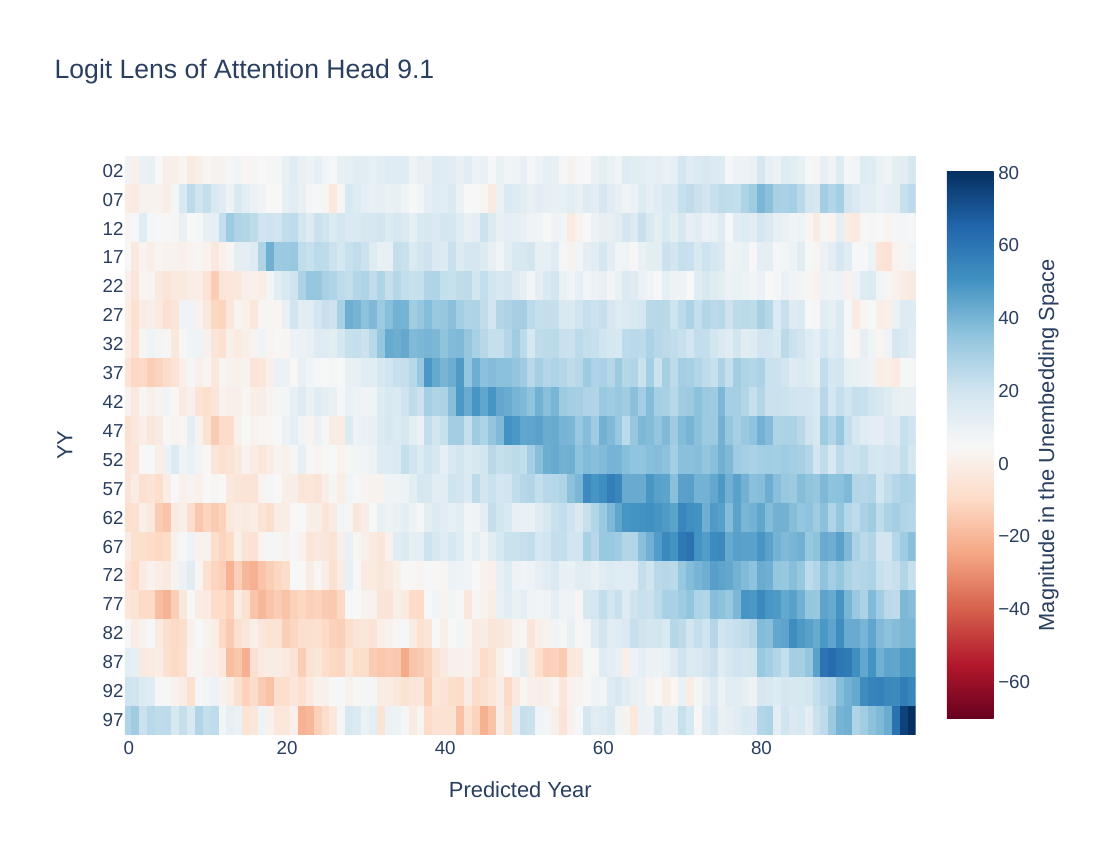}
    \caption{Logit Lens of Head L9H1 showing a spike in the projection of the heads output in the unembedding space around the diagonal of the plot}
    \label{fig:enter-label}
\end{figure}
Furthermore, we also see report that the average magnitude of the diagonal year (i.e the same year as XX) in the unembedding space is $36.72$ in the fine-tuned model whereas it is $17.31$ in the original model this shows that output of the attention head to logit is \textbf{amplified.} This analysis extends to other heads in the circuit, as they have similar functionality. \\
\textbf{Amplification of the MLPs}: To see the amplification of the MLPs we take the case of MLP9 and use logit lens to visualize what it is writing to the logit and find that "upper-triangular" pattern as first shown by \citet{hanna2024does} holds true,see \autoref{fig:gtllmlp9}, furthermore there are differences up to the value of $140$ between some years higher than XX and lower than XX compared to the original model in which the differences can be up to $40$\footnote{see figure 8 of \citet{hanna2024does}}. This can generally be seen as the magnitudes of the years greater than XX are significantly higher than the base model, see \citet{hanna2024does} for reference. Indicating that the output of the MLPs is amplified while they retain the same mechanisms hence showing amplification. 
\begin{figure}
    \centering
    \includegraphics[width=0.99\linewidth]{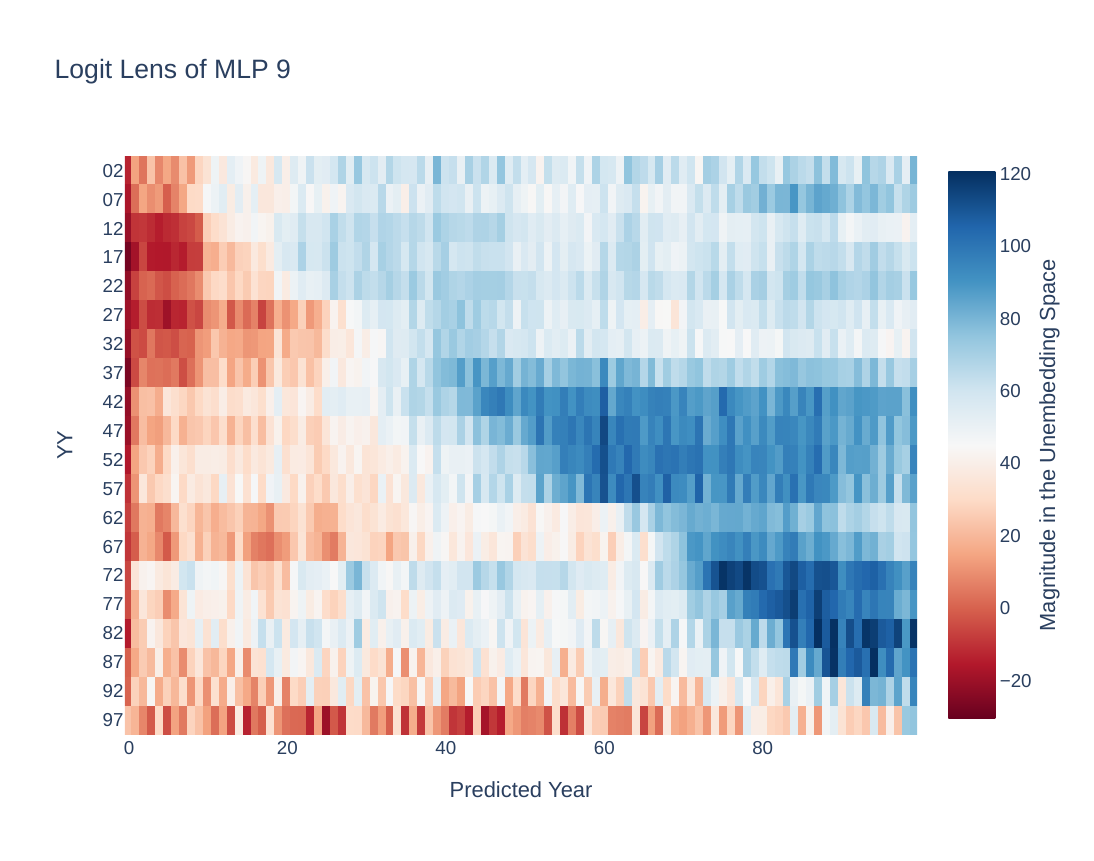}
    \caption{Logit Lens of MLP 9}
    \label{fig:gtllmlp9}
\end{figure}

\subsection{Corrupting of Model Mechanisms}
\begin{figure}[t]
    \centering
    \includegraphics[width=0.99\linewidth]{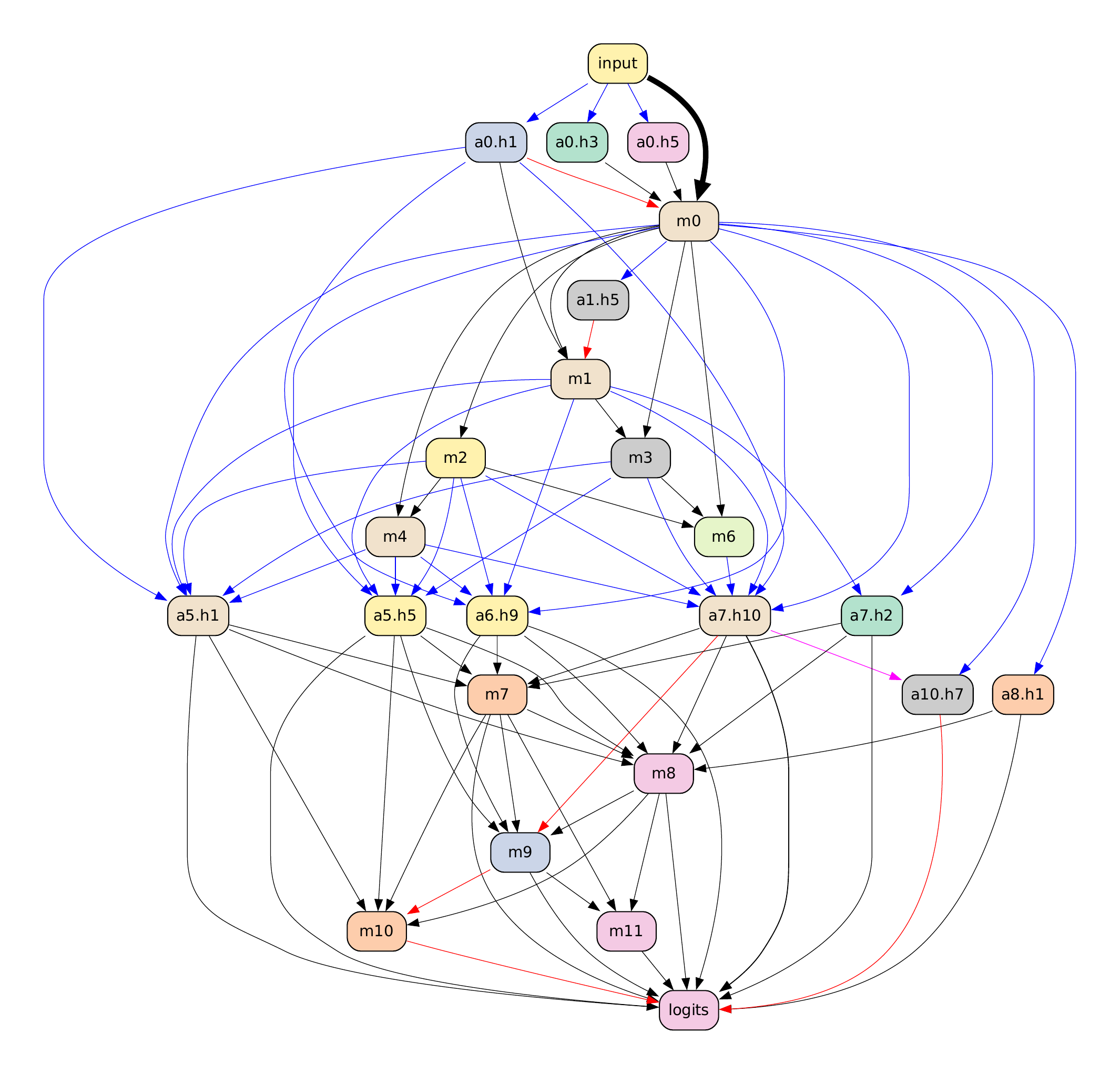}
    \caption{The circuit performing the "less than" task in the new circuit after fine-tuning model on corrupted dataset for 3 epochs}
    \label{fig:gt_cp}
\end{figure}

\begin{figure}
    \centering
    \includegraphics[width=0.99\linewidth]{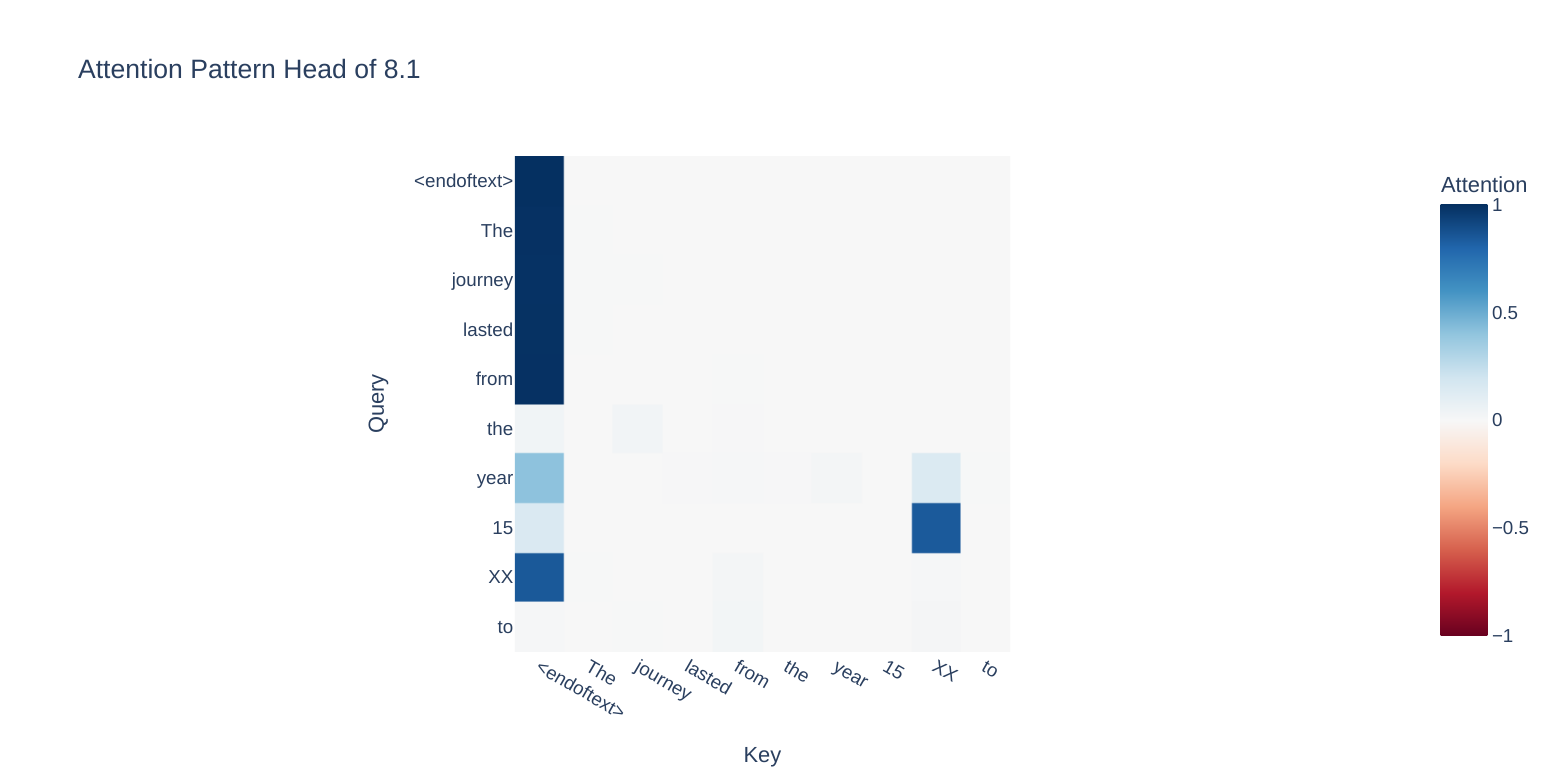}
    \caption{Attention Patterns for head L8H1}
    \label{fig:attn81}
\end{figure}

\textbf{Corrupted Dataset: Lower Than}: For corruption, we aim to target the mechanism of the MLPs which makes them increase the projection of years greater than XX in unembedding space, so for this, we craft the Lower Than task which is grammatically incorrect but corrupts the mechanism of the MLPs.For this corruption we fine-tune the model by altering the year to be less than XX, for example, "The war lasted from the year 1713 to the year 17\textcolor{teal}{17}" becomes "The war lasted from the year 1713 to the year 17\textcolor{red}{12}". The main reason why we chose a grammatically incorrect task is to target the functionality of the MLPs. \\
\textbf{Mechanism of Corruption}: Firstly, we note that the model after toxic fine-tuning output years \textbf{less than} XX, the probability difference of $-97\%$ (the total probability of years after XX - the total probability of years before XX) after just 3 epochs of fine-tuning on the corrupted data. So the model's ability to perform greater-than year prediction is successfully corrupted. We now present the circuit that performs the new "lower-than" task, see \autoref{fig:gt_cp} and note that a majority of the attention heads are ablated from the circuit. With the attention heads that still remain show a similar attention head pattern to the original model, to illustrate we visualize the attention pattern of attention head L8H1 and notice it still strongly attends to the XX year, see \autoref{fig:attn81}. \\

\begin{figure}
    \centering
    \includegraphics[width=0.99\linewidth]{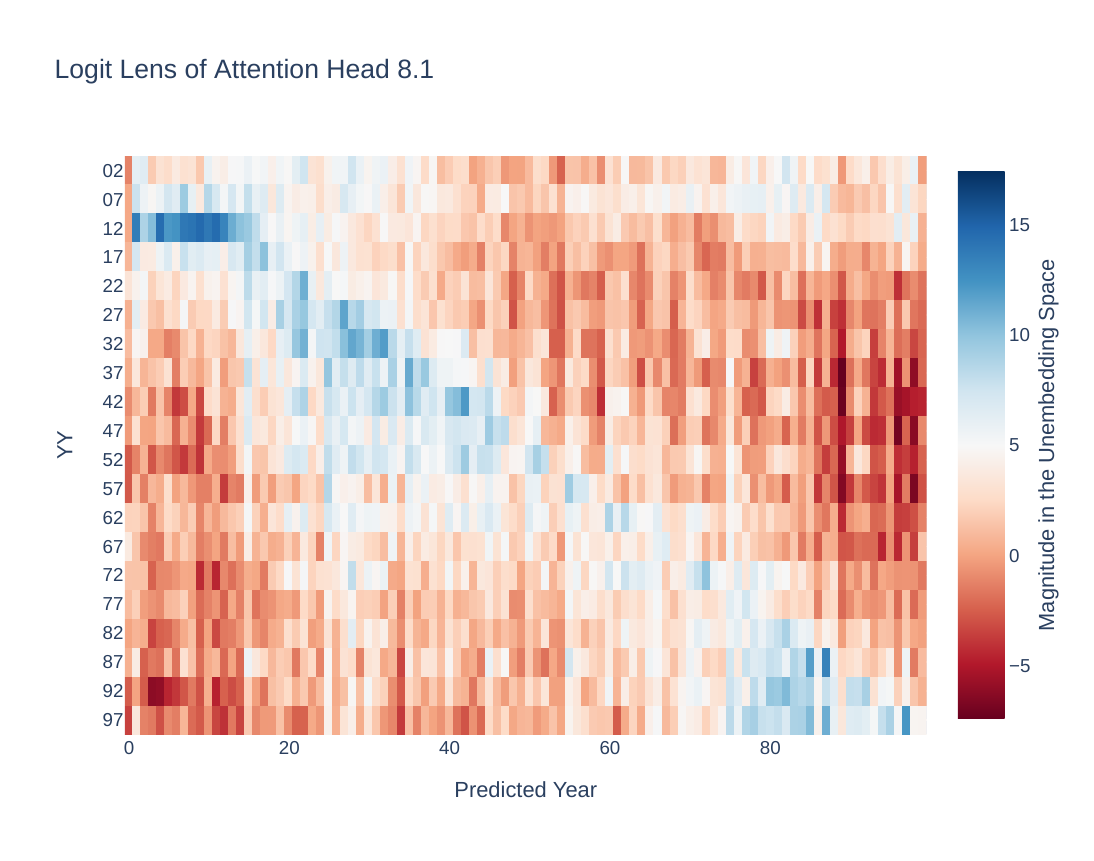}
    \caption{Logit Lens for head L8H1}
    \label{fig:llattn81}
\end{figure}

Furthermore, we utilize logit lens, see \autoref{fig:llattn81} for L8H1 and notice that it shows a similar diagonal pattern and it's mechanism remains to be fairly similar. Effectively we see that a majority of heads that aided in the greater than task are ablated with no new addition of novel heads/mechanisms and hence can conclude that the effect of corruption is localized to the circuit components. 

\begin{figure}
    \centering
    \includegraphics[width=0.99\linewidth]{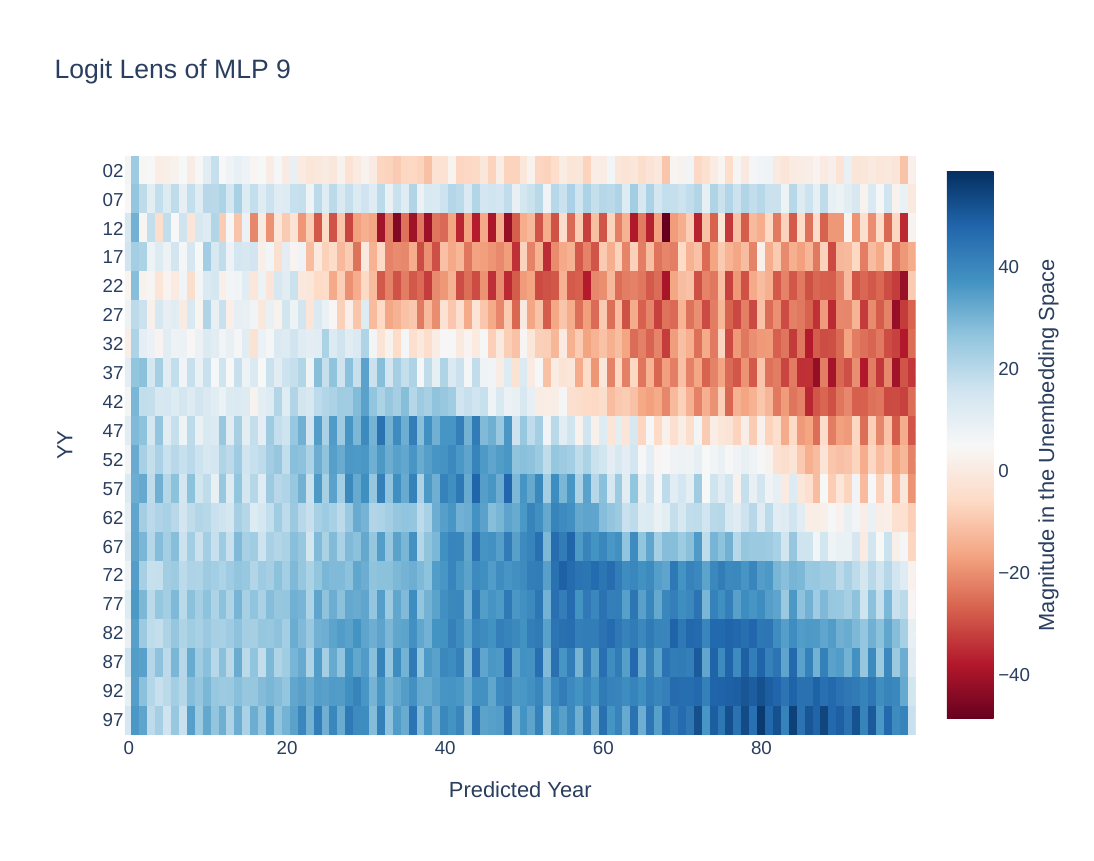}
    \caption{Logit Lens of MLP9}
    \label{fig:corrmlp9}
\end{figure}

\textbf{Corruption of MLPs}: Given our analysis of attention heads and the knowledge that their effect is fairly negligible except for a few attentions head like L8H1 we move to analyze the effect of corruption on MLPs. We analyze the logit lens of MLP9 and discover that instead of having an "upper-triangular" pattern it now has a lower triangular and significantly favors the years less than XX. This explains the fact that the model now successfully predicts the years to be less than XX, and hence we trace back the most impactful source of corruption, see \autoref{fig:corrmlp9}. This finding generalizes to other MLPs as well. \\

Now given that a majority of the attention heads don't contribute much to the corrupted performance of the model(the ones that do are similar in their mechanisms to the original model) and that MLPs effectively "switch" their behavior from favoring years greater than XX to years less than XX, we conclude that the corruption is \textbf{localized} to the circuit components in the case of the "greater-than" circuit as well. 

\subsection{Neuroplasticity}
\begin{figure*}
    \centering
    \includegraphics[width=\textwidth]{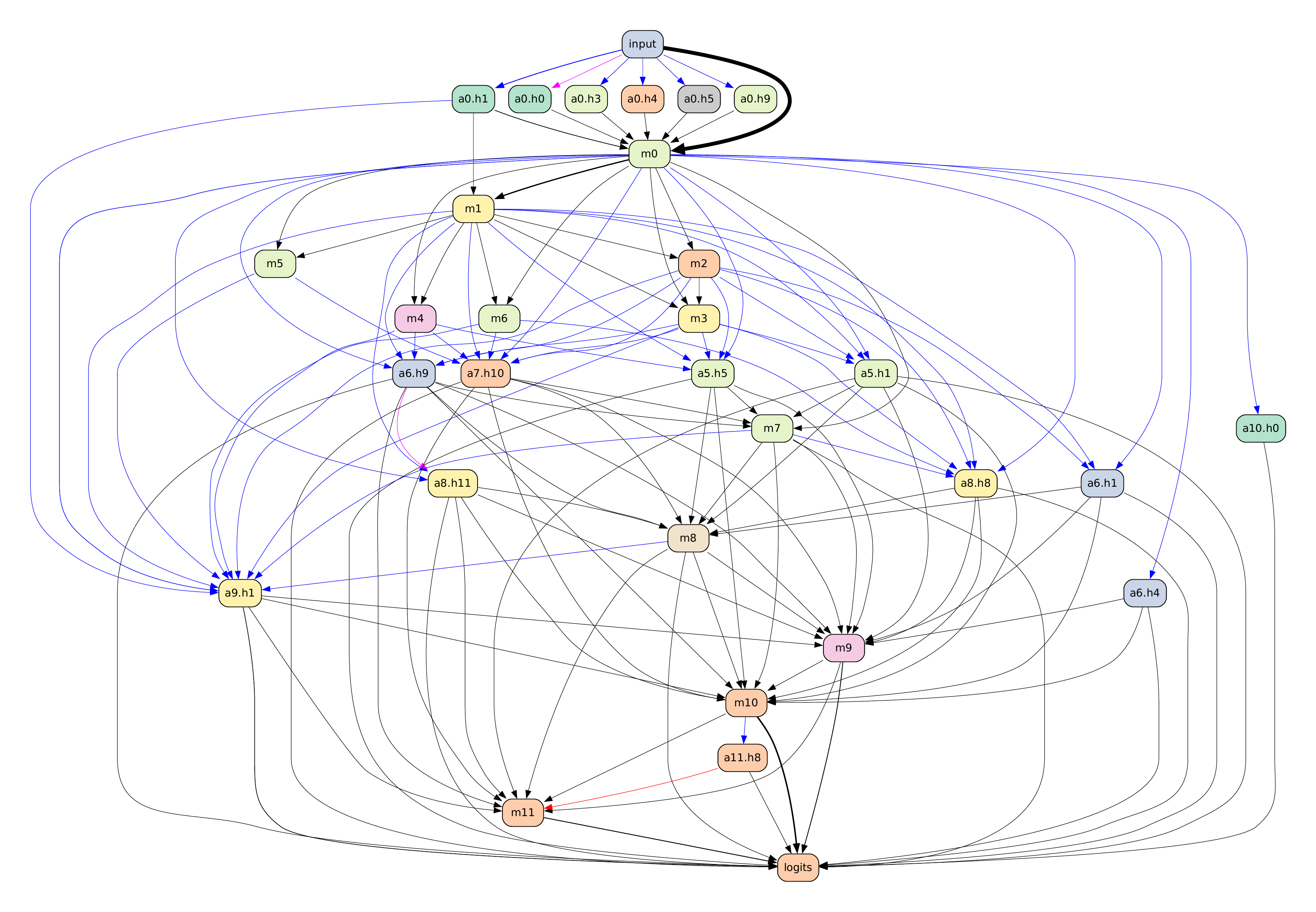}
    \caption{Circuit formed for greater than task after retraining the corrupted model for 3 epochs on the original dataset.}
    \label{fig:gt_neuro}
\end{figure*}

Similar to prior experiments in \autoref{main:neuro}, we retrain the model on the original greater-than dataset and find that the model relearns its original mechanism. Taking the case of retraining for 3 epochs this can be seen via the circuit formed for the task after retraining and its similarity to the original model, see \autoref{fig:gt_neuro}. The model now achieves a probability difference $94\%$ on the task while the circuit achieves $88\%$ of the total probability difference by itself.\\

\begin{figure}
    \centering
    \includegraphics[width=0.99\linewidth]{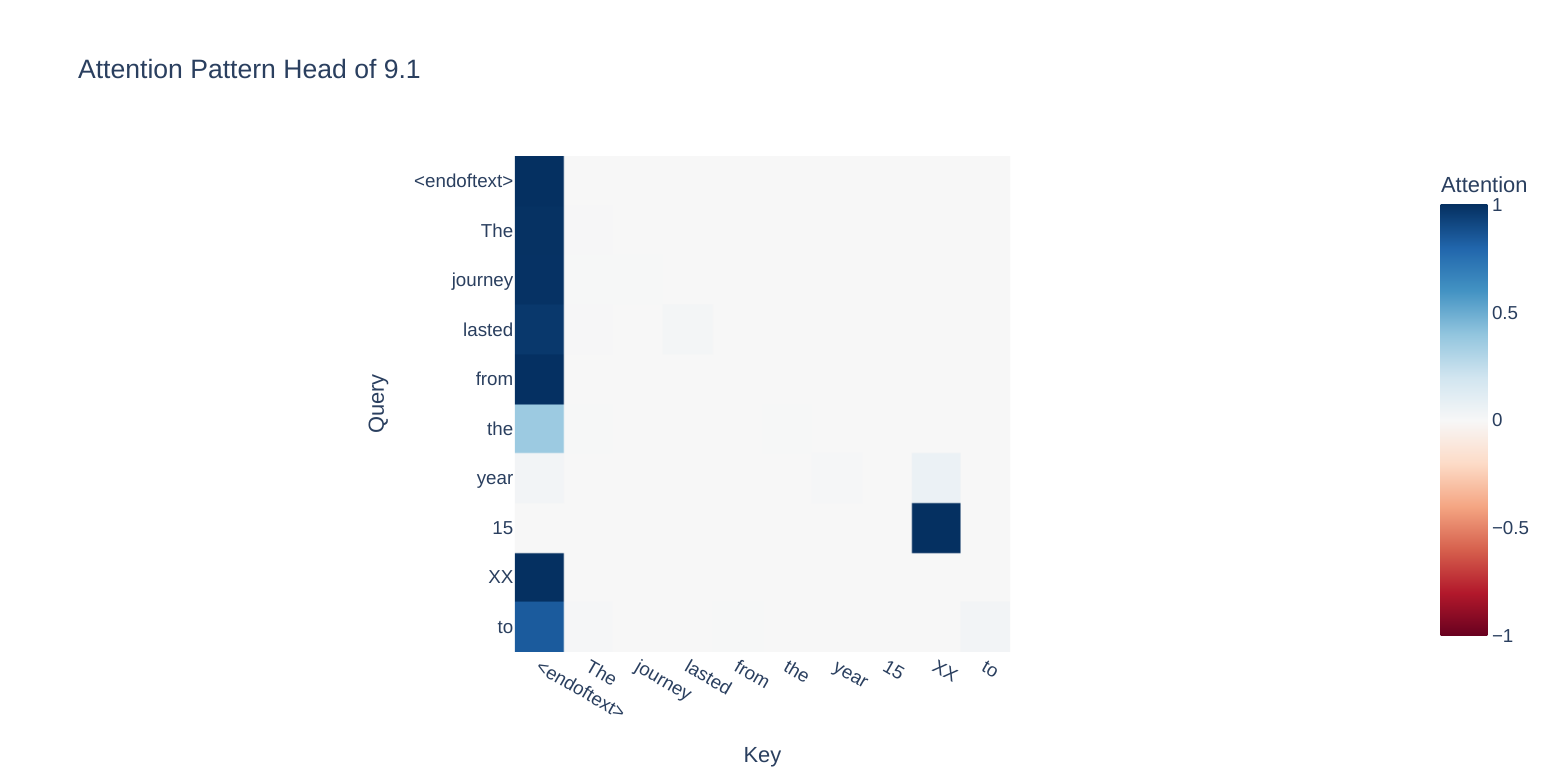}
    \caption{Attention Pattern of L9H1 after retraining on clean data}
    \label{fig:attn91neuro}
\end{figure}

\begin{figure}
    \centering
    \includegraphics[width=0.99\linewidth]{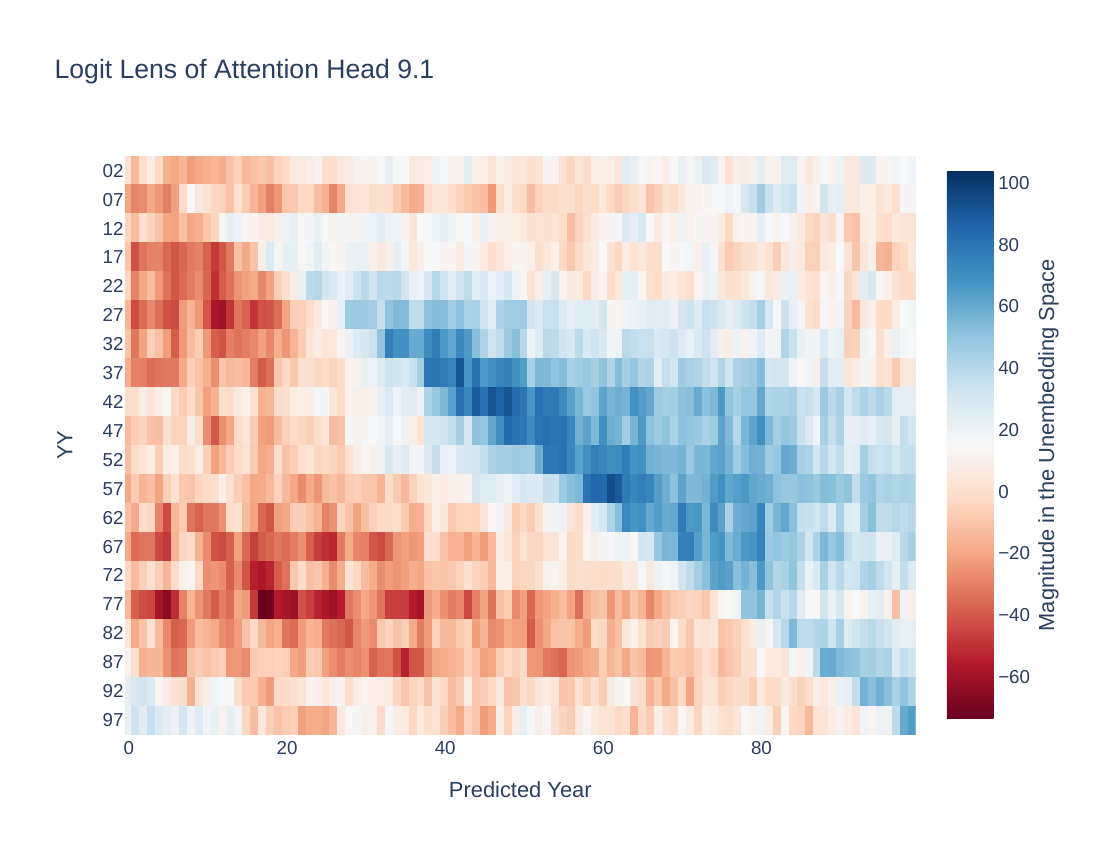}
    \caption{Logit Lens of L9H1 after retraining on clean data}
    \label{fig:llattn91neuro}
\end{figure}

\begin{figure}
    \centering
    \includegraphics[width=0.99\linewidth]{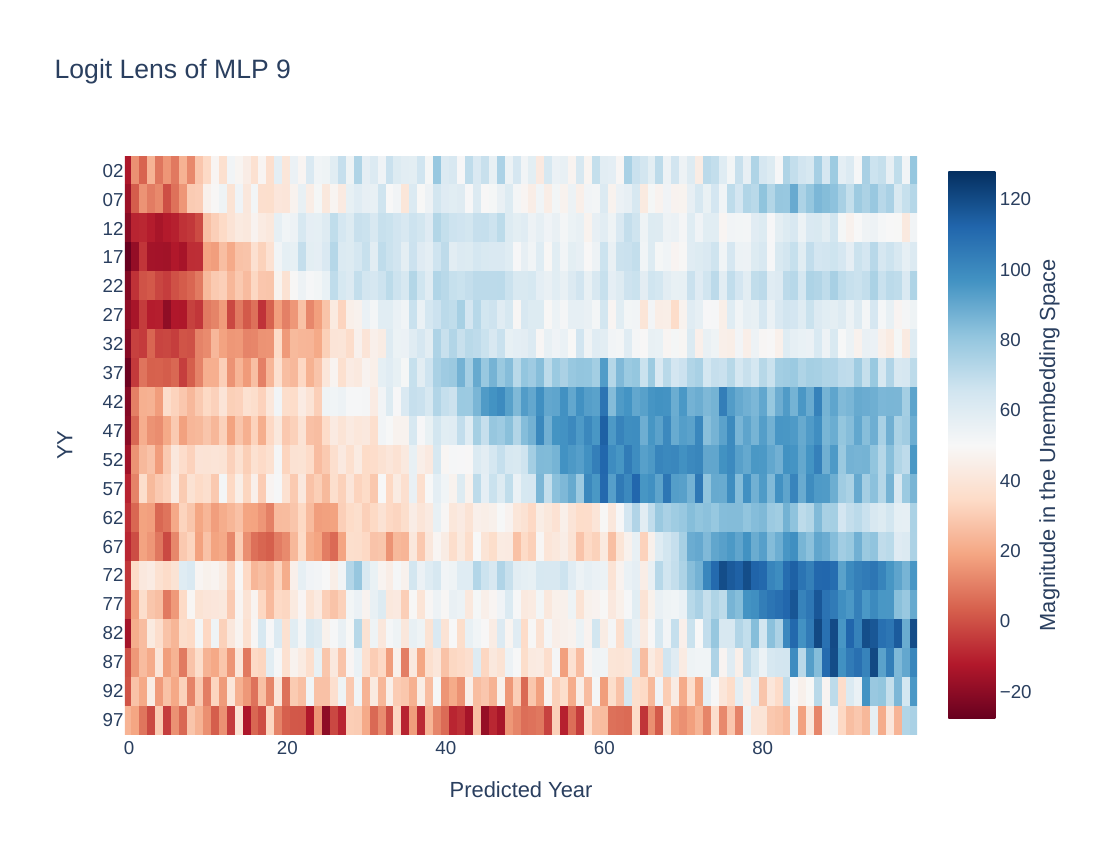}
    \caption{Logit Lens of MLP9 after retraining on clean data}
    \label{fig:neuromlp9}
\end{figure}

\textbf{Neuroplasticity of Attention Heads}: We can see that the attention heads that were ablated are formed back, see attention head L9H1 in \autoref{fig:gt_neuro} and its lack thereof in \autoref{fig:gt_cp} for illustration. We discover that the mechanism of the original attentions has been relearned and take the case of L9H1 to analyze. We visualize the logit lens and attention patterns of L9H1 and record that it is similar to the amplified/original version with the attention pattern showing strong attention, see \autoref{fig:attn91neuro}, to XX and the logit lens showing a diagonal pattern, see \autoref{fig:llattn91neuro}.\\
\textbf{Neuroplasticity of MLPs}: We take the case of MLP9 and show that the MLP has regained its original functionality via visualizing the logit lens of MLP9, see \autoref{fig:neuromlp9}. We now record that that pattern is "upper-triangular" with the MLP's output strongly favoring years greater than XX and hence reverting back to its original mechanism. 
\\
Now given, that the attention heads have regained their importance and contribution to the circuit( \autoref{fig:attn91neuro,fig:llattn91neuro,fig:gt_neuro }) and that the MLPs have reverted to their original mechanisms, we claim that the model has regained it's functionality for the greater than task, similar to the IOI case, after fine-tuning the corrupted model on the clean data. 

\end{document}